\documentclass{article}

    \PassOptionsToPackage{numbers, compress}{natbib}

\usepackage[final]{neurips_2025}

\usepackage[utf8]{inputenc} 
\usepackage[T1]{fontenc}    
\usepackage{hyperref}       
\usepackage{url}            
\usepackage{booktabs}       
\usepackage{amsfonts}       
\usepackage{nicefrac}       
\usepackage{microtype}      
\usepackage{xcolor}         

\usepackage{amsmath}
    \newcommand\numberthis{\addtocounter{equation}{1}\tag{\theequation}} 
\usepackage{amssymb}
\usepackage{amsthm} %
\usepackage{abbreviations}
\usepackage{algorithm}    
\usepackage{algpseudocode}
\usepackage{graphicx}
\usepackage{subcaption}
    
\usepackage{hyperref} 
    \hypersetup{ %
        pdftitle={},
        pdfsubject={},
        pdfkeywords={},
        pdfborder=0 0 0,
        pdfpagemode=UseNone,
        colorlinks=true,
        linkcolor=black,
        citecolor=mydarkblue, 
        filecolor=mydarkblue,
        urlcolor=mydarkblue,
    }

\addtocontents{toc}{\protect\setcounter{tocdepth}{0}} 

\title{Tractable Multinomial Logit Contextual Bandits \\with Non-Linear Utilities}

\author{%
Taehyun Hwang\\
Seoul National University\\
\texttt{th.hwang@snu.ac.kr} \\
\And
Dahngoon Kim\\
Seoul National University\\
\texttt{dahngoon.k@snu.ac.kr}\\
\And
Min-hwan Oh\\
Seoul National University\\
\texttt{minoh@snu.ac.kr}\\
}

\begin{document}

\maketitle

\begin{abstract}
We study the \textit{multinomial logit} (MNL) contextual bandit problem for sequential assortment selection. Although most existing research assumes utility functions to be linear in item features, this linearity assumption restricts the modeling of intricate interactions between items and user preferences. A recent work~\citep{zhang2024contextual} has investigated general utility function classes, yet its method faces fundamental trade-offs between computational tractability and statistical efficiency. To address this limitation, we propose a computationally efficient algorithm for MNL contextual bandits leveraging the upper confidence bound principle, specifically designed for non-linear parametric utility functions, including those modeled by neural networks. Under a realizability assumption and a mild geometric condition on the utility function class, our algorithm achieves a regret bound of $\tilde{\mathcal{O}}(\sqrt{T})$, where $T$ denotes the total number of rounds. Our result establishes that sharp $\tilde{\mathcal{O}}(\sqrt{T})$-regret is attainable even with neural network-based utilities, without relying on strong assumptions such as neural tangent kernel approximations. To the best of our knowledge, our proposed method is the first computationally tractable algorithm for MNL contextual bandits with non-linear utilities that provably attains $\tilde{\Ocal}(\sqrt{T})$ regret. Comprehensive numerical experiments validate the effectiveness of our approach, showing robust performance not only in realizable settings but also in scenarios with model misspecification.
\end{abstract}

\section{Introduction}
The \textit{multinomial logit} (MNL) contextual bandit~\cite{cheung2017thompson, ou2018multinomial, oh2019thompson, chen2020dynamic} is a model for sequential assortment selection under uncertainty prevalent in applications such as online recommendation systems, personalized marketing, and online retailing. 
In this setting, an agent repeatedly selects subsets of items to present to users, where user choice behavior is modeled using the MNL model~\cite{mcfadden1978modelling}. Specifically, the probability that a user selects an item is determined by its latent utility, which typically depends on contextual features associated with items and possibly user characteristics. 
The primary goal is to minimize expected regret, defined as the cumulative difference in rewards obtained by the algorithm compared to an oracle that knows the true utility parameters in advance.

Almost all prior studies on MNL contextual bandits assume the item utility functions to be linear in contextual features~\cite{cheung2017thompson, ou2018multinomial, oh2019thompson, chen2020dynamic, oh2021multinomial, perivier2022dynamic, agrawal2023tractable, zhang2023online, lee2024nearly, lee25improved}.
Although the linearity assumption simplifies computations and theoretical analyses, it significantly limits the expressive power required to capture complex and nonlinear user behaviors typically observed in real-world scenarios. 
Recent work by~\citet{zhang2024contextual} addresses this limitation by considering more general utility functions.
However, they encounter critical trade-offs: algorithms with computational tractability yield suboptimal regret (e.g., $\tilde{\Ocal}(T^{2/3})$ in stochastic and $\tilde{\Ocal}(T^{5/6})$ in adversarial settings), whereas their statistically efficient approaches become computationally infeasible due to non-convex optimization challenges.

Consequently, \textit{designing computationally efficient algorithms that achieve optimal statistical guarantees under general, non-linear utility functions remains an open and significant challenge}.

Motivated by these fundamental limitations, we investigate the MNL contextual bandit problem with utilities represented by general, non-linear parametric models, such as neural networks.
This setting introduces several fundamental challenges compared to the linear case.
First, the negative log-likelihood function used for parameter estimation lacks the self-concordant-like properties crucial for theoretical analysis in linear utility functions~\cite{tran2015composite}.
Second, gradient-based analyses that rely on parameter-independent gradients in linear models do not extend naturally to the non-linear setting, as gradients become parameter-dependent and more intricate.
Lastly, non-linear utility functions typically induce a non-convex optimization landscape characterized by multiple global minima due to symmetries (e.g., neuron permutations in neural networks), complicating both the optimization process and subsequent regret analysis.

Despite these challenges, this paper presents a computationally tractable algorithm for MNL contextual bandits that can 
handle non-linear utility functions and achieve a sharp regret bound. Our key contributions are as follows:

\begin{itemize}
    \item
    We propose an \textit{upper confidence bound} (UCB)-based algorithm designed explicitly for MNL contextual bandits with non-linear utilities.
    Under a realizability assumption and a newly proposed generalized geometric condition, our algorithm achieves a regret bound of $\tilde{\Ocal}(\sqrt{T})$.
    To our knowledge, this constitutes the first algorithm with such near-optimal regret guarantees independent of the total number of items $N$ in this non-linear utility setting.
    
    \item
    Unlike the existing approach~\cite{zhang2024contextual}, which either compromise statistical efficiency or computational feasibility, our algorithm simultaneously attains computational tractability and the sharp $\tilde{\Ocal}(\sqrt{T})$ regret bound under non-linear utility settings.
    
    \item
    To overcome technical challenges arising in parameter learning for non-linear utility based MNL models, we introduce a novel generalized geometric condition on the squared error of the utility function (Assumption~\ref{assm:generalized geometric condition}).
    This condition significantly generalizes and weakens assumptions used in previous studies~\cite{liu2023global, li2024communicationefficient, siam2025quantum} and accommodates broad classes of non-linear functions, including neural networks. 
    Moreover, we derive a new concentration inequality for parameter estimation in MNL models, independent of the regularization parameter up to logarithmic factors. 
    These technical advancements allow us to construct computationally tractable, optimistic utility estimators suitable for efficient exploration.
    
    Crucially, our approach circumvents the commonly used \textit{neural tangent kernel} (NTK) assumptions~\cite{jacot2018ntk}, eliminating the impractical requirement of extensive over-parameterization prevalent in NTK-based analyses.
    
    \item
    We demonstrate the superior performance of our algorithm through extensive numerical experiments.
    While benchmark methods suffer significant performance degradation in non-linear utility scenarios, our proposed method maintains robust performance in both realizable and misspecified contexts, underscoring its practical effectiveness and adaptability.
\end{itemize}

\begin{table*}[t!]
    \begin{center}
    \caption{
    Comparison of results and main assumptions for prior MNL contextual bandits with non-linear utilities.
    ~\citet{zhang2024contextual} consider a bounded and Lipschitz-continuous utility class, whereas our work assumes a bounded, Lipschitz, and smooth utility class.
    The geometric condition (Assumption~\ref{assm:generalized geometric condition}) is imposed in Phase I to establish the convergence rate of the pilot estimator.
    In contrast, the analysis in~\cite{zhang2024contextual} relies on the generalization error of the (offline or online) regression oracle, measured in terms of log-loss, rather than the distance between the estimator and the true parameter.
    The dimensionality $d_w$ appearing in the regret bounds of~\cite{zhang2024contextual} corresponds to the dimension of the Lipschitz function parameter, which may differ from the context dimension $d$.
    }
    \label{tab:my-table}
    \resizebox{\columnwidth}{!}{%
    \begin{tabular}{ccccc}
        \hline
        Algorithm                               & Utility                & Assumptions & Tractability\textsuperscript{$\dag$} & Regret
        \\
        \hline
        $\varepsilon$-greedy~\cite{zhang2024contextual} & Lipschitzness & Offline reg. oracle & \ding{51} & $\tilde{\Ocal}((d_w N K)^{1/3} T^{2/3})$
        \\
        log-barrier regularizer~\cite{zhang2024contextual} & Lipschitzness & Offline reg. oracle & \ding{55} & $\tilde{\Ocal}(K^2 \sqrt{d_w N T})$
        \\        
        $\varepsilon$-greedy~\cite{zhang2024contextual} & Lipschitzness & Online reg. oracle & \ding{51} & $\tilde{\Ocal}((NK)^{1/3} T^{5/6})$
        \\
        \texttt{Feel-Good TS}~\cite{zhang2024contextual} & Lipschitzness & Online reg. oracle & \ding{55} & $\tilde{\Ocal}(K^2 \sqrt{d_w N T})$
        \\
        $\algname$ (\textbf{this work}) & Smoothness & Geometric condition & \ding{51} & $\tilde{\Ocal}(\kappa^{-2} \mu^{-1} d_w \sqrt{T})$
        \\
        \hline        
    \end{tabular}%
    }
    \end{center}
    \vspace{0.5pt}
    {\footnotesize
    \textsuperscript{$\dag$}%
    Tractability of each method refers to the computational feasibility of the exploration strategy, i.e., the process of selecting an assortment given a current utility estimator. The computation required to obtain our pilot estimator is comparable to what is already used in~\cite{zhang2024contextual}.
    Following~\cite{zhang2024contextual}, any tractable empirical risk minimization method (e.g., SGD) is sufficient for computing our pilot estimator in practice.
    }    
\end{table*}

\section{Preliminary} \label{sec:preliminary}
\paragraph{Notations.}
For a positive integer $N \in \mathbb{N}$, we denote the set $\{1, \ldots, N\}$ by $[N]$.
For a vector $\xb \in \mathbb{R}^d$, its $\ell_p$-norm is defined as $\| \xb \|_p = ( \sum_{j=1}^d |x_j|^p )^{1/p}$ for $1 \le p < \infty$, and its $\ell_\infty$-norm is defined as $\| \xb \|_\infty = \max_{j \in [d]} |x_j|$.
For a matrix $\Ab$, its operator norm is denoted by $\| \Ab \|_{\mathrm{op}}$.
For a vector $\xb \in \mathbb{R}^d$ and a symmetric positive semi-definite matrix $\Ab \in \mathbb{R}^{d \times d}$, we define $\| \xb \|_{\Ab} := \sqrt{\xb^\top \Ab \xb}$.
We denote a real-valued function defined on $\Xcal$ and parameterized by $\wb$ as $f_{\wb}:\Xcal \rightarrow \RR$, and denote its gradient and Hessian with respect to $\wb$ by $\nabla f_{\wb}$ and $\nabla^2 f_{\wb}$, respectively.
Throughout, we use standard big-$\mathcal{O}$ notation to hide universal constants, and $\tilde{\mathcal{O}}$ notation to additionally hide logarithmic and poly-logarithmic factors.

\subsection{Problem Setting}

\paragraph{MNL contextual bandits.} 
We consider a sequential assortment selection problem in which, at each round $t$, the agent observes a set of feature vectors for all items, denoted by $X_t := \{ \xb_{t1}, \ldots, \xb_{tN} \} \subset \mathbb{R}^d$, where each feature vector belongs to a general context space $\mathcal{X} \subset \mathbb{R}^d$.
Based on this information, the agent selects an assortment $S_t = \{i_1, \dots, i_l\} \in \mathcal{S}:= \{ S \subset [N]: |S| \le K \}$, where $l \leq K$, and observes a user choice $i_t \in S_t \cup \{0\}$, where $0$ denotes the ``outside option'', i.e., the user chooses none of the offered items.
The user's selection $i_t \in S_t \cup \{0\}$ is modeled by the MNL choice model:
\begin{equation} \label{eq:true mnl}
    \begin{split}
        & \PP( i_t = i \mid X_t, S_t) =: p(i \mid X_t, S_t, \wb^*) := \frac{\exp(f_{\wb^*}(\xb_{ti}))}{1 + \sum_{j \in S_t} \exp(f_{\wb^*}(\xb_{tj}))}, 
        \\
        & \PP( i_t = 0 \mid X_t, S_t) =: p(0 \mid X_t, S_t, \wb^*) := \frac{1}{1 + \sum_{j \in S_t} \exp(f_{\wb^*}(\xb_{tj}))},     
    \end{split}
\end{equation}    
where $f_{\wb^*}(\cdot ): \Xcal \rightarrow \RR$ is an \textit{unknown} utility function, specifying the user's value for each item under the given context. 
At each round \( t \), the agent receives feedback in the form of a choice response indicating which item from the offered set \( S_t \cup \{0\} \) was selected.
The choice outcome is represented as a one-hot vector \( \yb_t := (y_{t0}, y_{t1}, \ldots, y_{tl}) \), where \( y_{ti} = 1 \) if item \( i \) was chosen and 0 otherwise.
By definition of the model, this choice is drawn according to the following MNL distribution :
\begin{equation*}
    \yb_t \sim \mathrm{Multinomial} \left\{1, \left[p(0 \mid X_t, S_t, \wb^*), p(i_1 \mid X_t, S_t, \wb^*),  \ldots, p(i_l \mid X_t, S_t, \wb^*) \right] \right\} \, ,
\end{equation*}
where $1$ denotes $\yb_t$ is a single-trial sample, i.e., $y_{t0} + \sum_{k=1}^l y_{tk} = 1$.

For a given revenue parameter vector \(\rb_t = [r_{t0}, r_{t1}, \ldots, r_{tN}]\), where \(r_{t0} = 0\) denotes the revenue associated with the outside option and \(r_{ti}\) is the revenue for item \(i \in [N]\),  
the expected reward of offering an assortment \(S \in \Scal\) under the context feature set \(X_t\) is defined as  
\begin{equation*}
    R_{X_t, \rb_t}(S, \wb^*) =: R_t(S, \wb^*) = \sum_{i \in S} p(i \mid X_t, S, \wb^*) r_{ti} = \sum_{i \in S} \frac{\exp(f_{\wb^*}(\xb_{ti})) r_{ti}}{1 + \sum_{j \in S} \exp(f_{\wb^*}(\xb_{tj}))} \, .
\end{equation*}

The goal of the agent is to minimize the cumulative regret over $T$ rounds, defined as the total difference in expected reward between the offline optimal assortment $S_t^* := \argmax_{S \in \Scal} R_t(S, \wb^*)$ and the assortment chosen by the agent,
$\displaystyle \mathrm{Regret}_T := \sum_{t=1}^T \left[ R_t(S_t^*, \wb^*) - R_t(S_t, \wb^*) \right]$.

\paragraph{Non-linear parametric functions.}

We consider a parametric function class \( \Fcal := \{ f_\wb: \Xcal \rightarrow \RR \mid \wb \in \Wcal \} \), where \( \Wcal \subset \RR^{d_w} \) is a set of \( d_w \)-dimensional parameter vectors.  
Each function \( f_\wb \in \Fcal \) may be highly non-convex and is not necessarily differentiable with respect to the context vector \( \xb \in \Xcal \), but we assume it is differentiable with respect to the parameter \( \wb \).  
We further assume that the function class \( \Fcal \) is expressive enough to contain the true (\textit{unknown}) utility function \( f_{\wb^*} \).

\begin{assumption}[Realizability] \label{assm:realizability}
    We assume $\wb^* \in \Wcal$ and $\Fcal$ is given to the agent.
\end{assumption}

Assumption~\ref{assm:realizability} is standard in the contextual bandit literature using general function approximation~\cite{foster2018practical, foster2020beyond, simchi2022bypassing, liu2023global, zhang2024contextual}. 
While this assumption is necessary for our theoretical guarantees, we will also demonstrate empirically that our proposed algorithm performs well even under model misspecification, particularly when the function class under consideration has high representational capacity, such as neural networks.

We also define the equivalence set $\Wcal^* \subset \Wcal$ as the set of parameter vectors that yield the same function values as \( f_{\wb^*} \) for all \( \xb \in \Xcal \), i.e.,  $\Wcal^* := \{ \wb \in \Wcal : f_\wb(\xb) = f_{\wb^*}(\xb) \text{ for all } \xb \in \Xcal \}$.
This equivalence set arises naturally in expressive function classes such as neural networks.  
For instance, consider a two-layer neural network defined as $f_{\wb^*}(\xb) = \sum_{k=1}^m \sigma(\xb^\top \wb^*_k)$, where $\wb^* = [\wb_1^*, \ldots, \wb_m^*] \in \RR^{d \times m}$ is the weight matrix, and $\sigma$ denotes the activation function.
Any column permutation of $\wb^*$ yields the same function output as $f_{\wb^*}$.
Therefore, rather than recovering $\wb^*$ exactly, it suffices to identify any parameter within the equivalence set $\Wcal^*$.

Following prior work on contextual bandits~\cite{abbasi2011improved, oh2019thompson, faury2020improved, lee2024nearly, liu2023global, li2024communicationefficient, siam2025quantum}, we make the following boundedness assumption.
\begin{assumption}[Boundedness] \label{assm:boundedness}
    For all $t \ge 1$ and $i \in [N]$, we assume that $\| \xb_{ti} \|_2 \le 1$ and $r_{ti} \in [0,1]$, and $\| \wb \|_2 \le 1$ for all $\wb \in \Wcal$. 
    Also, we assume that there exist constants $C_f, C_g, C_h > 0$ such that for all $\xb \in \Xcal$ and $\wb \in \Wcal$, it holds that
    \begin{equation*}
        |f_{\wb}(\xb)| \le C_f, 
        \quad \| \nabla f_{\wb}(\xb) \|_2 \le C_g, 
        \quad \| \nabla^2 f_{\wb}(\xb) \|_{\text{op}} \le C_h \, ,
    \end{equation*}
    where $\nabla f_{\wb} (\xb)$ and $\nabla^2 f_{\wb}(\xb)$ denote the gradient and Hessian of $f_{\wb}$ with respect to $\wb$.
\end{assumption}

\subsection{Challenges of Learning in Non-Linear Utility}

Learning utility parameters in non-linear MNL bandit models presents significant challenges compared to linear models. In linear utility settings, the negative log-likelihood function exhibits a self-concordant-like property~\cite{tran2015composite}, simplifying theoretical analysis and enabling tight regret bounds~\cite{perivier2022dynamic, lee2024nearly}. Non-linear utility models, however, typically lack this property, complicating analysis significantly.
Another key difference lies in the gradient structure. In linear models, gradients are simply feature vectors, independent of parameters, allowing straightforward concentration inequalities. In contrast, non-linear gradients depend explicitly on unknown parameters, complicating the application of these analytical methods.
Moreover, the optimization landscape for non-linear utilities is notably more complex. Linear models yield convex loss functions with unique minima, whereas non-linear models (e.g., neural networks) produce highly non-convex loss surfaces. Additionally, symmetries like neuron permutations cause multiple global minima, further complicating optimization and theoretical analysis in non-linear settings.

\section{Main Results} \label{sec:main results}
In this section, we introduce an algorithm for the MNL bandit problem with a non-linear parametric utility function that is both computationally tractable and statistically efficient.
We begin by explaining the key design principles of the algorithm, and then present its regret bound along with the main technical components of the analysis.

\subsection{\texorpdfstring{Algorithm: $\algname$}{Algorithm: ONL-MNL}}
\begin{algorithm}[t!] 
    \caption{$\algname$ (\textbf{O}ptimistic \textbf{N}on-\textbf{L}inear Utility for Contextual \textbf{MNL} Bandit)} \label{alg:algorithm 1}
    \begin{algorithmic}[1]
        \State \textbf{Inputs:} Regularization parameter $\lambda$, confidence radius $\{\beta_t\}_{t \ge 1}$, exploration length $t_0$, number of rounds $T$
        \For{$t=1,2, \ldots, t_0$} \Comment{\textit{Phase I: Uniform exploration}}
            \State Observe $X_t = \{ \xb_{t1}, \ldots, \xb_{tN} \}$, offer $S_t \sim \unif(\Scal)$ and observe $i_t$
        \EndFor
        \State Compute $\hat{\wb}_0 = \argmin_{\wb \in \Wcal} \Lcal (\wb)$
        \State \textbf{Initialization:} $\Vb_{t_0} = \lambda \Ib_{d_w}$
        \For{$t=t_0+1,\ldots, T$} \Comment{\textit{Phase II: Optimistic exploration}}
            \State Observe $X_t= \{ \xb_{t1}, \ldots, \xb_{tN} \}$
            \State Compute $\hat{\wb}_t = \argmin_{\wb \in \Wcal} \ell_t(\wb)$
            \State Compute $z_{ti} = f_{\hat{\wb}_t}(\xb_{ti}) + \sqrt{\beta_t} \|\nabla f_{\hat{\wb}_t} (\xb_{ti})\|_{\Vb_{t}^{-1}} + \frac{\beta_t C_h}{\lambda}$ for all $i \in [N]$
            \State Offer $S_t \in \argmax_{S \in \Scal} \tilde{R}_t(S)$ and observe $i_t$
            \State Update $\Vb_{t+1} = \Vb_t + \sum_{i \in S_t} \nabla f_{\hat{\wb}_t} (\xb_{ti}) \nabla f_{\hat{\wb}_t} (\xb_{ti})^\top$
        \EndFor
    \end{algorithmic}
\end{algorithm}

We propose $\algname$, an algorithm that leverages optimistic estimates derived from non-linear utility functions to guide exploration. The complete procedure is presented in Algorithm~\ref{alg:algorithm 1}, which consists of two main stages.

\paragraph{Uniform exploration phase.}
The first stage (Phase I) is a uniform exploration phase that lasts for \( t_0 \) rounds.
During this phase, the agent selects an assortment of size at most \( K \), uniformly at random. 
The goal of this exploration phase is to obtain a ``pilot'' estimate of the unknown parameter \( \wb^* \) or any parameter in the equivalence set $\Wcal^*$—that is, a good initial estimator.
At the end of Phase I, the agent invokes a regression oracle to estimate the initial parameter $\hat{\wb}_0$ based on the data collected during the exploration rounds.
This is done by minimizing the negative log-likelihood of the observed item choices under the MNL model in Eq.~\eqref{eq:true mnl} defined as follows:
\begin{equation*}
    \hat{\wb}_0 = \argmin_{\wb \in \Wcal} \Lcal (\wb) := - \sum_{t=1}^{t_0} \log p(i_t \mid X_t, S_t, \wb) \,.
\end{equation*}

Here, $\Lcal (\wb)$ denotes the empirical loss over the first $t_0$ rounds, where $i_t$ is the item chosen from assortment $S_t$ with the context set $X_t$.
Using standard results from empirical risk minimization over Lipschitz function classes, together with the reverse Lipschitz property of the MNL model, we show that the minimum squared error of the pilot estimator $\hat{\wb}_0$ with respect to the equivalence set $\Wcal^*$, namely $\min_{\tilde{\wb} \in \Wcal^*} \| \hat{\wb}_0 - \tilde{\wb} \|_2^2$, converges at a rate of $\Ocal(1/t_0)$ (Lemma~\ref{lemma:convergence of w_0}).

\paragraph{Optimistic exploration phase.}
In the second stage (Phase II), we adopt a more sophisticated strategy to balance exploration and exploitation.
In particular, we construct a confidence region for the parameter in the equivalence set $\Wcal^*$ that is closest to the pilot estimator $\hat{\wb}_0$.
We then estimate the utility parameters within $\Wcal^*$ by minimizing the following regularized negative log-likelihood: 
\begin{equation} \label{eq:linearized mnl loss}
    \hat{\wb}_t = \argmin_{\wb \in \Wcal} \ell_t(\wb) := - \sum_{s=t_0 + 1}^{t-1} \sum_{i \in S_s} y_{si} \log \hat{p}(i \mid X_s, S_s, \wb) + \frac{\lambda}{2} \| \wb - \hat{\wb}_0 \|_2^2 \, ,
\end{equation}
where $\hat{p}(i \mid X_s, S_s, \wb)$ is a \textit{linearized} MNL model given by
\begin{equation*}
        \hat{p}(i \mid X_s, S_s, \wb) := \frac{\exp(f_{\hat{\wb}_s} (\xb_{si}) + \nabla f_{\hat{\wb}_s}(\xb_{si})^\top (\wb - \hat{\wb}_s))}{1 + \sum_{j \in S_s} \exp(f_{\hat{\wb}_s}(\xb_{sj}) + \nabla f_{\hat{\wb}_s}(\xb_{sj})^\top (\wb - \hat{\wb}_s))} \, .
\end{equation*}
The key intuition is that we apply a first-order Taylor approximation of the non-linear utility function \( f_{\wb}(\cdot) \) around the estimate \( \hat{\wb}_s \), since we are dealing with a general non-linear model.  
Moreover, instead of regularizing around the origin \( \zero_{d_w} \), we regularize around the pilot estimator \( \hat{\wb}_0 \), as it provides a meaningful reference point that is already close to the parameters in $\Wcal^*$.
Based on the estimated parameter $\hat{\wb}_t$, we define the confidence set $\Ccal_t$ as follows:
\begin{equation*}
    \Ccal_t := \{ \wb \in \Wcal : \| \wb - \hat{\wb}_t \|_{\Vb_t}^2 \le \beta_t \} \, ,
\end{equation*}
where $\beta_t$ is a pre-defined monotonically increasing sequence specified later, and $\Vb_t:= \lambda \Ib_{d_w} + \sum_{s=t_0+1}^{t-1} \sum_{i \in S_s} \nabla f_{\hat{\wb}_s}(\xb_{si}) \nabla f_{\hat{\wb}_s}(\xb_{si})^\top$ is the Gram matrix. 
With suitably set $\beta_t$, we will show that with high probability, $\argmin_{\tilde{\wb} \in \Wcal^*} \| \hat{\wb}_0 - \tilde{\wb} \|_2 \in \Ccal_t$, i.e., the equivalent utility parameter in $\Wcal^*$ that is cloest to the pilot estimator $\hat{\wb}_0$ lies  within the constructed confidence set $\Ccal_t$.
Based on this observation, we compute the optimistic utility estimate $z_{ti}$ as follows:
\begin{equation*}
    z_{ti} = f_{ \hat{\wb}_{t}}(\xb_{ti}) + \sqrt{\beta_t} \|\nabla f_{ \hat{\wb}_{t}}(\xb_{ti})\|_{\Vb_{t}^{-1}} + \frac{\beta_t C_h}{\lambda} \, , \quad \forall i \in [N] \, .
\end{equation*}
The optimistic utility $z_{ti}$ is composed of the parts: the mean utility estimate $f_{ \hat{\wb}_{t}}(\xb_{ti})$ and the uncertainty estimate $\sqrt{\beta_t} \|\nabla f_{ \hat{\wb}_{t}}(\xb_{ti})\|_{\Vb_{t}^{-1}} + \frac{\beta_t C_h}{\lambda}$. 
We will show that $z_{ti}$ is an optimistic estimate of the true utility $f_{\wb^*}(\xb_{ti})$, assuming 
$\argmin_{\tilde{\wb} \in \Wcal^*} \| \hat{\wb}_0 - \tilde{\wb} \|_2 \in \Ccal_t$.
Based on $z_{ti}$, we construct the optimistic expected reward for the assortment, defined as $\displaystyle \tilde{R}_t (S) := \sum_{i \in S} \frac{\exp(z_{ti}) r_{ti}}{1 + \sum_{j \in S} \exp(z_{tj})}$.
Then, we offer the set $S_t$ that maximizes the optimistic expected reward, i.e., $S_t = \argmax \tilde{R}_t(S)$. 

\subsection{Regret Bound}
In this section, we present the cumulative regret upper bound for the proposed algorithm.
First, we introduce technical assumptions used to derive the regret bound. 
\begin{assumption}[Stochastic context in Phase I] \label{assm:stochastic context}
    We assume that during Phase I, the context feature vectors $X_t := \{ \xb_{t1}, \ldots, \xb_{tN} \}$ are independently and identically distributed (\textit{i.i.d.}) samples drawn from an unknown distribution $\Dcal$ supported on $\Xcal$.
\end{assumption}

\begin{assumption}[Generalized geometric condition on the squared loss]~\label{assm:generalized geometric condition}
    Let
    \begin{equation*}
        \ell_{\mathrm{sq}}(\wb) = \EE_{X \sim \Dcal, S \sim \unif(\Scal)} \left[ \sum_{i \in S} (f_{\wb} (\xb_j) - f_{\wb^*} (\xb_j))^2 \right]
    \end{equation*}
    be the expected squared loss function over both the context feature distribution $\Dcal$ and a uniformly sampled assortment.
    For the equivalence set $\Wcal^*$, we assume that $\ell_{\mathrm{sq}}(\wb)$ satisfies either a $(\tau, \gamma)$-growth condition or $\mu$-local strong convexity with respect to $\Wcal^*$, i.e., $\forall \wb \in \Wcal \setminus \Wcal^*$,
    \begin{equation*}
        \min_{\tilde{\wb} \in \Wcal^*} \left\{ \frac{\mu}{2} \| \wb - \tilde{\wb} \|_2^2 + \ell_{\mathrm{sq}}(\tilde{\wb}) , \frac{\tau}{2} \| \wb - \tilde{\wb} \|_2^\gamma + \ell_{\mathrm{sq}}(\tilde{\wb}) \right\} \le \ell_{\mathrm{sq}}(\wb) \, ,        
    \end{equation*}    
    for constants $\mu > 0$ and $0 < \gamma < 2$.
\end{assumption}

\begin{figure}[t]
  \centering
  \begin{minipage}[c]{0.48\textwidth}
    \centering
    \includegraphics[width=1\linewidth]{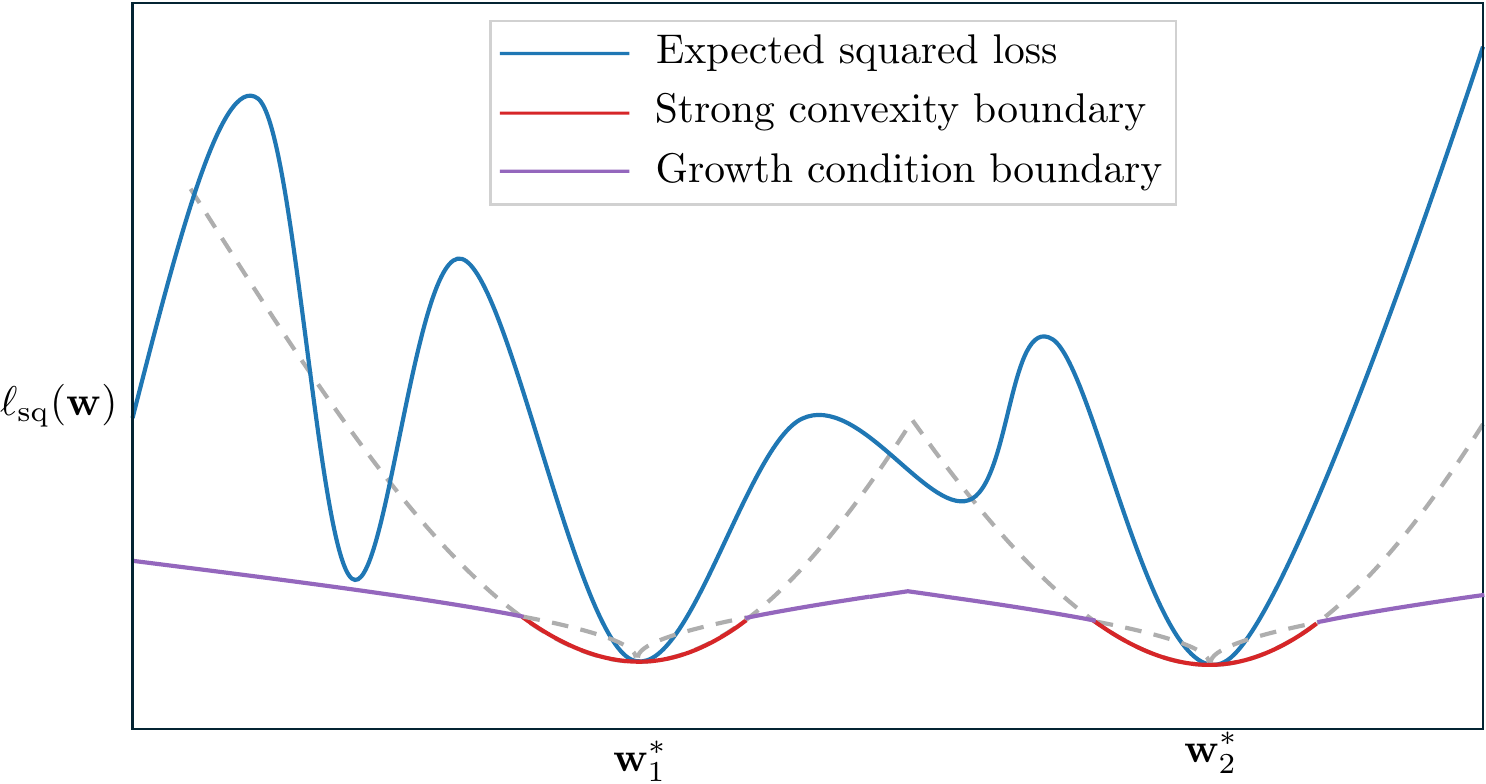}
  \end{minipage}
  \hfill
  \begin{minipage}[c]{0.48\textwidth}
    \captionsetup{justification=justified}
    \captionof{figure}{%
    \small Example of a highly non-convex loss $\ell_{\mathrm{sq}}(\wb)$ that satisfies Assumption~\ref{assm:generalized geometric condition}. The solid line shows the lower bound given by the pointwise minimum of the strong convexity and growth conditions over the equivalence set $\Wcal^*$. For neural networks, the squared loss may have multiple global minima due to parameter symmetries or over-parameterization. This highlights that Assumption~\ref{assm:generalized geometric condition} is strictly weaker than prior geometric conditions, enabling it to cover a wider range of utility functions.            
    }
    \label{fig:geometric condition}
  \end{minipage}
  \vspace{-0.5cm}
\end{figure}

\paragraph{Discussion of assumptions.}
Assumption~\ref{assm:stochastic context} is used to bound the difference between the expected negative log-likelihoods of the true utility parameter $\wb^*$ and the pilot estimator $\hat{\wb}_0$, a type of assumption commonly used in the empirical risk minimization literature~\cite{van2015fast, zhang2024contextual}.
Note that Assumption~\ref{assm:stochastic context} is only required for Phase I; after Phase I, the context vectors may be chosen even adversarially.

Assumption~\ref{assm:generalized geometric condition} is on the expected squared loss between utility functions over the context distribution, rather than the unknown true utility function $f_{\wb^*}$ itself.
Assumption~\ref{assm:generalized geometric condition} is also required in Phase I to establish the convergence rate of the pilot estimator $\hat{\wb}_0$.
The growth condition assumes that any utility function $f_\wb$, whose parameters lie far from the equivalence set $\Wcal^*$, cannot approximate the true utility function well over the context distribution.
The local strong convexity condition assumes that the expected loss exhibits quadratic growth within a local neighborhood around the equivalence set $\Wcal^*$.
Notably, this condition is strictly weaker than the global strong convexity assumptions commonly adopted in prior work on parametric bandits—such as linear and generalized linear bandits~\cite{filippi2010parametric, abbasi2011improved}—since it does not require the squared loss to be convex outside the local region near the optimal parameters.
Moreover, Assumption~\ref{assm:generalized geometric condition} is more general than the geometric conditions employed in prior works~\cite{liu2023global, li2024communicationefficient, siam2025quantum}.
That is, any function satisfying the geometric conditions in~\cite{liu2023global, li2024communicationefficient, siam2025quantum} also satisfies ours, but the converse does not necessarily hold.
Previous geometric assumptions typically require the existence of a unique global minimum, whereas we explicitly allow for multiple equivalent minima.
As previously discussed, this is a more realistic assumption in settings such as neural networks, where the squared error loss can admit multiple global minima—e.g., due to parameter symmetries or overparameterization.
This flexibility allows our framework to accommodate highly non-convex and non-linear expected loss landscapes, which may contain arbitrarily many spurious local minima and multiple global minima (Figure~\ref{fig:geometric condition}).

Now we are ready to state our main result that upper bounds the cumulative regret of the Algorithm~\ref{alg:algorithm 1}.\\

\begin{theorem}[Regret Bound of $\algname$] \label{thm:regret bound}
    Suppose Assumptions~\ref{assm:realizability},~\ref{assm:boundedness},~\ref{assm:stochastic context} and~\ref{assm:generalized geometric condition} hold.
    For any $\delta \in (0,1)$, if we set the algorithmic parameters in Algorithm~\ref{alg:algorithm 1} as follows: $T \ge \tilde{C} \kappa^{-1} C_f^2 \zeta^2 \left( \frac{\mu^{\gamma/(2-\gamma)}}{\tau^{2/(2-\gamma)}} \right)^2$, $t_0 = \lceil \kappa^{-3/2} d_w \sqrt{T} \rceil$, $\lambda = \tilde{\Ocal}(\kappa^{-5/2} \mu^{-1} d_w \sqrt{T})$, $\beta_t = \tilde{\Ocal} \left(  \mu^{-2} \kappa^{-4} d_w \right)$, 
    where $\tilde{C}$ is a universal constant, $\kappa:= \min_{\wb \in \Wcal, X \subset \Xcal, S \in \Scal, i \in S} p(0 \mid X, S, \wb) p(i \mid X, S, \wb)$, and, $\zeta$ is the logarithmic term depending on $K, t_0, C_g, 1/\delta$, then with probability at least $1 - 2\delta$, Algorithm~\ref{alg:algorithm 1} achieves the following cumulative regret:
    \begin{equation*}
        \mathrm{Regret}_{T} = \tilde{ \Ocal } \left( \kappa^{-3/2} d_w \sqrt{T} + \kappa^{-2} \mu^{-1}  d_w  \sqrt{T}  \right) \, .          
    \end{equation*}
\end{theorem}

\paragraph{Discussion of Theorem~\ref{thm:regret bound}.}
In terms of the total time step $T$, Algorithm~\ref{alg:algorithm 1} achieves $\tilde{\Ocal}(\sqrt{T})$ cumulative regret.
This implies that Algorithm~\ref{alg:algorithm 1} is a no-regret algorithm, as $\lim_{T \rightarrow \infty} \mathrm{Regret}_{T}/T = 0$.

When the utility function family consists of neural networks, our proposed method still achieves regret bound of $\tilde{\Ocal}(\sqrt{T})$ without relying on the NTK assumption, which is commonly used in the neural network-based bandit analysis~\citep{zhou2020neural, zhang2021neural, hwang2023combinatorial}, and thus eliminates the impractical requirement of extensive over-parameterization prevalent in NTK-based analyses.
In particular, existing neural bandits~\cite{xu2021robust, huang2021going, xu2024stochastic} that do not rely on NTK assumptions still require restrictive conditions—such as specific network architectures (e.g., two-layered ReLU networks with fixed second-layer weights set to one~\cite{xu2024stochastic}, or networks with quadratic activation functions~\cite{xu2021robust}) or specific context distributions (e.g., Gaussian~\cite{huang2021going}, or uniform distribution over the unit sphere~\cite{xu2024stochastic}).
In contrast, our approach applies to a broader class of neural networks that satisfy Assumption~\ref{assm:generalized geometric condition}, without imposing such architectural or distributional constraints.

Compared to the existing contextual MNL bandit algorithms with general utility functions~\cite{zhang2024contextual}, our algorithm is both computationally tractable and statistically efficient.
While the uniform exploration method in~\citet{zhang2024contextual} is computationally tractable,
it results in sub-optimal regret—for example, $\tilde{\Ocal}(T^{2/3})$ in the stochastic setting and \( \tilde{\Ocal}(T^{5/6}) \) in the adversarial setting.
Their other methods—log-barrier regularization-based exploration and the Feel-Good Thompson Sampling~\cite{zhang2022feel} approach—are computationally intractable, meaning they cannot be solved in polynomial time.
In contrast, our algorithm operates within polynomial time.
The main computational cost lies in minimizing the regularized negative log-likelihood, which can be efficiently approximated using gradient-based methods such as gradient descent with a per-round complexity of $\mathcal{O}(t)$~\cite{faury2022jointly}.
The assortment selection step is also tractable: by combining the optimistic utility estimates $z_{ti}$ with efficient assortment optimization techniques~\cite{rusmevichientong2010dynamic, davis2014assortment}, as done in~\citet{oh2019thompson, oh2021multinomial}, it can be solved with $\mathcal{O}(N)$ computational cost.    
In terms of regret bounds, although our regret depends on a problem-dependent instance factor $\kappa$, it does not scale directly with the total number of items~$N$ or directly with the assortment size $K$.
In contrast, the computationally intractable algorithm with $\tilde{\Ocal}(\sqrt{T})$-regret by~\citet{zhang2024contextual} has super-linear dependence on $K$—specifically $\mathcal{O}(K^2)$—, and additionally depend polynomially on $N$, limiting their applicability in settings with a large set of items, hence restricting real-world applications.
To the best of our knowledge, \textit{our method is the only known algorithm that is both computationally tractable and provably achieves a $\tilde{\Ocal}(\sqrt{T})$ regret in the contextual MNL bandit with non-linear utility functions.}

\begin{remark}
    By definition, $\kappa^{-1}$ can scale as $\Ocal(K^2)$ in the worst-case scenario. 
    However, in practice, the behavior of $\kappa$ is often much more favorable.
    For example, as shown in the experiments of~\citet{lee2024nearly} (Figure 1 in~\cite{lee2024nearly}), the gap between the regret of $\kappa^{-1}$-dependent algorithms~\cite{oh2019thompson, oh2021multinomial} and that of $\kappa^{-1}$-improved algorithm~\cite{lee2024nearly} does not grow significantly with $K$, even as $K$ increases.
    This observation suggests that equating $\kappa^{-1}$ with $K^2$ or treating it as equivalent to $K$ is overly pessimistic.
    Also, we note that the exponential dependence in $\kappa$ can be adjusted by rescaling the utility values. 
    For instance,~\citet{zhang2024contextual} define the MNL probability model using bounded utilities without applying the exponential transformation.
\end{remark}

\subsection{Regret Analysis} \label{sec:regret analysis}
In this section, we present key technical ingredients that enable the regret bound of Algorithm~\ref{alg:algorithm 1}, as established in Theorem~\ref{thm:regret bound}.
All the detailed proofs are provided in the appendix.

\paragraph{Convergence of $\hat{\wb}_0$.}
The purpose of Phase I in Algorithm~\ref{alg:algorithm 1} is to explore the unknown utility function $f_{\wb^*}$ through \( t_0 \) rounds of uniform sampling and to obtain a pilot estimate \( \hat{\wb}_0 \) that is sufficiently accurate. 
We establish a convergence guarantee for \( \hat{\wb}_0 \) relative to the equivalence set \( \Wcal^* \); specifically, we bound the quantity \( \min_{\tilde{\wb} \in \Wcal^*} \| \hat{\wb}_0 - \tilde{\wb} \|_2^2 \).

\begin{lemma}[Convergence rate of $\hat{\wb}_0$] \label{lemma:convergence of w_0}
    Suppose Assumptions~\ref{assm:realizability},~\ref{assm:boundedness},~\ref{assm:stochastic context} and~\ref{assm:generalized geometric condition} hold.
    There exist an absolute constant $C > 0$ such that after $t_0$ rounds in Phase I of Algorithm~\ref{alg:algorithm 1} where $t_0$ satisfies $t_0 \ge 2^{-1} C \kappa^{-2} d_w C_f^2 \zeta \frac{\mu^{\gamma/(2-\gamma)}}{\tau^{2/(2-\gamma)}}$, 
    with probability at least $1 - \delta$, we have
    \begin{equation*}
        \min_{\tilde{\wb} \in \Wcal^*} \| \hat{\wb}_0 - \tilde{\wb} \|_2^2 \le C \frac{\kappa^{-2} d_w C_f^2 \zeta}{\mu t_0} \, ,
    \end{equation*}
    where $\kappa:= \min_{\wb \in \Wcal, X \subset \Xcal, S \in \Scal, i \in S} p(0 \mid X, S, \wb) p(i \mid X, S, \wb)$ and $\zeta$ is the logarithmic term depending on $K, t_0, C_g, 1/\delta$.   
\end{lemma}

Compared to the contextual multi-armed bandit setting with general reward function~\cite{foster2020beyond,simchi2022bypassing}, one of the key challenges in estimating the utility parameter in the MNL bandit is that the agent does not receive direct feedback on the utility of each item. 
Instead, it only observes the user's choice from the offered assortment.
Despite this challenge, in Lemma~\ref{lemma:convergence of w_0}, we first adapt existing generalization error bounds of the empirical risk minimizer under \textit{i.i.d.} sampling, expressed in terms of the log-loss~\cite{van2015fast,zhang2024contextual}.
Then, leveraging the reverse Lipschitz property of the MNL model (Lemma~\ref{sub_lemma:reverse Lipschitzness of the MNL model}), we lower bound the difference in log-likelihoods by the squared error between the utility values.
Unlike~\citet{zhang2024contextual}, which derives a lower bound based on the assortment size $K$, our analysis incorporates a problem-dependent instance factor $\kappa$. 

\paragraph{Confidence set \& Optimism.}
Although the $1/t_0$ convergence rate of $\hat{\wb}_0$ is quite tight, it is not sufficient on its own.
For example,~\citet{xu2021robust} show that in the contextual bandit setting, one can achieve a regret of $\tilde{\Ocal}(T^{2/3})$ by first learning the parameters of a quadratic neural network through $t_0$ rounds of uniform exploration (yielding a $1/t_0$ convergence rate), and then switching entirely to exploitation using the learned parameters.
In contrast, we go beyond this by constructing a confidence set for parameters in equivalence set $\Wcal^*$.
We then use an optimistic estimate of the true utility within this set to guide assortment selection, ultimately achieving a regret of $\tilde{O}(\sqrt{T})$. The following lemma explains how this confidence set is constructed.  

\begin{lemma}[Confidence set] \label{lemma:confidence ball}
    Suppose Assumptions~\ref{assm:realizability},~\ref{assm:boundedness},~\ref{assm:stochastic context} and~\ref{assm:generalized geometric condition} hold.
    If we choose $t_0 = \lceil d_w \kappa^{-3/2} \sqrt{T} \rceil$ and $\lambda = \tilde{\Ocal}(\kappa^{-3/2} \mu^{-1} d_w \sqrt{T})$, then $\forall t \ge t_0 + 1$ in Phase II of Algorithm~\ref{alg:algorithm 1}, with probability at least $1 - 2\delta$, the following holds:
    \begin{equation*}
        \min_{\tilde{\wb} \in \Wcal^*} \| \hat{\wb}_t - \tilde{\wb} \|_{\Vb_t}^2
        \le \beta_t := \tilde{\Ocal} \left( \mu^{-2} \kappa^{-4} d_w \right) \, .        
    \end{equation*}
\end{lemma}
The construction of the confidence set in Lemma~\ref{lemma:confidence ball} relies on two key components.
The first is the careful choice of the exploration phase length $t_0$ and the regularization parameter $\lambda$, both of which play a critical role in ensuring the confidence radius $\beta_t$ scales as $\Ocal (\log T)$, which is essential for achieving a cumulative regret of $\tilde{\Ocal}(\sqrt{T})$.

The second challenge is that for $\lambda = \Ocal(\sqrt{T})$, existing concentration results for utility parameter estimation in MNL or GLM bandits~\cite{faury2020improved, perivier2022dynamic} are no longer applicable.
This is because these results rely on Bernstein-type inequalities, which yield a confidence radius of order $\Ocal\left(\sqrt{\lambda} + \frac{d \log T}{\sqrt{\lambda}}\right)$ where $d$ is the dimension of the context feature vector.
To ensure the confidence radius is $\Ocal (\log T)$, these bounds require $\lambda = \Ocal(d \log T)$, which limits their applicability when $\lambda = \Ocal(\sqrt{T})$.
To overcome this, we first establish a new concentration inequality for the utility parameter in the MNL choice model that is independent of $\lambda$ (up to logarithmic factors), presented in Lemma~\ref{sub_lemma:lambda-indep. concentration}.
Combined with carefully chosen values of $\lambda$ and $t_0$, we then use a mathematical induction argument to show that for all rounds $t$ in Phase II, the estimation error satisfies $\min_{\tilde{\wb} \in \Wcal^*} \| \hat{\wb}_t - \tilde{\wb} \|_2^2 = \tilde{\Ocal}(1/t_0)$.
Finally, we prove that this implies $\min_{\tilde{\wb} \in \Wcal^*} \| \hat{\wb}_t - \tilde{\wb} \|^2_{\Vb_t} \le \beta_t$.

Next, based on the confidence radius $\beta_t$ established in Lemma~\ref{lemma:confidence ball}, we show that the optimistic utility estimate $z_{ti}$ is greater than or equal to the true utility $f_{\wb^*}(\xb_{ti})$ as follows:
\begin{lemma}[Optimistic utility] \label{lemma:optimistic utility}
    Suppose Assumptions~\ref{assm:realizability},~\ref{assm:boundedness},~\ref{assm:stochastic context} and~\ref{assm:generalized geometric condition} hold and $\min_{\tilde{\wb} \in \Wcal^*} \| \hat{\wb}_t - \tilde{\wb} \|_{\Vb{t}}^2 \le \beta_t$, for all $t_0 + 1 \le t \le T$.
    Then, for all $t_0 + 1 \le t \le T$ and $i \in S_t$, we have
    \begin{equation*}
        0 \le z_{ti} - f_{\wb^*}(\xb_{ti}) \le 2 \sqrt{\beta_t} \|\nabla f_{\hat{\wb}_{t}}(\xb_{ti})\|_{\Vb_{t}^{-1}} + \frac{2 \beta_t C_h}{\lambda} \, .
    \end{equation*}
\end{lemma}
Unlike the linear utility case~\cite{oh2021multinomial}, our optimistic utility estimate for $f_{\wb^*}$ introduces an item-independent extra term of the form $2\beta_t C_h/\lambda$.
This term arises as a residual from the first-order approximation of the unknown utility function $f_{\wb^*}$, and unlike standard deviation terms $\| \nabla f_{\hat{\wb}_t}(\xb_{ti}) \|_{\Vb_t^{-1}}$, it does not vanish over time.
As a result, it poses an additional challenge in achieving a cumulative regret of $\tilde{\Ocal}(\sqrt{T})$.
Nevertheless, we show that with a regularization parameter $\lambda = \tilde{\Ocal}(\sqrt{T})$, it is still possible to achieve both optimism and cumulative regret of $\Ocal(\sqrt{T})$.
We then show that the expected reward of the assortment $S_t$, chosen based on the optimistic utility estimates $z_{ti}$, exceeds the true expected reward of the optimal assortment $S_t^*$ with high probability. 

\section{Numerical Experiments} \label{sec:numerical experiments}

\begin{figure*}[t]
  \centering
  \begin{subfigure}[b]{0.48\textwidth}
    \centering
    \includegraphics[width=\linewidth]{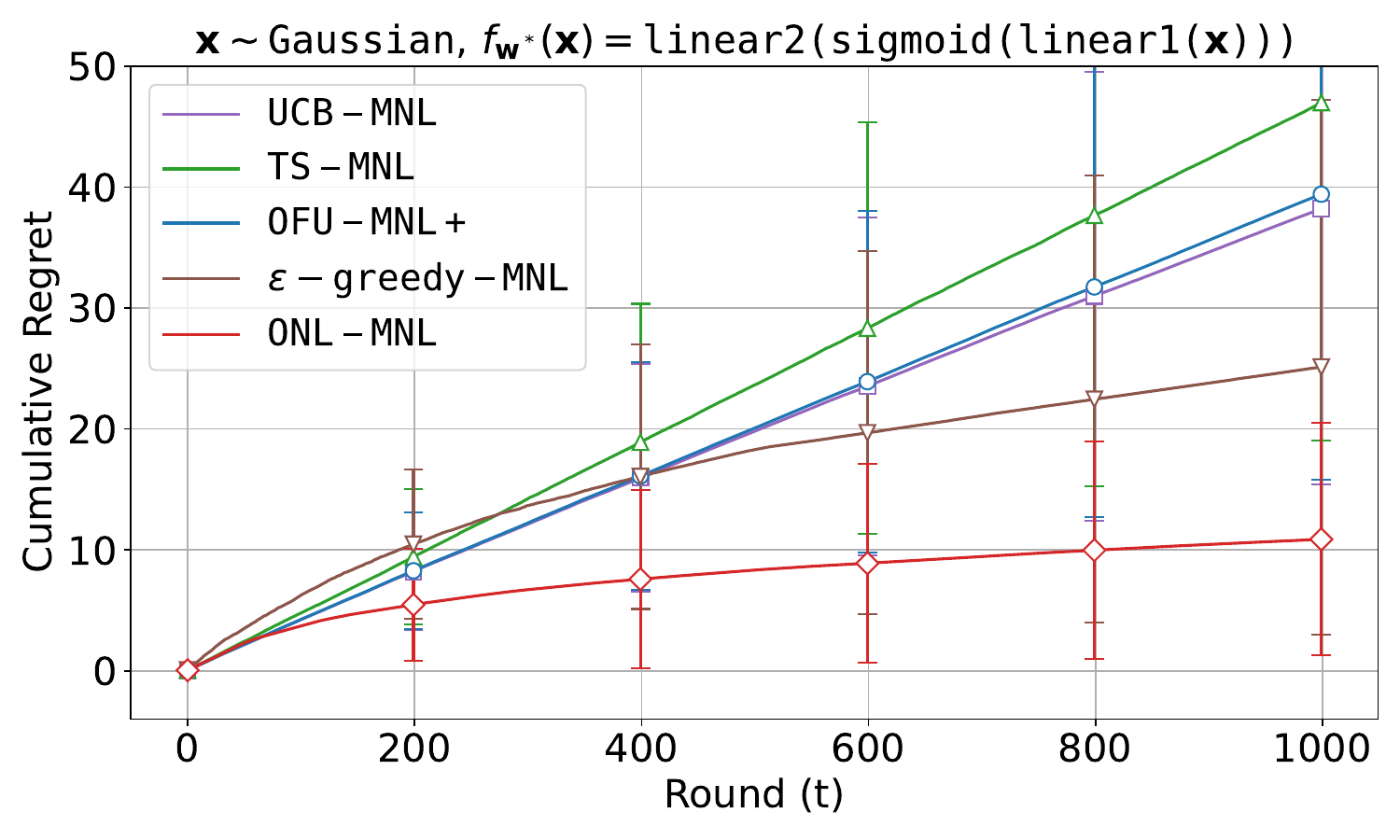}
    \caption{Realizability (Gaussian)}
  \end{subfigure}
  \begin{subfigure}[b]{0.48\textwidth}
    \centering
    \includegraphics[width=\linewidth]{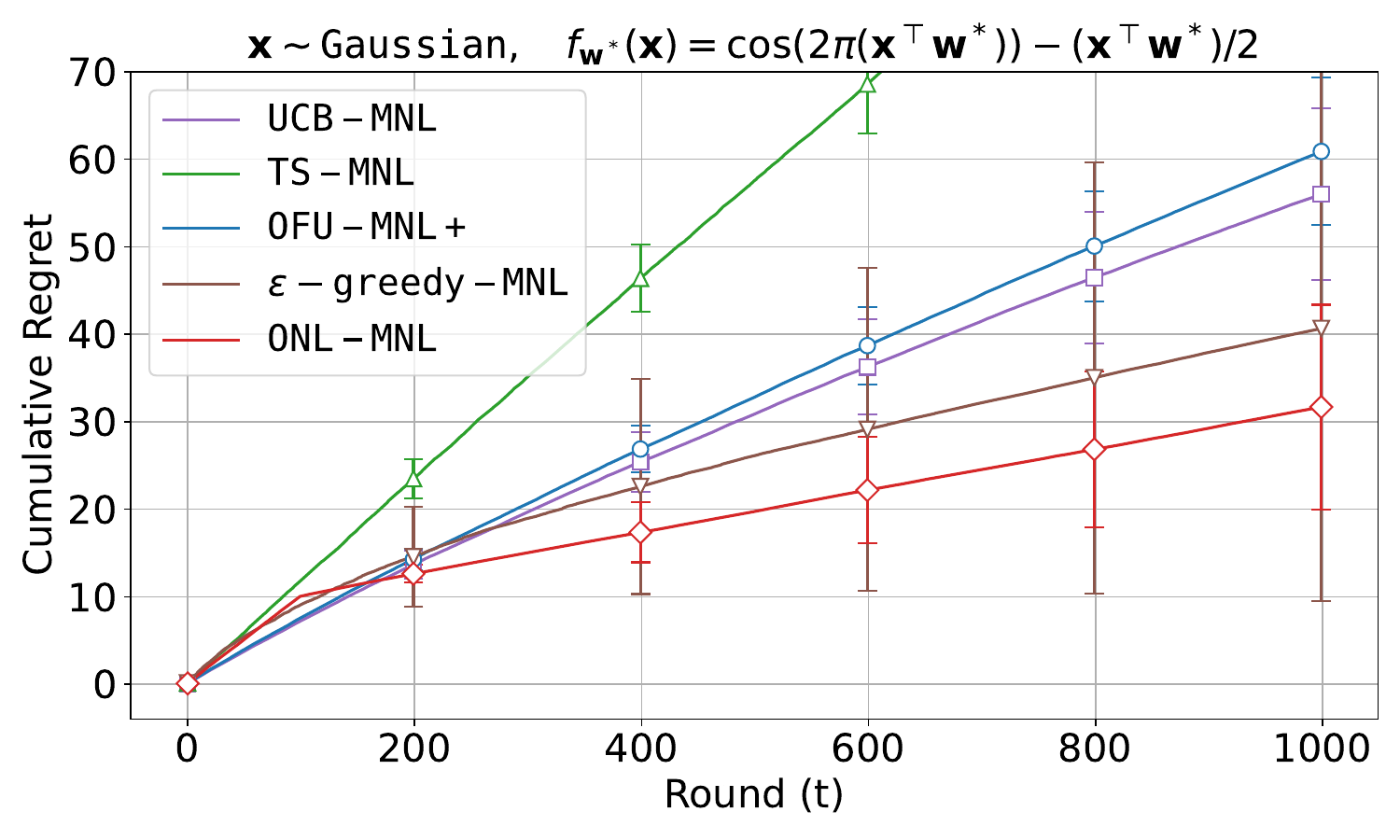}
    \caption{Misspecification (Gaussian)}
  \end{subfigure}
  
  \caption{
  Cumulative regret comparison between $\algname$ (ours) and baselines under Gaussian contexts.
  The results for the uniform context distribution are provided in Figure~\ref{fig:main exp_uniform}.
  }
  \label{fig:main exp_gaussian}
\end{figure*}

\paragraph{MNL choice model with non-linear utilities.}
In this section, we present numerical experiments to evaluate the performance of the proposed algorithm $\algname$.
As baselines, we compare with algorithms for contextual MNL bandits with linear utilities--\texttt{UCB-MNL}~\cite{oh2021multinomial}, \texttt{TS-MNL}~\cite{oh2019thompson}, and \texttt{OFU-MNL+}~\cite{lee2024nearly}--as well as $\varepsilon \texttt{-greedy-MNL}$~\cite{zhang2024contextual}, a regression-oracle-based method with uniform exploration designed for general utilities.
Each experiment was conducted independently over 30 random seeds, and we report the mean and standard deviation of the cumulative regret.
We set $N=100$, $K=5$, $d=3$, and $r_{ti}=1$. 
Experiments were performed under two different context distributions: $\xb \sim \Ncal(\zero_d, \Ib_d)$ and $x_i \sim \unif[-3,3], \forall i \in [d]$.
We use a two-layered neural network with sigmoid activation as the estimator $\hat{f}$ for our proposed algorithm, i.e., 
$\hat{f}(\xb) = \texttt{linear2}(\texttt{sigmoid}(\texttt{linear1}(\xb)))$.

For the realizable setting, the true utility function $f_{\wb^*}$ is defined as a two-layer neural network with the same architecture as $\hat{f}$, and we set the number of neurons in the hidden layer to $3$.
Each entry of the true utility parameter $\wb^*$ is independently sampled from the uniform distribution over the interval $[-1, 1]$.
We set the length of Phase I to $t_0 = 50$ and ran experiments for $T = 1000$.

For the misspecified setting, the true utility function is defined as $f_{\wb^*}(\xb) = \cos(2\pi \, (\xb^\top \wb^*)) - \frac{1}{2} (\xb^\top \wb^*)$, which differs in form from the estimator architecture.
To account for this increased complexity, we set the number of neurons in the hidden layer of the estimator to $15$.
Accordingly, we also increase the length of Phase I to $t_0 = 100$.

The experimental results in Figure~\ref{fig:main exp_gaussian} clearly demonstrate the effectiveness of our proposed algorithm, $\algname$, in both realizable and misspecified settings.
Existing linear utility-based methods exhibit steadily increasing regret under non-linear utilities, as they fail to capture the complex structure of the true utility function.
Compared to the $\varepsilon$-\texttt{greedy-MNL} algorithm designed for general utilities~\cite{zhang2024contextual}, our approach is not only statistically efficient--achieving low cumulative regret--but also exhibits robust and consistently strong performance across different scenarios in both realizable and misspecified settings.
See Appendix~\ref{appx:additional experiments} for additional experimental results and details.

\paragraph{Effect of the number of item on regret performance.}

\begin{figure*}[t]
  \begin{subfigure}[b]{0.48\textwidth}
    \centering
    \includegraphics[width=\linewidth]{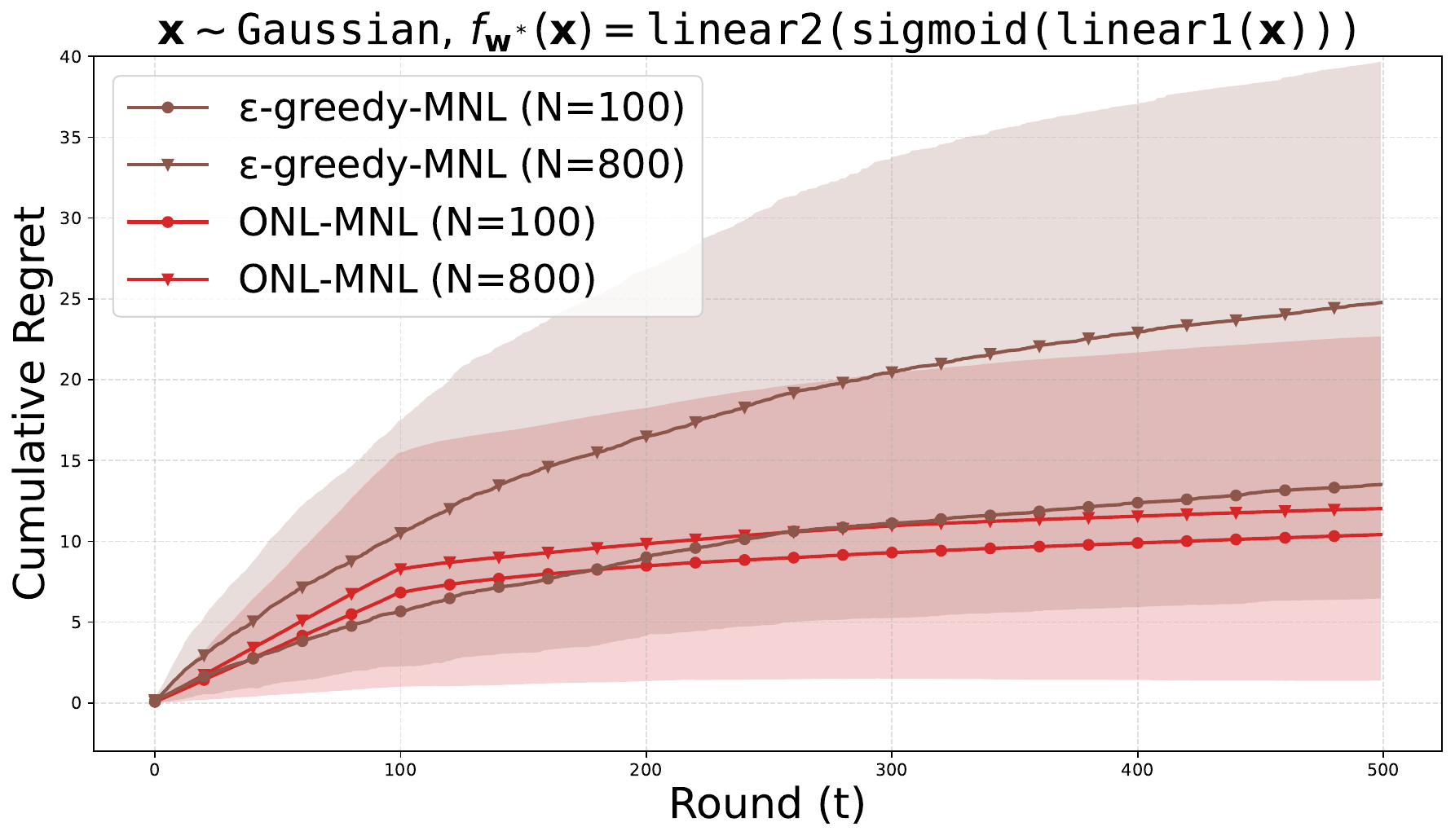}
    \caption{Gaussian}
  \end{subfigure}
  \hfill
  \begin{subfigure}[b]{0.48\textwidth}
    \centering
    \includegraphics[width=\linewidth]{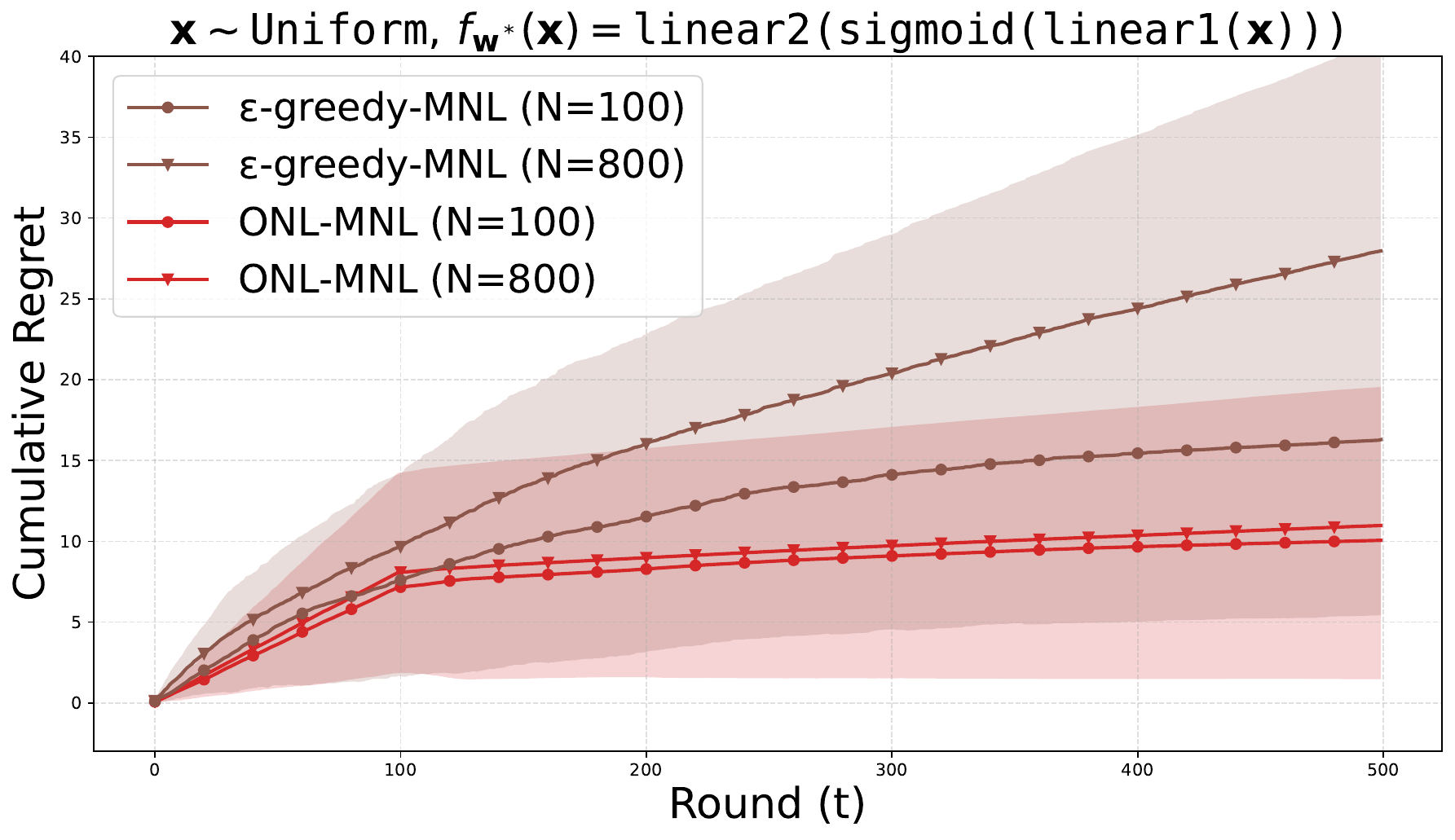}
    \caption{Uniform}
  \end{subfigure}

  \caption{Cumulative regret comparison between $\algname$ (ours) and $\varepsilon \texttt{-greedy-MNL}$~\cite{zhang2024contextual} under varying number of items $N$. 
  }
  \label{fig:add. exp}
\end{figure*}

To further highlight the distinct advantages of our proposed algorithm, $\algname$, over the $\varepsilon \texttt{-greedy-MNL}$~\cite{zhang2024contextual}, we compare the performances of those algorithms as the total number of items $N$ increases.
Beyond being computationally tractable, a key advantage of our algorithm is that its regret bound is independent of $N$, whereas the regret bound for $\varepsilon \texttt{-greedy-MNL}$ grows polynomially with $N$.
To verify this, we conduct experiments under a realizable setting where the true utility function is a two-layer neural network with sigmoid activation and $10$ hidden neurons.
We evaluate the cumulative regret of both algorithms at $T = 500$ for different values of $N \in \{100, 800\}$.
Following~\cite{zhang2024contextual}, where the optimal exploration rate satisfies $\varepsilon = \Ocal(N^{1/3})$ (Theorem 3.4 in~\cite{zhang2024contextual}), we set $\varepsilon = 0.1$ and $0.2$ for $N = 100$ and $800$, respectively.

Figure~\ref{fig:add. exp}-(a) and (b) show the results under Gaussian and uniform context distributions, respectively.
As expected, the cumulative regret of $\varepsilon \texttt{-greedy-MNL}$ increases as $N$ grows, which aligns with its theoretical regret bound that scales polynomially with $N$.
While our algorithm also exhibits a slight increase in cumulative regret as $N$ grows, this is primarily due to the increased regret incurred during the uniform exploration phase (Phase I).
As the total number of items increases, the size of the assortment set $\Scal$ also grows, leading to a larger gap between the expected reward of a uniformly sampled assortment and that of the optimal assortment.
Nevertheless, in Phase II, our algorithm consistently converges toward the optimal policy, regardless of the value of $N$.
In other words, the slope of the regret curve during Phase II remains relatively stable across different values of $N$, indicating that the learning efficiency of our method is unaffected by the size of the item set.
These results demonstrate that our method not only offers provable computational and statistical efficiency but also remains robust and effective even in settings with extremely large item pools.

\section{Conclusion} \label{sec:conclusion}
We studied contextual MNL bandits with non-linear parametric utilities and proposed a UCB-based algorithm that is both computationally tractable and statistically efficient. Our algorithm attains $\tilde{\Ocal}(\sqrt{T})$ regret independent of the number of items, to the best of our knowledge the first tractable method with this guarantee for non-linear utilities. Relative to prior approaches for general utility models, our result achieves the same $\tilde{\Ocal}(\sqrt{T})$ rate while remaining implementable under non-linearity.

Looking ahead, the current regret bound depends on the problem-dependent instance factor $\kappa$; reducing or removing this dependence in the non-linear setting likely requires new analytical techniques and remains an important direction. Another promising avenue is to extend our framework to MNL MDPs~\cite{hwang2023model}, where the transition kernel is specified by an MNL model. Prior work analyzes only the case in which the logit is a linear function of state–action features~\citep{hwang2023model,cho2024randomized,li2024provably,park2025infinite}; generalizing to \emph{non-linear} parameterizations while preserving computational tractability and performance guarantees would broaden the scope and impact of this line of research.

\clearpage
\section*{Acknowledgements}
This work was supported by the National Research Foundation of Korea~(NRF) grant funded by the Korea government~(MSIT) (No.  RS-2022-NR071853 and RS-2023-00222663), by Institute of Information \& communications Technology Planning \& Evaluation~(IITP) grant funded by the Korea government~(MSIT) (No. RS-2025-02263754), and by AI-Bio Research Grant and Ascending SNU Future Leader Fellowship through Seoul National University.

\bibliographystyle{abbrvnat}
\bibliography{references}

\newpage

\clearpage
\appendix

\renewcommand{\contentsname}{Contents of Appendix}
\addtocontents{toc}{\protect\setcounter{tocdepth}{3}}
{
  \hypersetup{hidelinks}
  \tableofcontents
}

\section{Related Works}

\paragraph{Contextual MNL bandits.}
For contextual MNL bandits under linear utility assumptions,
~\citet{ou2018multinomial} consider a setting in which the item-context features remain fixed over time and achieve $\tilde{\Ocal}(d K\sqrt{T})$ regret bound.
~\citet{chen2020dynamic} extended the fixed context setting to stochastic and time-varying contexts, deriving a regret bound of $\tilde{\Ocal}(d\sqrt{T} + d^2 K^2)$; however, their method is computationally intractable due to the need to enumerate all possible assortments.
~\citet{oh2021multinomial} introduced a polynomial-time algorithm that maintains confidence bounds in the parameter space and computes UCB scores for each item, achieving regret $\tilde{\Ocal}(dK\sqrt{T/\kappa})$, where $\kappa$ is the problem-dependent instance factor.
~\citet{perivier2022dynamic} further refined this under adversarial contexts and non-uniform rewards, improving the dependency on $\kappa$ and achieving $\tilde{\Ocal}(dK\sqrt{\kappa^* T} + d^2K^4/\kappa)$, but their method is not computationally tractable. 
~\citet{zhang2023online} proposed a computationally efficient algorithm using online parameter updates for a multi-parameter setting, where a shared contextual feature vector is used across items, but each item is associated with its own utility parameter.
~\citet{lee2024nearly} proposed a computationally efficient algorithm that achieves minimax optimal regret (up to logarithmic factors) in the single-parameter setting.

Extending beyond the linearity assumption,~\citet{zhang2024contextual} is, to the best of our knowledge, the only work that addresses contextual MNL bandits with a general function class.
For the stochastic context setting, they proposed two offline regression oracle-based algorithms.
Assuming the function class is Lipschitz, the uniform exploration method achieves a regret bound of $\tilde{\Ocal}((d K N)^{1/3} T^{2/3})$, while the log-barrier regularization method achieves a tighter bound of $\tilde{\Ocal}(K^2 \sqrt{d N T})$.
For the adversarial context setting, they proposed two online regression oracle-based algorithms.
Under the same Lipschitz assumption, the uniform exploration algorithm attains a regret of $\tilde{\Ocal}((N K)^{1/3} T^{5/6})$, and the log-barrier method again achieves $\tilde{\Ocal}(K^2 \sqrt{d N} T^{3/4})$.
Additionally, they showed that a variant of Feel-Good Thompson Sampling~\cite{zhang2022feel} achieves a regret of $\tilde{\Ocal}(K^2 \sqrt{d N T})$.
Even though these results offer significant theoretical insights for general utility function classes, the log-barrier regularized exploration and Thompson sampling-based algorithms are not computationally tractable—that is, they cannot be solved in polynomial time.
Furthermore, all of their algorithms exhibit polynomial dependence on the total number of items $N$, limiting their applicability to environments with large or infinite item sets.
In contrast, our proposed algorithm is not only computationally tractable but also achieves a regret bound of $\tilde{\Ocal}(\sqrt{T})$, which remains independent of $N$, even in the presence of non-linear utilities.

\paragraph{Beyond linear models in contextual bandits.}
For generalized linear reward models, a growing body of work has provided algorithms with provable guarantees~\cite{filippi2010parametric, li2017provably, faury2020improved, faury2022jointly}. Most of these methods achieve a regret bound of $\tilde{\mathcal{O}}(d \sqrt{T})$, with subsequent improvements focusing on instance-dependent analysis and computational efficiency.
For more general parametric reward models,~\citet{liu2023global} consider the infinite-armed bandit setting and establish a regret bound of $\tilde{\mathcal{O}}(d_w^2 \sqrt{T})$, where $d_w$ denotes the dimension of the parameterized reward model.
Their analysis also relies on a geometric condition on the squared loss of the parametric model, but crucially assumes a unique global minimum—a stronger condition than our Assumption~\ref{assm:generalized geometric condition}.

Beyond parametric reward models, there has been a surge of interest in bandit algorithms utilizing neural networks~\cite{zhou2020neural, xu2021neural, zhang2021neural, kassraie2022neural}.
However, these works typically rely on the assumption that the reward function lies in the reproducing kernel Hilbert space and heavily depend on NTK theory.
Moreover, regret bounds derived under NTK assumptions are known to suffer from unfavorable worst-case information gain of order $\tilde{\mathcal{O}}(T^{\frac{d-1}{d}})$, leading to a regret bound of $\tilde{\mathcal{O}}(T^{\frac{2d - 1}{2d}})$, which is significantly worse than the standard sublinear regret of $\tilde{\mathcal{O}}(\sqrt{T})$~\cite{kassraie2022neural}.
Without relying on NTK theory,~\citet{xu2021robust} analyze a quadratic neural network with a single hidden layer and establish a regret bound of $\tilde{\Ocal}(T^{2/3})$.
~\citet{huang2021going} consider a two-layer ReLU neural network with fewer hidden units than input dimensions under the assumption of Gaussian input distributions.
~\citet{xu2024stochastic} study a specific two-layer ReLU architecture in which all second-layer weights are fixed to one, assuming a unit sphere context distribution.
While these approaches achieve provably efficient algorithms without relying on NTK assumptions, their dependence on specific network architectures and restrictive context distributions limits their applicability to more general settings.

\section{Proof of Lemma~\ref{lemma:convergence of w_0}} \label{appx:convergence of w_0}

\begin{proof}[Proof of Lemma~\ref{lemma:convergence of w_0}]
    For notational simplicity, we abbreviate the following process:
    \begin{equation} \label{eq:uniform sampling process}
        X \sim \Dcal, \,  S \sim \unif(\Scal),  \, i \sim \mathrm{Multinomial} \left(1, [p(0 \mid X, S, \wb^*), \ldots, p(i_{|S|} \mid X, S, \wb^*)] \right)
    \end{equation}
    as $(X, S, i) \sim \Hcal$.    
    Under Assumption~\ref{assm:boundedness}, each function \( f_\wb \in \Fcal \) is Lipschitz continuous with constant \( C_f \).  
    Then, by the regression oracle guarantee for Lipschitz function classes (Lemma~\ref{aux_lemma:generalization bound of MLE for iid}), there exists a constant $C_1 > 0$ such that, with probability at least $1 - \delta$, the following bound holds:    
    \begin{equation*}
        \EE_{(X, S, i) \sim \Hcal} \left[ \log p(i \mid X, S, \wb^*) - \log p(i \mid X, S, \hat{\wb}_0) \right]
        \le C_1 \frac{d_w C_f^2 \zeta}{t_0} \, ,
    \end{equation*}
    where $\zeta = \log K \log (t_0 C_g) \log (1/\delta)$ is the logarithmic term.

    For any $X = \{ \xb_1, \ldots, \xb_N\}, S \in \Scal$, and $\wb \in \Wcal$, we define
    \begin{align*}
        & \pb(X, S, \wb) := \left[ p(i_1 \mid X, S, \wb) \, ,  \ldots \, ,  p(i_{|S|} \mid X, S, \wb) \right] \, , \quad
        \\
        & \pb_{0}(X, S, \wb) := \left[ p(0 \mid X, S, \wb) \, , p(i_1 \mid X, S, \wb) \, , \ldots, p(i_{|S|} \mid X, S, \wb) \right] \, .
    \end{align*}
    Then, we have
    \begin{align*}
        C_1 \frac{ d_w C_f^2 \zeta}{t_0}
        & \ge \EE_{(X, S, i) \sim \Hcal} \left[ \log p(i \mid X, S, \wb^*) - \log p(i \mid X, S, \hat{\wb}_0) \right]
        \\
        & = \EE_{X, S} \left[ \EE_{i \sim \pb_0(X, S, \wb^*)} \left[ \log p(i \mid X, S, \wb^*) - \log p(i \mid X, S, \hat{\wb}_0) \mid X, S\right] \right]
        \\
        & = \EE_{X, S} \left[ \kldiv( \pb_0 (X,S, \wb^*), \pb_0(X, S, \hat{\wb}_0)) \mid X, S \right]
        \\
        & \ge \EE_{X, S} \left[ \frac{1}{2} \| \pb_0 (X,S, \wb^*) - \pb_0(X, S, \hat{\wb}_0) \|_2^2 \mid X, S \right]
        \\
        & \ge \EE_{X, S} \left[ \frac{1}{2} \| \pb (X,S, \wb^*) - \pb(X, S, \hat{\wb}_0) \|_2^2 \mid X, S \right] \, , \numberthis \label{eq:w_0 eq 1}
    \end{align*}    
    where the second inequality follows from Pinsker’s inequality, and the final inequality follows from the definitions of $\pb$ and $\pb_0$.
    For the next step, we introduce the following lemma showing reverse Lipschitzness of the MNL model. 
    The proof of Lemma~\ref{sub_lemma:reverse Lipschitzness of the MNL model} is provided in Section~\ref{appx:technical lemma for convergence of w0}.

    \begin{lemma}[Reverse Lipschitzness of the MNL model] \label{sub_lemma:reverse Lipschitzness of the MNL model}
        For $\ab \in \Acal := \{ \xb \in \RR^d: \| \xb \|_2 \le C \}$, let $\hb(\ab) = [h_1(\ab), \ldots, h_d(\ab)]$ be defined as
        \begin{equation*}
            h_i (\ab) = \frac{\exp(a_i)}{1 + \sum_{j=1}^d \exp(a_j)} \, .
        \end{equation*}
        Then, for any $\ab, \bb \in \Acal$, 
        \begin{equation*}
            \| \hb (\ab) - \hb(\bb) \|_2 \ge \kappa_0 \| \ab - \bb \|_2 \, ,
        \end{equation*}
        where $\kappa_0 := \min_{\ab \in \Acal} \min_{i \in [d]} h_i(\ab) ( 1 - \sum_{j=1}^d h_j(\ab) )$.
    \end{lemma}

    Note that if we denote $\fb(X, S, \wb):= [ f_{\wb}(\xb_j) ]_{j \in S} \in \RR^{|S|}$, then we have $ \hb(\fb(X, S, \wb)) = \pb(X, S, \wb)$. Therefore by applying Lemma~\ref{sub_lemma:reverse Lipschitzness of the MNL model}, Eq.~\eqref{eq:w_0 eq 1} can be further bounded by
    \begin{align*}
        \eqref{eq:w_0 eq 1}
        & \ge \frac{ \kappa^2}{2} \EE_{X, S} \left[ \sum_{j \in S} \left( f_{\hat{\wb}_0}(\xb_j) - f_{\wb^*}(\xb_j) \right)^2 \mid X, S \right]
        = \frac{\kappa^2}{2} \ell_{\mathrm{sq}}(\hat{\wb}_0) \, .
    \end{align*}
    Therefore, we have
    \begin{equation*}
        \ell_{\mathrm{sq}}(\hat{\wb}_0) \le 2 C_1 \frac{\kappa^{-2} d_w C_f^2 \zeta}{t_0} \, .
    \end{equation*}
    
    On the other hand, let $\tilde{\wb}^* = \argmin_{\tilde{\wb} \in \Wcal^*} \| \hat{\wb}_0 - \tilde{\wb} \|_2$.
    Then, by the condition on $t_0$,
    \begin{equation} \label{eq:n condition}
        t_0 \ge 2^{-1} C \kappa^{-2} d_w C_f^2 \zeta \frac{\mu^{\gamma/(2-\gamma)}}{\tau^{2/(2-\gamma)}} \, ,
    \end{equation}
    where $C = 4 C_1$, we can obtain
    \begin{equation*}
        \| \hat{\wb}_0 - \tilde{\wb}^* \|_2 \le (\tau / \mu)^{1/(2-\gamma)} \, .
    \end{equation*}
    We can prove it by contradiction. Suppose that 
    \begin{equation*}
        \| \hat{\wb}_0 - \tilde{\wb}^* \|_2 > (\tau / \mu)^{1/(2-\gamma)} \, .
    \end{equation*}
    Then, we have 
    \begin{equation} \label{eq:w0 eq 1}
         \frac{\mu}{2} \| \hat{\wb}_0 - \tilde{\wb}^* \|_2^2 > \frac{\tau}{2} \| \hat{\wb}_0 - \tilde{\wb}^* \|_2^\gamma \, , 
    \end{equation}
    which implies that the $(\tau, \gamma)$-growth condition holds by Assumption~\ref{assm:generalized geometric condition}.
    On the other hand, by the $(\tau, \gamma)$-growth condition in Assumption~\ref{assm:generalized geometric condition}, Eq.~\eqref{eq:w0 eq 1} implies 
    \begin{equation*}
        \frac{\tau}{2} \| \hat{\wb}_0 - \tilde{\wb}^* \|_2^\gamma \le \ell_{\mathrm{sq}}(\hat{\wb}_0) - \ell_{\mathrm{sq}}(\tilde{\wb}^*) = \ell_{\mathrm{sq}}(\hat{\wb}_0)
        \le 2 C_1 \frac{\kappa^{-2} d_w C_f^2 \zeta}{t_0} \, .
    \end{equation*}
    By the condition on $t_0$ in Eq.~\eqref{eq:n condition}, this implies 
    \begin{equation*}
        \| \hat{\wb}_0 - \tilde{\wb}^* \|_2 \le (\tau / \mu)^{1/(2-\gamma)} \, ,
    \end{equation*}
    which leads to a contradiction.
    This shows that when $t_0$ satisfies Eq.~\eqref{eq:n condition}, then 
    $\hat{\wb}_0$ lies in the region where the local strong convexity of $\ell_{\mathrm{sq}} (\cdot)$ at $\tilde{\wb}^*$ holds. 
    Therefore, we obtain
    \begin{equation*}
        \frac{\mu}{2} \| \hat{\wb}_0 - \tilde{\wb}^* \|_2^2 \le \ell_{\mathrm{sq}}(\hat{\wb}_0) - \ell_{\mathrm{sq}}(\tilde{\wb}^*) = \ell_{\mathrm{sq}}(\hat{\wb}_0) - \ell_{\mathrm{sq}}(\wb^*) \le  2 C_1 \frac{\kappa^{-2} d_w C_f^2 \zeta}{t_0} \, , 
    \end{equation*}
    which results in
    \begin{equation*}
        \| \hat{\wb}_0 - \tilde{\wb}^* \|_2^2 \le C \frac{\kappa^{-2} d_w C_f^2 \zeta}{\mu t_0} \, ,
    \end{equation*}
    where we again denote $C = 4 C_1$.
\end{proof}

\subsection{Proofs of Technical Lemmas} \label{appx:technical lemma for convergence of w0}
\subsubsection{Proof of Lemma~\ref{sub_lemma:reverse Lipschitzness of the MNL model}}
\begin{proof}[Proof of Lemma~\ref{sub_lemma:reverse Lipschitzness of the MNL model}]
    For any $\ab, \bb \in \Acal$, by the mean value theorem there exists $\cbb \in \mathrm{conv}(\ab, \bb)$ such that
    \begin{equation*}
        \hb(\ab) - \hb(\bb) = \nabla \hb(\cbb)^\top (\ab - \bb) \, .
    \end{equation*}
    Now, it suffices to show that $\lambda_{\min} (\nabla(\hb(\cbb)^\top) \ge \kappa_0$.
    Since $\nabla \hb(\cbb) = \mathrm{diag}(\hb(\cbb)) - \hb(\cbb) \hb(\cbb)^\top$, 
    for any $\xb \in \RR^d \setminus \{ \zero_d \}$, 
    \begin{align*}
        \xb^\top \nabla \hb(\cbb) \xb & = \xb^\top \mathrm{diag}(\hb(\cbb)) \xb - \left( \xb^\top \hb(\cbb) \right)^2
        \\
        & = \sum_{i=1}^d h_i (\cbb) x_i^2 - \left( \sum_{i=1}^d h_i(\cbb) x_i \right)^2
        \\
        & \ge \sum_{i=1}^d h_i (\cbb) x_i^2 - \left( \sum_{i=1}^d h_i(\cbb) \right) \left( \sum_{i=1}^d h_i(\cbb) x_i^2 \right)
        \\
        & = \left( \sum_{i=1}^d h_i(\cbb) x_i^2 \right) \left( 1 - \sum_{i=1}^d h_i(\cbb) \right)
        \\
        & = \sum_{i=1}^d h_i(\cbb) \left( 1 - \sum_{i=1}^d h_i(\cbb) \right) x_i^2 
        \\
        & \ge  \sum_{i=1}^d \kappa_0  x_i^2 \, ,
    \end{align*}
    where the first inequality uses Cauchy-Schwarz inequality, and the last inequality follows by the definition of $\kappa_0$. This concludes the proof.
\end{proof}

\section{Proof of Lemma~\ref{lemma:confidence ball}} \label{appx:confidence ball}
\begin{proof}[Proof of Lemma~\ref{lemma:confidence ball}]
    Let $\tilde{\wb}^* = \argmin_{\tilde{\wb} \in \Wcal^*} \| \hat{\wb}_0 - \tilde{\wb} \|_2$. 
    And for notational simplicity, we will abbreviate $\hat{p}(i \mid X_t, S_t, \wb)$ as $\hat{p}_{ti}(\wb)$, and $p(i \mid X_t, S_t, \wb)$ as $p_{ti}(\wb)$ in this proof.
    Recall that $\ell_t(\wb)$ is the regularized negative log-likelihood defined as
    \begin{equation*}
        \ell_t (\wb) := - \sum_{s=t_0 + 1}^{t-1} \sum_{i \in S_s} y_{si} \log \hat{p}_{si}(\wb) + \frac{\lambda}{2} \| \wb - \hat{\wb}_0 \|_2^2 \, .
    \end{equation*}
    Since $\hat{\wb}_t = \argmin_{\wb \in \Wcal} \ell_t( \wb)$, we have $\nabla \ell_t(\wb_t) = \zero_{d_w}$, and this implies
    \begin{equation} \label{eq:gradient l equal 0}
        \sum_{s=t_0 + 1}^{t-1} \sum_{i \in S_s} \hat{p}_{si}(\hat{\wb}_t) \nabla f_{\hat{\wb}_s}(\xb_{si}) + \lambda (\hat{\wb}_t - \hat{\wb}_0) = \sum_{s=t_0 + 1}^{t-1} \sum_{i \in S_s} y_{si} \nabla f_{\hat{\wb}_s}(\xb_{si}) \, .
    \end{equation}
    Let us define $\gb_t (\wb) = \sum_{s=t_0 + 1}^{t-1} \sum_{i \in S_s} \hat{p}_{si}(\wb) \nabla f_{\hat{\wb}_s}(\xb_{si}) + \lambda \wb$. 
    Then, from Eq.~\eqref{eq:gradient l equal 0} we have
    \begin{align*}
        \gb_t (\hat{\wb}_t) - \gb_t (\tilde{\wb}^*) & = \sum_{s=t_0 + 1}^{t-1} \sum_{i \in S_s} \left( y_{si} - \hat{p}_{si}(\tilde{\wb}^*) \right) \nabla f_{\hat{\wb}_{s}}(\xb_{si}) + \lambda(\hat{\wb}_0 - \tilde{\wb}^*) \, .
    \end{align*}    
    Note that since $\EE[y_{si} \mid \Fcal_s] \ne \hat{p}_{si}(\tilde{\wb}^*)$, we cannot apply conventional self-normalized martingale inequality used in previous MNL bandit~\cite{oh2019thompson, oh2021multinomial, perivier2022dynamic, zhang2023online, lee2024nearly}. 
    Instead since $\EE[y_{si} \mid \Fcal_s] = p_{si}(\tilde{\wb}^*)$ we add and subtract $p_{si}(\tilde{\wb}^*)$ as follows:
    \begin{align*}
        \gb_t (\hat{\wb}_t) - \gb_t (\tilde{\wb}^*) & = \sum_{s=t_0 + 1}^{t-1} \sum_{i \in S_s} \left( y_{si} - p_{si}(\tilde{\wb}^*) \right) \nabla f_{\hat{\wb}_{s}}(\xb_{si})
            + \sum_{s=t_0 + 1}^{t-1} \sum_{i \in S_s} \left( p_{si}(\tilde{\wb}^*) - \hat{p}_{si}(\tilde{\wb}^*) \right) \nabla f_{\hat{\wb}_{s}}(\xb_{si})
            \\
            & \quad 
            + \lambda(\hat{\wb}_0 - \tilde{\wb}^*) \, .
    \end{align*}
    By inequality $(a+b+c)^2 \le 4a^2 + 4b^2 + 4c^2$, we have
    \begin{align*}
        & \| \gb_t (\hat{\wb}_t) - \gb_t (\tilde{\wb}^*) \|_{\Vb_{t}^{-1}}^2 
        \\
        & \le \underbrace{ 4 \left\| \sum_{s=t_0 + 1}^{t-1} \sum_{i \in S_s} \left( y_{si} - p_{si}(\tilde{\wb}^*) \right) \nabla f_{\hat{\wb}_{s}}(\xb_{si}) \right\|^2_{\Vb_{t}^{-1}}}_{I_1}
            + \underbrace{4\left \| \sum_{s=t_0 + 1}^{t-1} \sum_{i \in S_s} \left( p_{si}(\tilde{\wb}^*) - \hat{p}_{si}(\tilde{\wb}^*) \right) \nabla f_{\hat{\wb}_{s}}(\xb_{si}) \right \|^2_{\Vb_t^{-1}}}_{I_2}
        \\
            & \quad + \underbrace{4 \lambda^2 \| \hat{\wb}_0 - \tilde{\wb}^* \|^2_{\Vb_t^{-1}}}_{I_3} 
    \end{align*}

    For the term $I_1$, we introduce the following lemma.
    \begin{lemma} \label{sub_lemma:lambda-indep. concentration}
        For all $t \ge t_0 + 1$, with probability at least $1 - \delta$,
        \begin{equation*}
            \left\| \sum_{s=t_0 + 1}^{t-1} \sum_{i \in S_s} \left( y_{si} - p_{si}(\tilde{\wb}^*) \right) \nabla f_{\hat{\wb}_{s}}(\xb_{si}) \right\|^2_{\Vb_{t}^{-1}}
            \le \alpha_t \, ,
        \end{equation*}
        where $\alpha_t = \tilde{C}_\alpha d_w \log(1 + T C_g^2 /(d_w \lambda)) \log(t^2 \pi^2/\delta)$ for an absolute constant $\tilde{C}_\alpha > 0$.
    \end{lemma}

    If we denote each $y_{si} - p_{si}(\wb^*)$ as $\varepsilon_{si}$ and note $\EE[\varepsilon_{si}]=0$, the quantity $\left\| \sum_{s=t_0 + 1}^{t-1} \sum_{i \in S_s} \left( y_{si} - p_{si}(\tilde{\wb}^*) \right) \nabla f_{\hat{\wb}_{s}}(\xb_{si}) \right\|^2_{\Vb_{t}^{-1}}$ corresponds to the self-normalized vector-valued martingale term in~\citet{abbasi2011improved}.
    However, due to correlations among items within each selected assortment $\{ \varepsilon_{si} \}_{i \in S_s}$, standard self-normalized inequalities cannot be directly applied.
    To address this,~\citet{perivier2022dynamic} introduced an analysis based on a global random vector $\zb_s := \sum_{i \in S_s} \varepsilon_{si} \nabla f_{\hat{\wb}_{s}}(\xb_{si}) $, which, combined with a Bernstein-type tail inequality for the logistic bandit (Theorem 1 in~\citet{faury2020improved}), yields a concentration bound of order $\tilde{\Ocal}\left( \sqrt{\lambda} + \frac{d_w \log T}{\sqrt{\lambda}} \right)$ (Theorem C.6 in~\citet{perivier2022dynamic}).
    As a result, achieving a regret bound of $\tilde{\Ocal}(\sqrt{T})$ requires the regularization parameter as $\lambda = \Ocal(d \log T)$.
    In contrast, by adapting techniques from~\citet{dani2008stochastic}, we derive a concentration result with a bound that is independent of $\lambda$ (up to logarithmic factors). 
    The proof of Lemma~\ref{sub_lemma:lambda-indep. concentration} is provided in Section~\ref{appx:technical lemma for confidence ball}.

    For the term $I_2$, we introduce several notations to facilitate understanding.
    Let $K_s = |S_s|$ denote the size of the assortment selected at round $s$.
    Then we can write $S_s = \{ i_1, \ldots, i_{K_s} \}$.
    Now we define $\Gb_s \in \RR^{d \times K_s}$ as the matrix whose $k$-th column is the gradient of $f(\xb_{s i_k})$ evaluated at $\hat{\wb}_s$, i.e.,
    
    \begin{equation*}
        \Gb_s := \left[ \nabla f_{\hat{\wb}_s}(\xb_{s i_1}), \ldots, \nabla f_{\hat{\wb}_s}(\xb_{s i_{K_s}}) \right] \in \RR^{d_w \times K_s} \, .    
    \end{equation*}
    
    For any vector $\ub := [u_1, \ldots, u_{K_s}]^\top \in \RR^{K_s}$, we define $\hb_s(\ub) := [h_1(\ub), \ldots, h_{K_s}(\ub)]^\top$, where each $h_k$ is given by:
    
    $$
    h_k(\ub) := \frac{\exp(u_k)}{1 + \sum_{j=1}^{K_s} \exp(u_j)}.
    $$
    
    In addition, we define:
    \begin{align*}
        & \hat{\ub}_s := [\hat{u}_{s i_1}, \ldots, \hat{u}_{s i_{K_s}}]^\top \in \RR^{K_s}  \quad \text{where} \quad \hat{u}_{s i_k}=f_{\hat{\wb}_s} (\xb_{s i_k}) + (\tilde{\wb}^* - \hat{\wb}_s)^\top \nabla f_{\hat{\wb}_s}(\xb_{s i_k}) \,,
        \\
        & \ub^*_s := [u^*_{s i_1}, \ldots, u^*_{s i_{K_s}}]^\top \in \RR^{K_s} \quad \text{where} \quad u^*_{s i_k}=f_{\tilde{\wb}^*} (\xb_{s i_k}) \, .
    \end{align*}

    Now we bound $I_2$ as follows:
    \begin{align*}
        I_2 
        & = 4 \left \| \sum_{s=t_0+1}^{t-1} \Gb_s (\hb_s(\hat{\ub}_s) - \hb_s(\ub^*_s)) \right \|^2_{\Vb_t^{-1}}
        \\
        & = 4 \left \| \sum_{s=t_0+1}^{t-1} \Gb_s \nabla \hb_s (\tilde{\ub}_s)^\top (\hat{\ub}_s - \ub^*_s) \right \|^2_{\Vb_t^{-1}}
        \\
        & \le 4 \left( \sum_{s=t_0+1}^{t-1} \left \| \Gb_s \nabla \hb_s(\tilde{\ub}_s)^\top (\hat{\ub}_s - \ub^*_s) \right \|_{\Vb_t^{-1}} \right)^2
        \\
        & \le 4 \left( \sum_{s=t_0+1}^{t-1} \| \Gb_s \|_{\Vb^{-1}_t} \| \nabla \hb_s(\tilde{\ub}_s)^\top (\hat{\ub}_s - \ub^*_s) \|_{2} \right)^2
    \end{align*}
    where the first inequality follows from the triangle inequality, and the second inequality uses the following bound: for any matrices $\Ab, \Bb, \Vb \in \RR^{m \times m}$ and vector $\xb \in \RR^m$,
    \begin{equation*}
        \| \Ab \Bb \xb \|^2_{\Vb} = (\Ab \Bb \xb)^\top \Vb (\Ab \Bb \xb) = (\Bb \xb)^\top \Ab^\top \Vb \Ab (\Bb \xb) \le \lambda_{\max}(\Ab^\top \Vb \Ab) \| \Bb \xb \|_2^2 = \| \Ab \|_{\Vb}^2 \| \Bb \xb \|_2^2 \, .
    \end{equation*}
    On the other hand, we can bound each $\| \nabla \hb_s(\tilde{\ub}_s)^\top (\hat{\ub}_s - \ub^*_s) \|_{2}$ as follows:
    \begin{align*}
        \| \nabla \hb_s(\tilde{\ub}_s)^\top (\hat{\ub}_s - \ub^*_s) \|_{2}
        & \le \| \nabla \hb_s(\tilde{\ub}_s)^\top \|_{\infty, 2} \| \hat{\ub}_s - \ub^*_s \|_\infty \, ,
    \end{align*}
    where the norm $\| \cdot \|_{\infty, 2}$ is defined as follows:
    \begin{equation*}
        \| \Ab \|_{\infty, 2} = \sup \{ \| \Ab \xb \|_2 : \| \xb \|_{\infty} \le 1 \} .
    \end{equation*}

    Since $\nabla \hb_s(\tilde{\ub}_s)^\top = \diag( \hb_s(\tilde{\ub}_s)) - \hb_s(\tilde{\ub}_s) \hb_s(\tilde{\ub}_s)^\top$, for any $\xb := [x_1, \ldots, x_{K_s}]^\top$ with $ \| \xb \|_\infty \le 1$, we have
    \begin{align*}
        \| \nabla \hb_s(\tilde{\ub}_s)^\top \xb \|_2^2 
        & = \left \| \diag(\hb_s(\tilde{\ub}_s)) \xb - (\hb_s(\tilde{\ub}_s)^\top \xb) \hb_s(\tilde{\ub}_s) \right \|_2^2
        \\
        & = \sum_{k=1}^{K_s} \left( h_k(\tilde{\ub}_s) x_k - \left( \sum_{j=1}^{K_s} x_j h_j(\tilde{\ub}_s) \right) h_k(\tilde{\ub}_s) \right)^2
        \\
        & = \sum_{k=1}^{K_s} h^2_k(\tilde{\ub}_s) \left(  x_k - \sum_{j=1}^{K_s} x_j h_j(\tilde{\ub}_s)  \right)^2   \, .
    \end{align*}
    If we denote $\sum_{j=1}^{K_s} h_j(\tilde{\ub}_s) =: q$, since $\| \xb \|_\infty \le 1$, for all $1 \le k \le K_s$, we have
    \begin{equation*}
        \left| x_k - \sum_{j=1}^{K_s} x_j h_j(\tilde{\ub}_s) \right| 
        \le 1 + \left| \sum_{j=1}^{K_s} x_j h_j(\tilde{\ub}_s) \right|
        \le 1 + q \, .
    \end{equation*}
    Then, $\| \nabla \hb_s(\tilde{\ub}_s)^\top \xb \|_2^2$ can be further bounded as
    \begin{align*}
        \| \nabla \hb_s(\tilde{\ub}_s)^\top \xb \|_2^2
        \le \sum_{i=1}^{K_s} h^2_i(\tilde{\ub}_s) (1 + q)^2
        \le \sum_{i=1}^{K_s} h_i(\tilde{\ub}_s) (1 + q)^2
        \le q (1 + q)^2  
        \le \max_{p \in (0,1)} p (1+p)^2  \le 4 \, ,
    \end{align*}
    where the fourth inequality holds because $0 < q < 1$.
    This implies $\| \nabla \hb_s(\tilde{\ub}_s)^\top \|_{\infty, 2} \le 2$ for all $s$, therefore we bound the term $I_2$ as follows:
    \begin{align*}
        I_2 & \le  16 \left( \sum_{s=t_0+1}^{t-1} \| \Gb_s \|_{\Vb^{-1}_t} \| \hat{\ub}_s - \ub^*_s \|_{\infty} \right)^2
        \\
        & = 16 \left( \sum_{s=t_0+1}^{t-1} \| \Gb_s \|_{\Vb^{-1}_t} \max_{1 \le k \le K_s} | \hat{u}_{s i_k} - u^*_{s i_k} | \right)^2
        \\
        & = 16 \left( \sum_{s=t_0+1}^{t-1} \| \Gb_s \|_{\Vb^{-1}_t} \max_{ i \in S_s} \frac{1}{2} \| \hat{\wb}_s - \tilde{\wb}^* \|^2_{\nabla^2 f_{\tilde{\wb}_s}(\xb_{si})} \right)^2
        \\
        & \le 4 C^2_h \left( \sum_{s=t_0+1}^{t-1} \| \Gb_s \|_{\Vb^{-1}_t} \| \hat{\wb}_s - \tilde{\wb}^* \|^2_2 \right)^2 \, , 
    \end{align*}
    where the third equality follows from the second-order Taylor’s theorem with $\tilde{\wb}_s$ lying between $\hat{\wb}_s$ and $\tilde{\wb}^*$, and the final inequality follows from Assumption~\ref{assm:boundedness}.
  
    For $I_3$, we have
    \begin{equation*}
        4 \lambda^2 \| \hat{\wb}_0 - \tilde{\wb}^* \|^2_{\Vb_t^{-1}} \le 4 \lambda \| \hat{\wb}_0 - \tilde{\wb}^* \|^2_{2} \le \frac{4 \lambda C \kappa^{-2} d_w C_f^2 \zeta}{\mu t_0} \, .
    \end{equation*}

    For next step, we introduce the following lemma, whose proof is provided in Section~\ref{appx:technical lemma for confidence ball}.
    \begin{lemma} \label{sub_lemma:bridge between w and g}
        For any $t_0 + 1 \le t \le T$, it holds:
        \begin{equation*}
            \| \hat{\wb}_t - \tilde{\wb}^* \|_{\Vb_t}^2 \le (1 + 6 \sqrt{2})^2 \kappa^{-1} \| \gb_t(\hat{\wb}_t) - \gb_t(\tilde{\wb}^*) \|_{\Vb_t^{-1}}^2 \, .
        \end{equation*}
    \end{lemma}

    Combining Lemma~\ref{sub_lemma:bridge between w and g} with the bounds of $I_1, I_2, I_3$, we have
    \begin{align*}
        \| \hat{\wb}_t - \tilde{\wb}^* \|_2^2 
        & \le \frac{(1 + 6\sqrt{2})^2 \kappa^{-1}}{\lambda} \| \gb_t(\hat{\wb}_t) - \gb_t(\tilde{\wb}^*) \|_{\Vb_t^{-1}}^2
        \\
        & \le \frac{4 (1 + 6\sqrt{2})^2 \kappa^{-1}}{\lambda}
            \left(
                \alpha_t
                + C^2_h \left( \sum_{s=t_0+1}^{t-1} \| \Gb_s \|_{\Vb^{-1}_t} \| \hat{\wb}_s - \tilde{\wb}^* \|^2_2 \right)^2
                + \frac{\lambda \kappa^{-2} d_w C_f^2 \zeta}{\mu t_0}
            \right)
        \\
        & = \frac{C_0 \kappa^{-1} \alpha_t}{\lambda}
            + \frac{C_0 C_h^2 \kappa^{-1} }{\lambda} \left( \sum_{s=t_0+1}^{t-1} \| \Gb_s \|_{\Vb^{-1}_t} \| \hat{\wb}_s - \tilde{\wb}^* \|^2_2 \right)^2
            + \frac{C_0 \kappa^{-3} d_w C_f^2 \zeta}{\mu t_0} \, , \numberthis \label{eq:induction eq}
    \end{align*}
    where we denote $C_0 := 4 (1 + 6\sqrt{2})^2$.
    We now proceed by induction to establish the convergence rate of $\| \hat{\wb}_t - \tilde{\wb}^* \|_2^2$ for every time step $t = t_0 +1, \ldots, T$.
    Recall that by Lemma~\ref{lemma:convergence of w_0}, with probability at least $1 - \delta$, we have
    \begin{equation*}
        \| \hat{\wb}_0 - \tilde{\wb}^* \|^2_{2} \le \frac{C \kappa^{-2} d_w C_f^2 \zeta}{\mu t_0} \, .
    \end{equation*}
    To formally set up the inductive argument, we assume that at round $\tau$, there exists a universal constant $\tilde{C}$ such that, with probability at least $1 - 2 \delta$, the following holds:
    \begin{equation*}
        \| \hat{\wb}_\tau - \tilde{\wb}^* \|^2_{2} \le \frac{\tilde{C} \kappa^{-2} d_w C_f^2 \zeta}{\mu t_0} \, .
    \end{equation*}
    Then, we want to show that at round $\tau+1$ with probability at least $1 - 2\delta$, 
    \begin{equation*}
        \| \hat{\wb}_{\tau+1} - \tilde{\wb}^* \|^2_{2} \le \frac{\tilde{C} \kappa^{-2} d_w C_f^2 \zeta}{\mu t_0} \, .
    \end{equation*}
    From Eq.~\eqref{eq:induction eq}, at round $\tau+1$, we have
    \begin{align*}
        & \| \hat{\wb}_{\tau+1} - \tilde{\wb}^* \|^2_{2}
        \\
        & \le \frac{C_0 \kappa^{-1} \alpha_{\tau+1}}{\lambda}
            + \frac{C_0 C_h^2 \kappa^{-1}}{\lambda} \left( \sum_{s=t_0 + 1}^{\tau} \| \Gb_s \|_{\Vb^{-1}_{\tau + 1}} \| \hat{\wb}_s - \tilde{\wb}^* \|^2_2 \right)^2
            + \frac{C_0 \kappa^{-3} d_w C_f^2 \zeta}{\mu t_0}
        \\
        & \le \frac{C_0 \kappa^{-1} \alpha_{\tau+1}}{\lambda}
            + \frac{C_0 C_h^2 \tilde{C}^2 \kappa^{-5} d_w^2 C_f^4 \zeta^2}{\lambda \mu^2 t_0^2}  \left( \sum_{s=t_0 + 1}^{\tau}  \| \Gb_s \|_{\Vb^{-1}_{\tau + 1}}   \right)^2 
            + \frac{C_0 \kappa^{-3} d_w C_f^2 \zeta}{\mu t_0}
        \\
        & \le \frac{C_0 \kappa^{-1} \alpha_{\tau+1}}{\lambda}
            + \frac{C_0 C_h^2 \tilde{C}^2 \kappa^{-5} d_w^2 C_f^4 \zeta^2}{\lambda \mu^2 t_0^2} \left( \sqrt{ \sum_{s=t_0 + 1}^\tau } \sqrt{ \sum_{s=t_0 + 1}^{\tau} \| \Gb_s \|^2_{\Vb^{-1}_{\tau + 1}}} \right)^2
            + \frac{C_0 \kappa^{-3} d_w C_f^2 \zeta}{\mu t_0}
        \\
        & \le \frac{C_0 \kappa^{-1} \alpha_{\tau+1}}{\lambda}
            + \frac{C_0 C_h^2 \tilde{C}^2 \kappa^{-5} d_w^2 C_f^4 \zeta^2 \tau}{\lambda \mu^2 t_0^2} \left(  \sum_{s=t_0 + 1}^{\tau} \| \Gb_s \|^2_{\Vb^{-1}_{\tau + 1}} \right)
            + \frac{C_0 \kappa^{-3} d_w C_f^2 \zeta}{\mu t_0}
        \\
        & \le \frac{C_0 \kappa^{-1} \alpha_{\tau+1}}{\lambda}
            + \frac{C_0 C_h^2 \tilde{C}^2 \kappa^{-5} d_w^3 C_f^4 \zeta^2 \tau}{\lambda \mu^2 t_0^2} 
            + \frac{C_0 \kappa^{-3} d_w C_f^2 \zeta}{\mu t_0} \, , \numberthis \label{eq:induction eq 3}
    \end{align*}
    where the last inequality invokes the following lemma, with its proof provided in Section~\ref{appx:technical lemma for confidence ball}.
    \begin{lemma}[] \label{sub_lemma:big elliptical}
        For all $t_0 + 1 \le t \le T$, 
        \begin{equation*}
            \sum_{s=t_0+1}^{t-1} \| \Gb_s \|^2_{\Vb_t^{-1}} \le d_w \, .
        \end{equation*}
    \end{lemma}

    Now, our goal is to show that there exist a universal constant $\tilde{C}$ such that the following bound holds:
    \begin{equation} \label{eq:induction eq 1}
        \frac{C_0 \kappa^{-1} \alpha_{\tau+1}}{\lambda}
            + \frac{C_0 C_h^2 \tilde{C}^2 \kappa^{-5} d_w^3 C_f^4 \zeta^2 \tau}{\lambda \mu^2 t_0^2} 
            + \frac{C_0 \kappa^{-3} d_w C_f^2 \zeta}{\mu t_0} 
        \le \frac{\tilde{C} \kappa^{-2} d_w C_f^2 \zeta}{\mu t_0}  \, .
    \end{equation}
    For notational simplicity, let us denote $\alpha_t := \tilde{C}_\alpha d_w \zeta_1$, where $\zeta_1$ denotes a logarithmic term that depends on $T, d_w, \lambda,$ and $\delta$.
    Then, Eq.~\eqref{eq:induction eq 1} can be rewritten as:
    \begin{equation} \label{eq:induction eq 2}
        \frac{C_0 \tilde{C}_{\alpha} \kappa^{-1} d_w \zeta_1}{\lambda}
            + \frac{C_0 C_h^2 \tilde{C}^2 \kappa^{-5} d_w^3 C_f^4 \zeta^2 \tau}{\lambda \mu^2 t_0^2} 
            + \frac{C_0 \kappa^{-3} d_w C_f^2 \zeta}{\mu t_0} 
        \le \frac{\tilde{C} \kappa^{-2} d_w C_f^2 \zeta}{\mu t_0}  \, .        
    \end{equation}
    Note that the left-hand side of Eq.~\eqref{eq:induction eq 2} is monotonically increasing in $\tau$, and therefore the inequality must also hold at $\tau = T$; that is,
    \begin{align*}
        & \frac{C_0 \tilde{C}_{\alpha} \kappa^{-1} d_w \zeta_1}{\lambda}
            + \frac{C_0 C_h^2 \tilde{C}^2 \kappa^{-5} d_w^3 C_f^4 \zeta^2 T}{\lambda \mu^2 t_0^2} 
            + \frac{C_0 \kappa^{-3} d_w C_f^2 \zeta}{\mu t_0} 
        \le \frac{\tilde{C} \kappa^{-2} d_w C_f^2 \zeta}{\mu t_0} 
        \\
        & \iff \left( \frac{C_0 C_h^2 \kappa^{-5} d_w^3 C_f^4 \zeta^2 T}{\lambda \mu^2 t_0^2} \right) \tilde{C}^2
            - \left( \frac{ \kappa^{-2} d_w C_f^2 \zeta}{\mu t_0} \right) \tilde{C}
            + \left( \frac{C_0 \kappa^{-3} d_w C_f^2 \zeta}{\mu t_0} + \frac{C_0 \tilde{C}_{\alpha} \kappa^{-1} d_w \zeta_1}{\lambda} \right) \le 0 \, .
    \end{align*}
    A feasible choice of $\tilde{C}$ must satisfy the following condition:
    \begin{align*}
        \left( \frac{ \kappa^{-2} d_w C_f^2 \zeta}{\mu t_0} \right)^2
            - 4 \left( \frac{C_0 C_h^2 \kappa^{-5} d_w^3 C_f^4 \zeta^2 T}{\lambda \mu^2 t_0^2} \right)
            \left( \frac{C_0 \kappa^{-3} d_w C_f^2 \zeta}{\mu t_0} + \frac{C_0 \tilde{C}_{\alpha} \kappa^{-1} d_w \zeta_1}{\lambda} \right)
            \ge 0 \, .        
    \end{align*}

    By substituting $\lambda = C_{\lambda} \sqrt{T}$ and $t_0 = \kappa^{-3/2} d_w \sqrt{T}$, this condition becomes:
    \begin{align*}
        & \left( \frac{ \kappa^{-1/2} C_f^2 \zeta}{\mu \sqrt{T}} \right)^2
            - 4 \left( \frac{C_0 C_h^2 \kappa^{-2} d_w C_f^4 \zeta^2}{C_{\lambda} \sqrt{T} \mu^2} \right)
            \left( \frac{C_0 \kappa^{-3/2} C_f^2 \zeta}{\mu \sqrt{T}} + \frac{C_0 \tilde{C}_{\alpha} \kappa^{-1} d_w \zeta_1}{C_{\lambda} \sqrt{T}} \right)
            \ge 0
        \\        
        & \iff
        (\kappa^{-1} C_f^4 \zeta^2 \mu) C_{\lambda}^2 - (4 C_0^2 C_h^2 \kappa^{-7/2} d_w C_f^6 \zeta^3) C_{\lambda} - (4 C_0^2 C_h^2 \tilde{C}_{\alpha} \kappa^{-3} d_w^2 C_f^4 \zeta^2 \zeta_1 \mu) \ge 0 \, .
    \end{align*}    
    Solving the above quadratic inequality for $C_\lambda$, it suffices to choose
    \begin{equation*}
        C_\lambda \ge \frac{2 C_0 C_h d_w \kappa^{-5/2} \left(C_0 C_f^2 C_h \zeta + \sqrt{C_0^2 C_f^4 C_h^2 \kappa^4 \zeta^2 + \tilde{C}_{\alpha} \kappa^7 \mu^2 \zeta_1} \right)}{\mu} = \tilde{\Ocal}(\kappa^{-5/2} \mu^{-1} d_w) \, .
    \end{equation*}    
    Note that $C_\lambda$ depends only logarithmically on $T$, and this choice guarantees that Eq.\eqref{eq:induction eq 2} holds, thereby confirming the existence of a universal constant $\tilde{C}$.
    Hence, by induction, we conclude that for all $t_0+1 \le t \le T$, there exists a universal constant $\tilde{C}$ such that, with probability at least $1 - 2\delta$, the desired bound holds:
    \begin{equation*}
        \| \hat{\wb}_t - \tilde{\wb}^* \|_2^2 \le \frac{\tilde{C} \kappa^{-2} d_w C_f^2 \zeta}{\mu t_0} \, .
    \end{equation*}

    Finally, combining again the results of Lemma~\ref{sub_lemma:bridge between w and g} and similar argument used in Eq.~\eqref{eq:induction eq 3}, we have
    \begin{align*}
        & \| \hat{\wb}_t - \tilde{\wb}^* \|_{\Vb_t}^2
        \\
        & \le (1 + 6 \sqrt{2})^2 \kappa^{-1} \| \gb_t (\hat{\wb}_t) - \gb_t (\tilde{\wb}^*) \|_{\Vb_t^{-1}}^2
        \\
        & \le 4(1 + 6 \sqrt{2})^2 \kappa^{-1} 
            \left(
                \alpha_t
                + C^2_h \left( \sum_{s=t_0+1}^{t-1} \| \Gb_s \|_{\Vb^{-1}_t} \| \hat{\wb}_s - \tilde{\wb}^* \|^2_2 \right)^2
                + \frac{\lambda \kappa^{-2} d_w C_f^2 \zeta}{\mu t_0}
            \right)
        \\
        & \le C_0 \kappa^{-1} \bigg( \alpha_t
            + \frac{C_h^2 \tilde{C}^2 \kappa^{-4} d_w^2 C_f^4 \zeta^2}{\mu^2 t_0^2} \left( \sum_{s=t_0 + 1}^{t-1} \| \Gb_s \|_{\Vb^{-1}_t} \right)^2
            + \frac{\lambda \kappa^{-2} d_w C_f^2 \zeta}{\mu t_0}
            \bigg)
        \\
        & \le C_0 \kappa^{-1} \bigg( \alpha_t
            + \frac{C_h^2 \tilde{C}^2 \kappa^{-4} d_w^2 C_f^4 \zeta^2}{\mu^2 t_0^2}  \left( \sqrt{ \sum_{s=t_0+1}^{t-1} } \sqrt{ \sum_{s=t_0+1}^{t-1} \| \Gb_s \|^2_{\Vb^{-1}_{t}}} \right)^2
            + \frac{\lambda \kappa^{-2} d_w C_f^2 \zeta}{\mu t_0}
            \bigg)
        \\
        & \le C_0 \kappa^{-1} \bigg( \alpha_t
            + \frac{C_h^2 \tilde{C}^2 \kappa^{-4} d_w^2 C_f^4 \zeta^2 t}{\mu^2 t_0^2}  \left( \sum_{s=t_0+1}^{t-1} \| \Gb_s \|^2_{\Vb^{-1}_{t}} \right)
            + \frac{\lambda \kappa^{-2} d_w C_f^2 \zeta}{\mu t_0}
            \bigg)
        \\
        & \le C_0 \kappa^{-1} \bigg( \alpha_t
            + \frac{C_h^2 \tilde{C}^2 \kappa^{-4} d_w^3 C_f^4 \zeta^2 t}{\mu^2 t_0^2} 
            + \frac{\lambda \kappa^{-2} d_w C_f^2 \zeta}{\mu t_0}
            \bigg)
        \\
        & = \tilde{\Ocal} \left( \mu^{-2} \kappa^{-4} d_w \right) \, ,            
    \end{align*}
    where, in the last inequality, we substitute $\alpha_t = \tilde{\Ocal}(d_w)$, $\lambda = C_\lambda \sqrt{T} = \tilde{\Ocal}(\kappa^{-5/2} \mu^{-1} d_w \sqrt{T})$, $t_0= \lceil d_w \kappa^{-3/2} \sqrt{T} \rceil$.     
\end{proof}

\subsection{Proofs of Technical Lemmas} \label{appx:technical lemma for confidence ball}
\subsubsection{Proof of Lemma~\ref{sub_lemma:lambda-indep. concentration}}
\begin{proof}[Proof of Lemma~\ref{sub_lemma:lambda-indep. concentration}]
    If we denote $\sbb_t = \sum_{s=t_0+1}^{t-1} \sum_{i \in S_s} \left( y_{si} - p_{si}(\tilde{\wb}^*) \right) \nabla f_{\hat{\wb}_s}(\xb_{si})$, 
    then,
    \begin{equation*}
        \sbb_{t+1} = \sbb_t + \sum_{i \in S_t} (y_{ti} - p_{ti}(\tilde{\wb}^*)) \nabla f_{\hat{\wb}_t}(\xb_{ti}) = \sbb_t + \phib_t \, , 
    \end{equation*}
    where we denote $\phib_t := \sum_{i \in S_t} (y_{ti} - p_{ti}(\tilde{\wb}^*)) \nabla f_{\hat{\wb}_t}(\xb_{ti})$.
    Then, we have
    \begin{align*}
        \| \sbb_{t+1} \|_{\Vb_{t+1}^{-1}}^2
        & = (\sbb_{t+1} + \phib_t)^\top \Vb_{t+1}^{-1} (\sbb_{t+1} + \phib_t)
        \\
        & = \sbb_{t+1}^\top \Vb_{t+1}^{-1} \sbb_{t+1} + 2 \sbb_{t+1}^\top \Vb_{t+1}^{-1} \phib_t + \phib_{t}^\top \Vb_{t+1}^{-1} \phib_{t}
        \\
        & \le \sbb_{t+1}^\top \Vb_{t}^{-1} \sbb_{t+1} + 2 \sbb_{t+1}^\top \Vb_{t+1}^{-1} \phib_t + \phib_{t}^\top \Vb_{t}^{-1} \phib_{t} \, ,
    \end{align*}
    where the last inequality follows by $\Vb_{t} \preceq \Vb_{t+1}$.
    Using this argument recursively, we have
    \begin{equation} \label{eq:recursive eq}
        \| \sbb_t \|_{\Vb_t^{-1}}^2 \le \sum_{s=t_0+1}^{t-1} \phib_s^\top \Vb_{s}^{-1} \phib_s + \sum_{s=t_0+1}^{t-1} 2 \phib_s \Vb_{s+1}^{-1} \sbb_s \, .
    \end{equation}

    Now we will define a martingale difference sequence. 
    We start to define an event $E_t = \ind \left\{ \| \sbb_s \|_{\Vb_{s}^{-1}}^2 \le \beta_s, \forall s \le t \right\}$, and define
    \begin{equation*}
        J_t = 2 E_t \phib_t^\top \Vb_{t+1}^{-1} \sbb_t \, .
    \end{equation*}
    Then, we have $\EE[J_t \mid \Fcal_t] = 2 \EE[\phib_t^\top \mid \Fcal_t] \Vb_{t+1}^{-1} \sbb_t = 0$.
    And since we have
    \begin{align*}
        | J_t | 
        & \le 2 E_t | \phib_t^\top \Vb_{t+1}^{-1} \sbb_t |
        \\
        & \le 2 E_t \| \phib_t^\top \Vb_{t+1}^{-1/2} \|_2 \| \Vb_{t+1}^{-1/2} \sbb_t \|_2
        \\
        & = 2 E_t \sqrt{\phib_t^\top \Vb_{t+1}^{-1} \phib_t} \sqrt{\sbb_t^\top \Vb_{t+1}^{-1} \sbb_t}
        \\
        & \le 2 E_t \sqrt{\phib_t^\top \Vb_{t}^{-1} \phib_t} \sqrt{\sbb_t^\top \Vb_{t}^{-1} \sbb_t}
        \\
        & \le 2 E_t \sqrt{\alpha_t} \| \phib_t\|_{\Vb_{t}^{-1}}
    \end{align*}
    where the last inequality follows trivially when $E_t = 0$, and by definition of $E_t$ when $E_t = 1$.
    Additionally this gives us a family of uniform upper bound:
    \begin{equation*}
        | J_s | \le 2 \sqrt{\alpha_t}, \quad \forall s \le t \, .
    \end{equation*}
        
    Now we bound the conditional variance of $J_t$ as follows:
    \begin{align*}
        V_t := \sum_{s=t_0+1}^{t-1} \mathrm{Var}(J_s \mid J_0, \ldots, J_{s-1})
        & \le \sum_{s=t_0+1}^{t-1} 4 E_s \alpha_s \| \phib_s\|^2_{\Vb_{s}^{-1}} 
        \\
        & \le 4 (\max_{s \le t} \alpha_s) \sum_{s=t_0+1}^{t-1} \| \phib_s\|^2_{\Vb_{s}^{-1}}
        \\
        & \le 16 \alpha_t \sum_{s=t_0+1}^{t-1} \max_{i \in S_s} \| \nabla f_{\wb_s}(\xb_{si}) \|_{\Vb_s^{-1}}^2
    \end{align*}
    where the last inequality uses the following inequality
    \begin{align*}
        \| \phib_s \|_{\Vb_s^{-1}}
        & \le  \sum_{i \in S_s} | \varepsilon_{si} | \| \nabla f_{\wb_s}(\xb_{si}) \|_{\Vb_s^{-1}} 
        \\
        & \le \left( \sum_{i \in S_s} | y_{si} | \| \nabla f_{\wb_s}(\xb_{si}) \|_{\Vb_s^{-1}}
            + \sum_{i \in S_s} | p_{si}(\tilde{\wb}^*) | \| \nabla f_{\wb_s}(\xb_{si}) \|_{\Vb_s^{-1}} \right)
        \\
        & \le 2 \max_{i \in S_s} \| \nabla f_{\wb_s}(\xb_{si}) \|_{\Vb_s^{-1}} \, .
    \end{align*}
    For the next step, we introduce the following lemma:
    \begin{lemma} \label{sub_lemma:gradient elliptical}
        For any $t \ge t_0+1$, it holds that
        \begin{equation*}
            \sum_{s=t_0+1}^{t-1} \min \left\{ 1,  \max_{i \in S_s} \| \nabla f_{\hat{\wb}_s}(\xb_{si}) \|_{\Vb_s^{-1}}^2 \right\}
            \le 2 d_w \log \left( 1 + \frac{t C_g^2}{d_w \lambda} \right) \, .
        \end{equation*}
    \end{lemma}
    Lemma~\ref{sub_lemma:gradient elliptical} is conceptually similar to the elliptical potential lemma commonly used in the linear utility setting~\cite{oh2019thompson, oh2021multinomial, lee2024nearly}.
    As a result, the proof of Lemma~\ref{sub_lemma:gradient elliptical} can be readily extended from its linear counterpart. 
    For completeness, we provide the proof of Lemma~\ref{sub_lemma:gradient elliptical} at the end of this section.
    Suppose that $\lambda \ge C_g^2$, which guarantees that $\| \nabla f_{\hat{\wb}_t}(\xb_{ti}) \|_{\Vb_t^{-1}}^2 \le 1$ for all $t \ge t_0+1$ and $i \in [N]$.
    By using Lemma~\ref{sub_lemma:gradient elliptical}, we bound $V_t$ as follows:  
    \begin{equation*}
        V_t \le 32 \alpha_t d_w \log (1 + T C_g^2/(d_w \lambda)) =: v_t \, .
    \end{equation*}   
    
    Now we have established the sum of conditional variance of martingale difference sequence, denoted by $V_t$, is bounded by $v_t = \tilde{\Ocal}(\alpha_t d_w)$. 
    Then, by Freedman's inequality (Lemma~\ref{aux_lemma:Freedman}) with parameters $a = \alpha_t/2, b = 2 \sqrt{\alpha_t}$, and $v = v_t$, we have
    \begin{align*}
        \PP \left( \sum_{s=t_0+1}^{t-1} J_s \ge \alpha_t / 2 \right)
        & = \PP \left( \sum_{s=t_0+1}^{t-1} J_s \ge \alpha_t / 2, V_t \le v_t \right)
        \\
        & \le \exp \left( \frac{-(\alpha_t/2)^2}{2v_t + 2(\alpha_t/2)(2\sqrt{\alpha_t})/3} \right)
        \\
        & \le \exp \left( \max \left\{ -\frac{\alpha_t^2}{16 v_t} \right\}, \left\{ - \frac{3 \sqrt{\alpha_t}}{16}
        \right\} \right)
        \\
        & \le \frac{\delta}{t^2 \pi^2}
    \end{align*}
    where the last inequality holds if we set
    \begin{equation} \label{eq:beta condition 1}
        \alpha_t \ge C_\beta d_w \log(1 + T C_g^2/(d_w \lambda)) \log(t^2 \pi^2/\delta)
    \end{equation}
    for some absolute constant $C_\beta > 0$ satisfying that
    \begin{equation*}
        \alpha_t^2 \ge 16 v_t \log(t^2 \pi^2/\delta) + (16/3)^2 \log^2 (t^2 \pi^2 / \delta) \, .
    \end{equation*}

    By taking union bound,
    \begin{align*}
        \PP \left( \exists t, \quad \sum_{s=t_0+1}^{t-1} J_s \ge \frac{\alpha_t}{2} \right) 
        & \le \sum_{t=t_0+1}^{\infty} \PP \left( \sum_{s=t_0+1}^{t-1} J_s \ge \frac{\alpha_t}{2} \right) 
        \\
        & \le \sum_{t=t_0+1}^{\infty} \PP \left( \sum_{s=t_0+1}^{t-1} J_s \ge \frac{\alpha_t}{2} \right) 
        \\
        & \le \sum_{t=t_0+1}^{\infty} \frac{\delta}{t^2 \pi^2} = \frac{\pi^2}{\delta} \cdot \frac{\pi^2}{6} = \delta \, .
    \end{align*}
    Until now, we have established the event $\{ \forall t\ge t_0+1, \sum_{s=t_0+1}^{t-1} J_s \le \alpha_t/2 \}$ holds with probability at least $1 - \delta$. 
    Now we will show that the event $E_t = 1$ under the event $\{ \forall t\ge t_0+1, \sum_{s=t_0+1}^{t-1} J_s \le \alpha_t/2 \}$.
    We prove this by induction on $t$.
    Recall that $E_t = \ind \{ \| \sbb_s \|_{\Vb_s^{-1}}^2 \le \alpha_s, \forall t_0+1 \le s \le t \}$.
    For the base case ($t = t_0+1$), by definition of $\sbb_{t_0+1} = \zero_{d_w}$, it holds that $\| \sbb_{t_0+1} \|_{\Vb_{t_0 +1}^{-1}}^2 \le \alpha_{t_0+1}$.
    For the induction step, suppose that $E_s = 1$ for $t_0+1 \le s < t$. By Eq.~\eqref{eq:recursive eq}, 
    \begin{align*}
        \| \sbb_t \|_{\Vb_t^{-1}}^2 
        & \le \sum_{s=t_0+1}^{t-1} \phib_s^\top \Vb_{s}^{-1} \phib_s + \sum_{s=t_0+1}^{t-1} 2 \phib_s \Vb_{s+1}^{-1} \sbb_s
        \\
        & \le \sum_{s=t_0+1}^{t-1} \min \{1, \| \phib_s \|_{\Vb_s^{-1}}^2 \} + \frac{\alpha_t}{2}
        \\
        & \le 2 d_w \log (1 + T C_g^2/(d_w \lambda)) + \frac{\alpha_t}{2} \, ,
    \end{align*}
    where we invoke Lemma~\ref{sub_lemma:gradient elliptical} again for the second inequality.
    Then, with Eq.~\eqref{eq:beta condition 1}, if we have
    \begin{equation*}
        \alpha_t = \tilde{C}_\alpha d_w \log(1 + T C_g^2/(d_w \lambda)) \log(t^2 \pi^2/\delta)
    \end{equation*}
    for sufficiently large constant $\tilde{C}_\alpha >0$, then we have $\| \sbb_t \|_{\Vb_t^{-1}}^2 \le \alpha_t$. 
    This completes the proof. 
\end{proof}

\subsubsection{Proof of Lemma~\ref{sub_lemma:bridge between w and g}}
\begin{proof}[Proof of Lemma~\ref{sub_lemma:bridge between w and g}]
    For any $\wb_1, \wb_2 \in \Wcal$, by the mean value theorem, we have
    \begin{equation*}
        \gb_t (\wb_1) - \gb_t (\wb_2) = \nabla \ell_t (\wb_2) - \nabla \ell_t (\wb_1) = \left( \int_{0}^1 \nabla^2 \ell_t (\wb_1 + z (\wb_2 - \wb_1)) dx \right) (\wb_2 - \wb_1) \, .
    \end{equation*}
    This implies
    \begin{equation} \label{eq:bridge eq 1}
        \| \gb_t (\wb_1) - \gb_t (\wb_2) \|_{\Gb^{-1}_t(\wb_1, \wb_2)} = \| \wb_1 - \wb_2 \|_{\Gb_t(\wb_1, \wb_2)} \, ,
    \end{equation}
    where we denote $\Gb_t(\wb_1, \wb_2) := \int_{0}^1 \nabla^2 \ell_t (\wb_1 + z (\wb_2 - \wb_1)) dz$.
    On the other hand,~\citet{lee2024nearly} show that the multinomial logistic loss with linear utility is $3\sqrt{2}$-self-concordant-like (Lemma~\ref{aux_lemma:self-concordant-like in lee2024nearly}), then by the property of self-concordant-like function (Lemma~\ref{aux_lemma:self-concordence}), we have
    \begin{align*}
        \nabla^2 \ell_t(\wb_1 - z(\wb_2 - \wb_1)) \succeq \exp \left( -3 \sqrt{2} \| z(\wb_2 - \wb_1) \|_2 \right) \nabla^2 \ell_t(\wb_1) \, .
    \end{align*} 
    This implies
    \begin{align*}
        \Gb_t(\wb_1, \wb_2) 
        & \succeq \int_{0}^1 \exp \left( -3 \sqrt{2} \| z(\wb_2 - \wb_1) \|_2 \right) dz \nabla^2 \ell_t(\wb_1)
        \\
        & \succeq \frac{1 - \exp \left( \| \wb_2 - \wb_1 \|_2 \right)}{3 \sqrt{2} \| \wb_2 - \wb_1 \|_2} \nabla^2 \ell_t(\wb_1)
        \\
        & \succeq \frac{1 }{1 + 3 \sqrt{2} \|\wb_2 - \wb_1 \|_2} \nabla^2 \ell_t(\wb_1) 
        \\
        & \succeq \frac{1 }{1 + 6 \sqrt{2}} \nabla^2 \ell_t(\wb_1) \, ,
    \end{align*}
    where the third inequality uses the inequality $\frac{1 - \exp(-x)}{x} \ge \frac{1}{1+x}$ for $x \ge 0$.
    By applying this to Eq.~\eqref{eq:bridge eq 1}, we obtain
    \begin{align*}
        \| \wb_1 - \wb_2 \|^2_{\nabla^2 \ell_t(\wb_1)} 
        & \le (1 + 6 \sqrt{2}) \| \wb_1 - \wb_2 \|^2_{\Gb_t(\wb_1, \wb_2)}
        \\
        & \le (1 + 6 \sqrt{2}) \| \gb_t (\wb_1) - \gb_t (\wb_2) \|^2_{\Gb^{-1}_t(\wb_1, \wb_2)}
        \\
        & \le (1 + 6 \sqrt{2})^2 \| \gb_t (\wb_1) - \gb_t (\wb_2) \|^2_{\left(\nabla^2 \ell_t (\wb_1)\right)^{-1}} \, .
        \numberthis \label{eq:bridge eq 2}
    \end{align*}

    On the other hand, since
    \begin{align*}
        \nabla^2 \ell_t (\wb) 
        & := \lambda \Ib_{d_w} + \sum_{s=t_0+1}^{t-1} \left[ \sum_{i \in S_s} \hat{p}_{si}(\wb) \nabla f_{si} \nabla f_{si}^\top - \left(  \sum_{i \in S_s} \hat{p}_{si}(\wb) \nabla f_{si} \right) \left(  \sum_{i \in S_s} \hat{p}_{si}(\wb) \nabla f_{si} \right)^\top \right]
        \\
        & \succeq \lambda \Ib_{d_w} + \kappa \sum_{s=t_0+1}^{t-1} \sum_{i \in S_s} \nabla f_{si} \nabla f_{si}^\top
            \succeq \kappa \Vb_t \, ,
    \end{align*}
    where we abbreviate $\nabla f_{\hat{\wb}_s}(\xb_{si}) =: \nabla f_{si}$.
    Then, by applying this to Eq.~\eqref{eq:bridge eq 2} we have
    \begin{equation*}
        \| \wb_1 - \wb_2 \|_{\Vb_t}^2 \le (1 + 6 \sqrt{2})^2 \kappa^{-1} \| \gb_t (\wb_1) - \gb_t (\wb_2) \|^2_{\Vb_t^{-1}} \, .
    \end{equation*}
    Substituting $\wb_1 = \hat{\wb}_t$ and $\wb_2 = \tilde{\wb}^*$ completes the proof.
\end{proof}

\subsubsection{Proof of Lemma~\ref{sub_lemma:big elliptical}}
\begin{proof}[Proof of Lemma~\ref{sub_lemma:big elliptical}]
    Note that
    \begin{align*}
        \sum_{s=t_0+1}^{t-1} \| \Gb_s \|_{\Vb_t^{-1}}^2 
        &
        = \sum_{s=t_0+1}^{t-1} \lambda_{\max}(\Gb_s^\top \Vb_t^{-1} \Gb_s)
        \\
        & \le \sum_{s=t_0+1}^{t-1} \tr (\Gb_s^\top \Vb_t^{-1} \Gb_s)
        \\
        & = \sum_{s=t_0+1}^{t-1} \sum_{i \in S_s} \nabla f_{\hat{\wb}_s}(\xb_{si})^\top \Vb^{-1}_{t} \nabla f_{\hat{\wb}_s}(\xb_{si})
        \\
        & = \sum_{s=t_0+1}^{t-1} \sum_{i \in S_s} \tr \left( \nabla f_{\hat{\wb}_s}(\xb_{si})^\top \Vb^{-1}_{t} \nabla f_{\hat{\wb}_s}(\xb_{si}) \right)
        \\
        & = \tr \left( \Vb^{-1}_{t} \sum_{s=t_0+1}^{t-1} \sum_{i \in S_s} \nabla f_{\hat{\wb}_s}(\xb_{si}) \nabla f_{\hat{\wb}_s}(\xb_{si})^\top \right) \, .
    \end{align*}
    If we denote $\lambda_1, \ldots, \lambda_{d_w}$ the eigenvalues of $\sum_{s=t_0+1}^{t-1} \sum_{i \in S_s} \nabla f_{\hat{\wb}_s} (\xb_{si}) \nabla f_{\hat{\wb}_s} (\xb_{si})^\top$, then we have
    \begin{align*}
        \tr \left( \Vb^{-1}_{t} \sum_{s=t_0+1}^{t-1} \sum_{i \in S_s} \nabla f_{\hat{\wb}_s}(\xb_{si}) \nabla f_{\hat{\wb}_s}(\xb_{si})^\top \right) 
        & = \sum_{j=1}^{d_w} \lambda_j / (\lambda_j + \lambda)
        \le d_w \, .
    \end{align*}
\end{proof}

\subsubsection{Proof of Lemma~\ref{sub_lemma:gradient elliptical}} \label{sec:proof of sub_lemma:gradient elliptical}
\begin{proof}[Proof of Lemma~\ref{sub_lemma:gradient elliptical}]
    For notational simplicity, we abbreviate $\nabla f_{\hat{\wb}_t} (\xb_{ti})$ as $\nabla f_{ti}$.
    Note that
    \begin{align*}
        \Vb_{t+1} = \Vb_t + \sum_{i \in S_t} \nabla f_{ti} \nabla f_{ti}^\top \, ,
    \end{align*}
    which results in
    \begin{align*}
        \det \Vb_{t+1} = \det \Vb_{t} \left( 1 + \sum_{i \in S_{t}} \| \nabla f_{ti} \|_{\Vb_t^{-1}}^2 \right)
        & \ge \det \Vb_t \left( 1 + \max_{i \in S_t} \| \nabla f_{ti} \|_{\Vb_t^{-1}}^2 \right)
        \\
        & \ge \det (\lambda \Ib_{d_w}) \prod_{s=t_0+1}^t \left( 1 + \max_{i \in S_s} \| \nabla f_{si} \|_{\Vb_s^{-1}}^2 \right) \, .
    \end{align*}
    Then, we have
    \begin{align*}
        \sum_{s=t_0+1}^t \left( 1 + \max_{i \in S_s} \| \nabla f_{si} \|_{\Vb_s^{-1}}^2 \right) \le \log \frac{\det \Vb_{t+1}}{\det (\lambda \Ib_{d_w})} \, .
    \end{align*}
    Using the inequality $z \le 2 \log (1 + z)$ for $z \in [0,1]$, we have
    \begin{align*}
        \sum_{s=t_0+1}^t \min \left \{ 1, \max_{i \in S_s} \| \nabla f_{si} \|_{\Vb_s^{-1}}^2 \right\}
        & \le \sum_{s=t_0+1}^t 2 \log \left(1 +  \max_{i \in S_s} \| \nabla f_{si} \|_{\Vb_s^{-1}}^2 \right)
        \\
        & \le 2 \log \frac{\det \Vb_{t+1}}{\det (\lambda \Ib_{d_w})}
        \\ 
        & \le 2 d_w \log \left( 1 + \frac{(t+1) C_g^2}{d_w \lambda}\right) \, ,
    \end{align*}
    where the last inequality follows by the determinant-trace inequality (Lemma~\ref{aux_lemma:det-trace ineq.}). 
\end{proof}

\section{Proof of Lemma~\ref{lemma:optimistic utility}}
\begin{proof}[Proof of Lemma~\ref{lemma:optimistic utility}]
    Note that
    \begin{align*}
        z_{ti} - f_{\wb^*}(\xb_{ti}) 
        & = f_{\hat{\wb}_{t}}(\xb_{ti}) - f_{\wb^*}(\xb_{ti}) + \sqrt{\beta_t} \|\nabla f_{\hat{\wb}_{t}}(\xb_{ti})\|_{\Vb_{t}^{-1}} + \frac{\beta_t C_h}{\lambda}
        \\
        & = (\hat{\wb}_{t} - \wb^*)^\top \nabla f_{\dot{\wb}}(\xb_{ti}) + \sqrt{\beta_t} \|\nabla f_{\hat{\wb}_{t}}(\xb_{ti})\|_{\Vb_{t}^{-1}} + \frac{\beta_t C_h}{\lambda} \numberthis \label{eq:optimistic eq 1}
        \\
        & = (\hat{\wb}_{t} - \wb^*)^\top \nabla f_{\hat{\wb}_{t}}(\xb_{ti}) + (\hat{\wb}_{t} - \wb^*)^\top (\nabla f_{\dot{\wb}}(\xb_{ti}) - \nabla f_{\hat{\wb}_{t}}(\xb_{ti}))
        \\
        & \quad + \sqrt{\beta_t} \|\nabla f_{\hat{\wb}_{t}}(\xb_{ti})\|_{\Vb_{t}^{-1}} + \frac{\beta_t C_h}{\lambda}        
        \\
        & \ge (\hat{\wb}_{t} - \wb^*)^\top (\nabla f_{\dot{\wb}}(\xb_{ti}) - \nabla f_{\hat{\wb}_{t}}(\xb_{ti}))
            + \frac{\beta_t C_h}{\lambda}        
        \\
        & = (\hat{\wb}_{t} - \wb^*)^\top \nabla^2 f_{\ddot{\wb}}(\xb_{ti}) (\dot{\wb} - \hat{\wb}_{t}) + \frac{\beta_t C_h}{\lambda} \numberthis \label{eq:optimistic eq 2}
        \\
        & \ge - \| \hat{\wb}_{t} - \wb^* \|_{\Vb_{t}} \| \Vb_t^{-1/2} \nabla^2 f_{\ddot{\wb}}(\xb_{ti}) \Vb_t^{-1/2} \| \| \dot{\wb} - \hat{\wb}_{t} \|_{\Vb_{t}} + \frac{\beta_t C_h}{\lambda}
        \\
        & \ge 0 \, ,
    \end{align*}
    where Eq.~\eqref{eq:optimistic eq 1} applies the mean value theorem with $\dot{\wb}$ lying between $\hat{\wb}_t$ and $\wb^*$, Eq.~\eqref{eq:optimistic eq 2} uses it again for $\ddot{\wb}$ lying between $\dot{\wb}$ and $\hat{\wb}_t$, and the last inequality follows from the bound $\| \dot{\wb} - \hat{\wb}_t \|_{\Vb_t} \le \| \hat{\wb}_t - \wb^* \|_{\Vb_t} \le \sqrt{\beta_t}$.
    The upper bound can be derived by the similar argument.
\end{proof}

\section{Proof of Theorem~\ref{thm:regret bound}} \label{appx:main theorem}
\begin{proof}[Proof of Theorem~\ref{thm:regret bound}]
    Suppose that Lemma~\ref{lemma:confidence ball} holds.
    Then, by Lemma~\ref{lemma:optimistic utility}, for all $t_0 + 1 \le t \le T$ and $i \in S_t$, we have $f_{\wb^*} (\xb_{ti}) \le z_{ti}$.
    By Lemma~\ref{aux_lemma:monotonicity of expected reward of optimal assortment}, the expected reward corresponding to the optimal assortment $S_t^*$ is monotone in the MNL utilities, for all $t = t_0+1, \ldots, T$, we have
    \begin{equation} \label{eq:regret 1}
        R_t(S^*_t, \wb^*) \le \tilde{R}_t(S_t^*) \le \tilde{R}_t(S_t) \, .
    \end{equation}
    If we denote $z_{ti}' = f_{\wb^*}(\xb_{ti}) + 2 \sqrt{\beta_t} \|\nabla f_{\hat{\wb}_{t}}(\xb_{ti})\|_{\Vb_{t}^{-1}} + 2\beta_t C_h/\lambda$, by Lemma~\ref{aux_lemma:monotonicity of R tilde}, which states that the optimistic expected reward $\tilde{R}_t(S)$ is monotonically increasing with respect to the utilities of the items in $S$, and by Lemma~\ref{lemma:optimistic utility} we have $z_{ti} \le z_{ti}'$, which implies
    \begin{equation} \label{eq:regret 2}
        \tilde{R}_t(S_t) = \sum_{i \in S_t} \frac{\exp(z_{ti}) r_{ti}}{1 + \sum_{j \in S_t} \exp(z_{tj})}
        \le \sum_{i \in S_t} \frac{\exp(z'_{ti}) r_{ti}}{1 + \sum_{j \in S_t} \exp(z'_{tj})} \, .
    \end{equation}

    Let $\tilde{\wb}^* := \argmin_{\tilde{\wb} \in \Wcal^*} \| \hat{\wb}_0 - \tilde{\wb} \|_2$ denote the parameter in the equivalence set $\Wcal^*$ that is closest to $\hat{\wb}_0$.
    By combining Eq.~\ref{eq:regret 1} and Eq.~\ref{eq:regret 2}, we bound the cumulative regret in Phase II as follows:
    \begin{align*}
        \sum_{t=t_0+1}^T R_t(S_t, \wb^*) - R_t(S_t, \wb^*)
        & = \sum_{t=t_0+1}^T R_t(S_t, \tilde{\wb}^*) - R_t(S_t, \tilde{\wb}^*)
        \\
        & \le \sum_{t=t_0+1}^T \tilde{R}_t(S_t) - R_t(S_t, \tilde{\wb}^*)
        \\
        & \le \sum_{t=t_0+1}^T \left(
             \frac{\sum_{i \in S_t} \exp(z'_{ti}) r_{ti}}{1 + \sum_{j \in S_t} \exp(z'_{tj})}
            -  \frac{\sum_{i \in S_t} \exp(f_{\tilde{\wb}^*}(\xb_{ti})) r_{ti}}{1 + \sum_{j \in S_t} \exp(f_{\tilde{\wb}^*}(\xb_{tj}))}
            \right) 
        \\
        & \le \sum_{t=t_0+1}^T \max_{i \in S_t} | z_{ti}' - f_{\tilde{\wb}^*}(\xb_{ti})|
        \numberthis \label{eq:regret 3}
        \\
        & = \sum_{t=t_0+1}^T 2 \sqrt{\beta_t} \|\nabla f_{\hat{\wb}_{t}}(\xb_{ti})\|_{\Vb_{t}^{-1}} + \frac{2 \beta_t C_h}{\lambda} \, ,
    \end{align*}
    where Eq.~\eqref{eq:regret 3} follows by Lemma~\ref{aux_lemma:difference between mnl expected reward}.
    By applying Cauchy-Schwarz inequality, we have
    \begin{align*}
        \sum_{t=t_0+1}^T \left( 2 \sqrt{\beta_t} \|\nabla f_{\hat{\wb}_{t}}(\xb_{ti})\|_{\Vb_{t}^{-1}} + \frac{2 \beta_t C_h}{\lambda} \right)
        & \le \sqrt{T \sum_{t=t_0+1}^T \left( 2 \sqrt{\beta_t} \|\nabla f_{\hat{\wb}_{t}}(\xb_{ti})\|_{\Vb_{t}^{-1}} + \frac{2 \beta_t C_h}{\lambda} \right)^2}
        \\
        & \le \sqrt{T \sum_{t=t_0+1}^T \left( 8 \beta_t \|\nabla f_{\hat{\wb}_{t}}(\xb_{ti})\|^2_{\Vb_{t}^{-1}} + \frac{8 \beta^2_t C^2_h}{\lambda^2}\right)} 
        \\
        & \le \sqrt{8 \beta_T T \sum_{t=t_0+1}^T \|\nabla f_{\hat{\wb}_{t}}(\xb_{ti})\|^2_{\Vb_{t}^{-1}} 
            +  \frac{8 \beta^2_t C^2_h T^2}{\lambda^2}}
        \\
        & \le \sqrt{16 T \beta_T d_w \log (1 + (T+1)C_g^2/(d_w \lambda)) +  \frac{8 \beta^2_t C^2_h T^2}{\lambda^2}}
        \\
        & = \tilde{\Ocal}
        \left( \kappa^{-2} \mu^{-1} d_w \sqrt{T} \right) \, ,            
    \end{align*}
    where the second inequality uses the inequality $(a+b)^2 \le 2 a^2 + 2 b^2$, the third inequality uses the monotonicity of $\beta_t$, and the last inequality follows by Lemma~\ref{sub_lemma:gradient elliptical}, and the final bound is obtained by setting 
    $\beta_T = \tilde{\Ocal}(\kappa^{-4} d_w \mu^{-2}), \lambda = \tilde{\Ocal}(\kappa^{-5/2} \mu^{-1} d_w \sqrt{T})$.
    Then, the cumulative regret incurred over Phases I and II is bounded by
    \begin{equation*}
        \mathrm{Regret}_{T} = t_0 + \tilde{ \Ocal } \left(\kappa^{-2} \mu^{-1}  d_w  \sqrt{T}  \right)
        = \tilde{ \Ocal } \left( \kappa^{-3/2} d_w \sqrt{T} + \kappa^{-2} \mu^{-1}  d_w  \sqrt{T}  \right) \, .        
    \end{equation*}
\end{proof}

\section{Auxiliary Lemmas}

\begin{lemma}[Lemma D.1 in~\citet{zhang2024contextual}] \label{aux_lemma:generalization bound of MLE for iid}
    Let $\Xcal$ be an instance space, $\Ycal$ a target space, and $p(y \mid x)=f^*(x,y)$ be the true conditional density for $(x,y) \in \Xcal \times \Ycal$, where $f^* \in \Fcal$.
    Assume $\Fcal= \{f_{\thetab}: \Xcal \times \Ycal \rightarrow [\beta, 1] \mid \thetab \in [0,1]^d, \beta > 0\}$ is a $1$-Lipschitz function class in the parameter $\thetab \in [0,1]^d$, i.e., $\|f_{\thetab_1}(x, \cdot ) - f_{\thetab_2}(x, \cdot )\|_\infty \le \| \thetab_1 - \thetab_2 \|_\infty\, $ for all $\thetab_1, \thetab_2 \in [0,1]^d$ and $x \in \Xcal$.
    Let $D = \{ (x_i, y_i) \}_{i=1}^n$ be i.i.d. samples with $x_i \sim \Hcal$ (unknown) and $y_i \sim p(\cdot \mid x_i)$, and define the empirical risk minimizer $\hat{f} = \argmin_{f \in \Fcal} \sum_{i=1}^n - \log f(x_i, y_i)$.
    Then for any $\delta \in (0,1)$, with probability at least $1-\delta$, 
    \begin{equation*}
        \EE_{x \sim \Hcal, y \sim p(\cdot \mid x)}[\log \hat{f}(x, y) - \log f^*(x, y)] \le \Ocal \left( \frac{d \log \frac{|\Ycal|}{\beta} \log \frac{n}{\beta} \log \frac{1}{\delta}}{n} 
        \right) \, .
    \end{equation*}
\end{lemma}

\begin{lemma}[Freedman's inequality~\cite{freedman1975tail}] \label{aux_lemma:Freedman}
    Suppose $X_1, \ldots, X_T$ is a martingale difference sequence with $|X_t| \le b$ for all $t \in [T]$.
    Let $V$ denote the sum of conditional variances,
    \begin{equation*}
        V = \sum_{t=1}^T \mathrm{Var}(X_i \mid X_1, \ldots, X_{i-1}) \, .
    \end{equation*}
    Then, for every $a, v > 0$, 
    \begin{equation*}
        \PP \left( \sum_{i=1}^T X_i \ge a, V \le v \right) \le \exp \left( \frac{- a^2}{2 v + 2ab/3} \right) \, .
    \end{equation*}
\end{lemma}

\begin{lemma}[Proposition C.1 in~\citet{lee2024nearly}] \label{aux_lemma:self-concordant-like in lee2024nearly}
    Let $f_{\wb}(\xb) = \xb^\top \wb$. Then the loss $\ell_t(\wb)$ in Eq.~\eqref{eq:linearized mnl loss} is $3 \sqrt{2}$-self-concordant-like.
\end{lemma}

\begin{lemma}[Theorem 4 in~\citet{tran2015composite}] \label{aux_lemma:self-concordence}
    Let $f: \RR^n \rightarrow \RR$ be a $M_f$-self-concordant-like function and let $x, y \in \mathrm{dom}(f)$, then it holds:
    \begin{equation*}
        \exp \left( - M_f \| y - x \|_2 \right) \nabla^2 f(x) \preceq \nabla^2 f(y) \, .
    \end{equation*}
\end{lemma}

\begin{lemma}[Lemma 10 in~\citet{abbasi2011improved}] \label{aux_lemma:det-trace ineq.}
    Let $\xb_1, \ldots, \xb_t \in \RR^d$ with $\| \xb_s \|_2 \le L$ for any $1 \le s \le t$.
    Let $\bar{\Vb}_t = \lambda \Ib_d + \sum_{s=1}^t \xb_s \xb_s^\top$ for some $\lambda > 0$.
    Then,
    \begin{equation*}
        \det \bar{\Vb}_t \le (\lambda + t L^2 / d )^d \, .
    \end{equation*}
\end{lemma}

\begin{lemma}[Lemma A.3 in~\citet{agrawal2019mnl}] \label{aux_lemma:monotonicity of expected reward of optimal assortment} 
    Let $\vb = [v_1, \ldots, v_N]^\top \in \RR_+^N$ be a utility vector and $\rb = [r_1, \ldots, r_N]^\top \in \RR_+^N$ a revenue vector.  
    The expected revenue of an assortment \( S \subset [N] \) under the utility vector $\vb$ is defined as
    \[
        R(S, \vb) := \sum_{i \in S} \frac{r_i v_i}{1 + \sum_{j \in S} v_j} \, .
    \]
    Let $\ub = [u_1, \ldots, u_N]^\top \in \RR_+^N$ be another utility vector such that $0 \le v_i \le u_i$ for all $i \in [N]$.  
    Then the following holds:
    \[
        R(S_{\vb}, \vb) \le R(S_{\vb}, \ub) \le R(S_{\ub}, \ub) \, ,
    \]
    where $S_{\vb} := \argmax_{S \subset [N]} R(S, \vb)$ and $S_{\ub} := \argmax_{S \subset [N]} R(S, \ub)$.
\end{lemma}

\begin{lemma}[Lemma H.2 in~\citet{lee2024nearly}] \label{aux_lemma:monotonicity of R tilde}
    Let $\rb = [r_1, \ldots, r_N] \in \RR^N$ with $0 \le r_i \le 1, \forall i \in [N]$ be given.
    For $\vb = [v_1, \ldots, v_N] \in \RR^N$ and $S \subset [N]$, let $R(S, \vb) = \frac{\sum_{i \in S} \exp(v_i) r_i}{1 + \sum_{j \in S} \exp(v_j)}$.
    Then for any $\ub = [u_1, \ldots, u_N]^\top \in \RR^N$ with $u_i \ge v_i$ for all $i \in [N]$, the following holds:
    \begin{equation*}
        R(S_{\vb}, \vb) \le R(S_{\vb}, \ub) \, ,
    \end{equation*}
    where $S_{\vb} = \argmax_{S \subset [N]} R(S, \vb)$.
\end{lemma}

\begin{lemma}[Lemma 3 in~\citet{oh2019thompson}] \label{aux_lemma:difference between mnl expected reward}
    Let $\ub = [u_1, \ldots, u_d], \vb=[v_1, \ldots, v_d] \in \RR^d$.
    Then for any $\rb = [r_1, \ldots, r_d] \in \RR^d$ with $|r_i|\le 1$, 
    \begin{equation*}
        \frac{\sum_{i=1}^d r_i \exp(u_i)}{1 + \sum_{j=1}^d \exp(u_j)} - \frac{\sum_{i=1}^d r_i \exp(v_i)}{1 + \sum_{j=1}^d \exp(v_j)} \le \max_{i \in [d]} | u_i - v_i | \, .
    \end{equation*}
\end{lemma}

\section{Details on Experiments \& Additional Results} \label{appx:additional experiments}
In this section, we provide additional details on the experimental settings and results discussed in the main text. 
All experiments are run on a computing cluster with Intel\textsuperscript{\textregistered} Xeon\textsuperscript{\textregistered} Gold 6526R (16-core, 2.8\,GHz, 37.5\,MB cache, 3 UPI, 195\,W).

\begin{figure*}[t]
  \centering
  \begin{subfigure}[b]{0.48\textwidth}
    \centering
    \includegraphics[width=\linewidth]{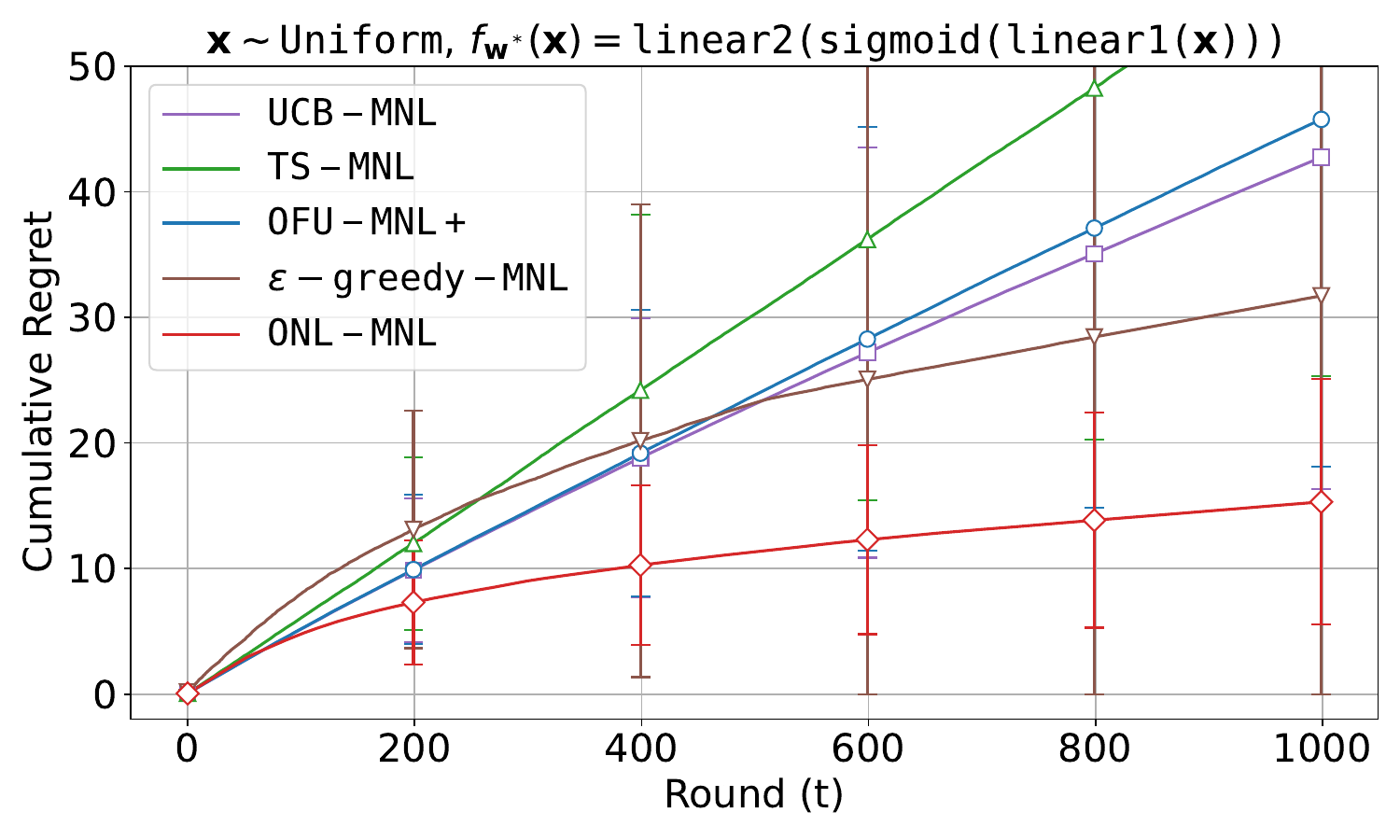}
    \caption{Realizability (Uniform)}
  \end{subfigure}
  \begin{subfigure}[b]{0.48\textwidth}
    \centering
    \includegraphics[width=\linewidth]{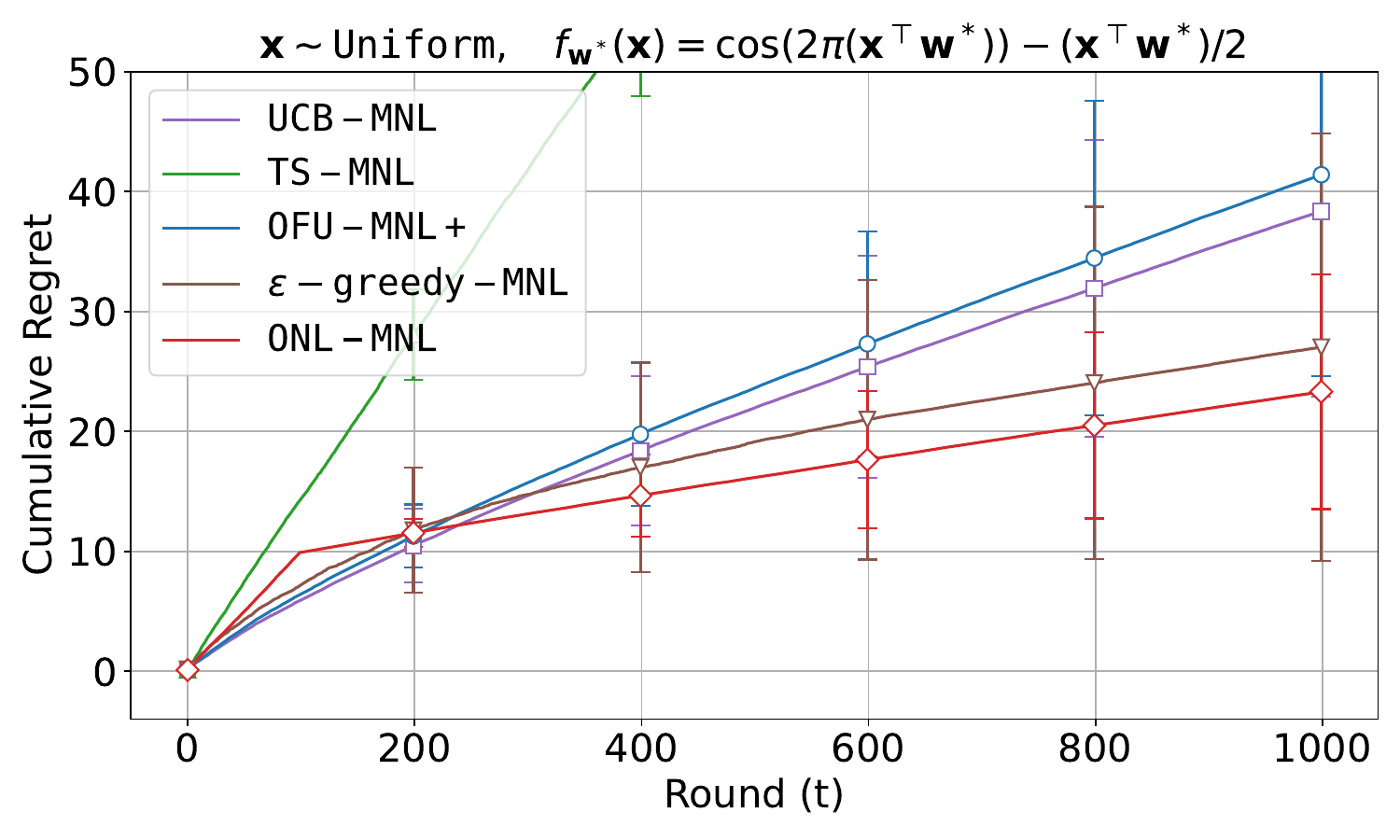}
    \caption{Misspecification (Uniform)}
  \end{subfigure}
  
  \caption{Cumulative regret comparison between $\algname$ (ours) and baselines under realizable and misspecified settings with uniform context distributions.}
  \label{fig:main exp_uniform}
\end{figure*}

\paragraph{Additional experiments under varying parameters.}
We conduct experiments over a wide range of values for $K$ and $d$.
The experimental setup is identical to that described in Section~\ref{sec:numerical experiments}.
Figures~\ref{fig:add_exp_gaussian} and~\ref{fig:add_exp_uniform} present additional evaluations of our proposed algorithm and baseline methods under Gaussian and uniform context distributions, respectively.
The results provide strong evidence that our proposed algorithm consistently outperforms existing MNL contextual bandit algorithms.

\paragraph{Semi-synthetic experiment with real-world dataset.}
In this section, we provide an additional semi-synthetic experiment leveraging a real-world dataset.
We used the IMDB Large Movie Review dataset~\cite{maas2011learning}, which consists of 50,000 movie reviews in text form, each labeled as either positive or negative.
To evaluate online assortment selection algorithms, one needs access to ground-truth choice probabilities for given assortments, which are typically unobservable in real-world datasets.
To address this, we first transformed the review texts into vector representations using TF-IDF (Term Frequency–Inverse Document Frequency), implemented via the `TfidfVectorizer` from scikit-learn.
We then applied truncated SVD to reduce the dimensionality of the vectors to $d = 30$.
This process yielded a dataset consisting of 50,000 context feature vectors with corresponding binary labels.

We then used 40,000 of these samples as training data to fit a binary classification model.
Specifically, we trained a two-layered neural network with 32 hidden nodes to classify the binary labels.
The learned model was subsequently used to define the true utility for each movie based on its extracted context feature, allowing us to formulate an online assortment selection task.
We then evaluated both the MNL bandit baselines and our proposed algorithm under this setting.

\begin{figure*}[t]
    \includegraphics[width=\linewidth]{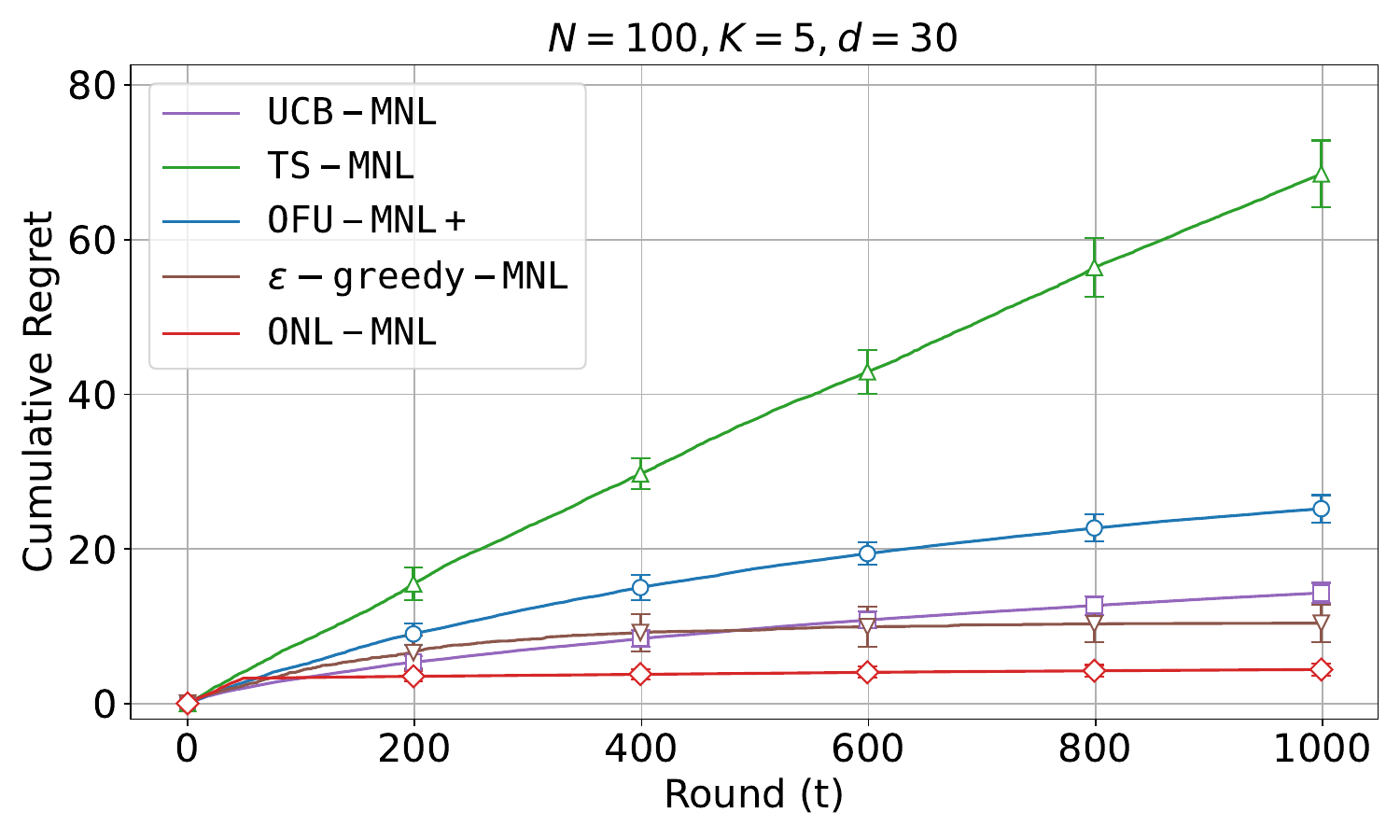}
    \caption{
    Cumulative regret comparison between $\algname$ (ours) and baseline methods in the semi-synthetic experimental setting based on the IMDB Large Movie Review dataset~\cite{maas2011learning}.
    }
    \label{fig:main exp_realworld}
\end{figure*}

\begin{figure*}[t!]
  \centering

  \begin{subfigure}[b]{0.49\textwidth}
    \centering
    \includegraphics[width=\linewidth]{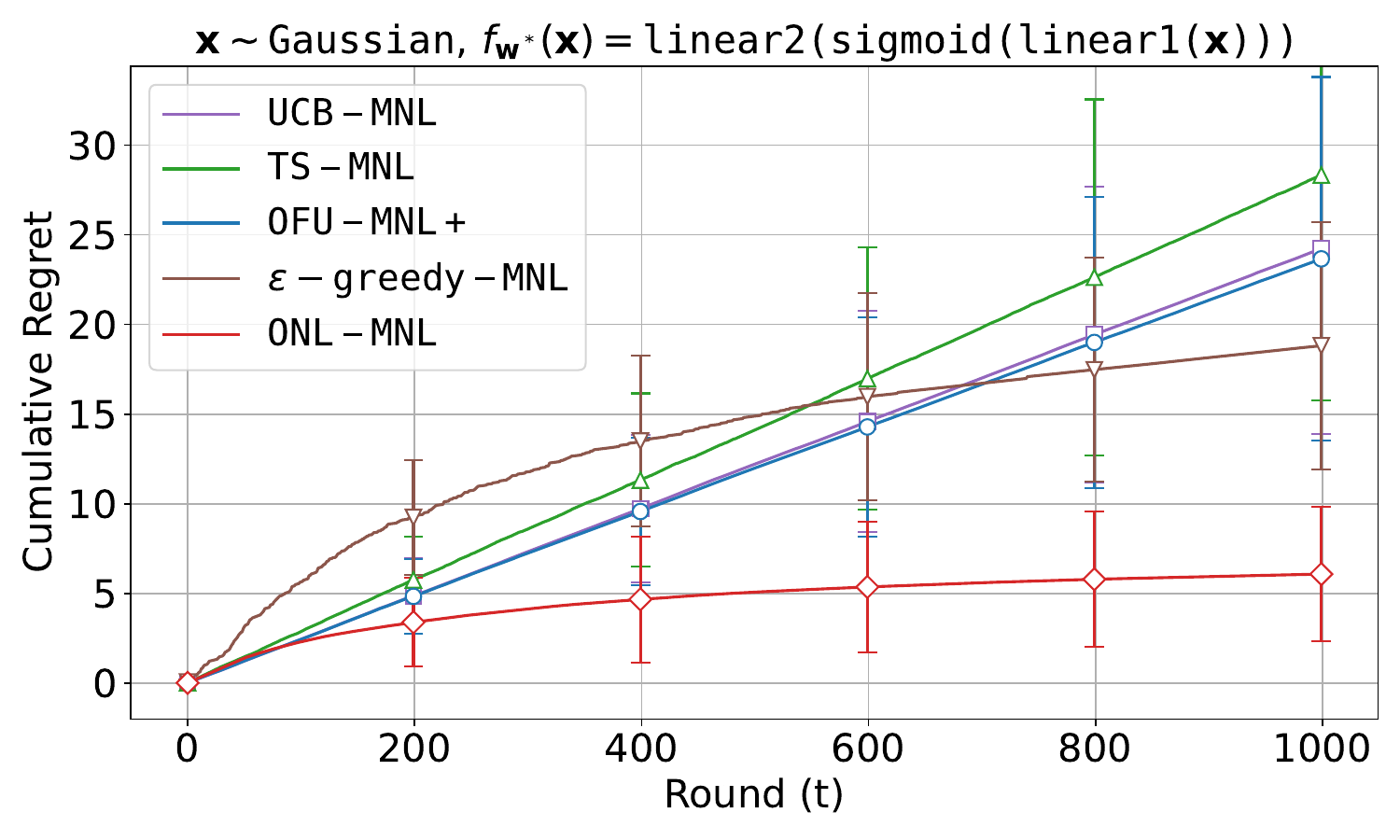}
    \caption{$N=100$, $d=3$, $K=10$}
  \end{subfigure}\hfill
  \begin{subfigure}[b]{0.49\textwidth}
    \centering
    \includegraphics[width=\linewidth]{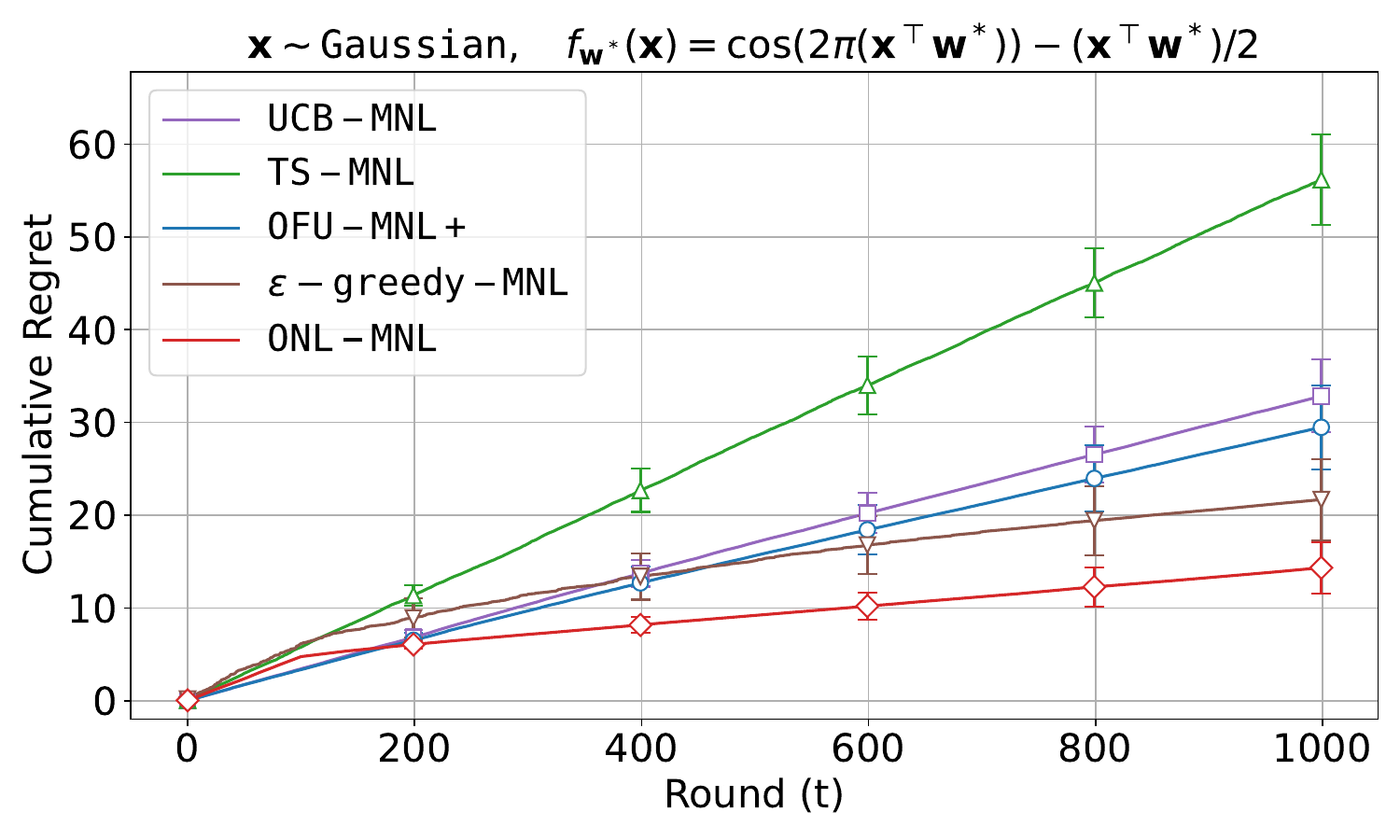}
    \caption{$N=100$, $d=3$, $K=10$}
  \end{subfigure}

  \begin{subfigure}[b]{0.49\textwidth}
    \centering
    \includegraphics[width=\linewidth]{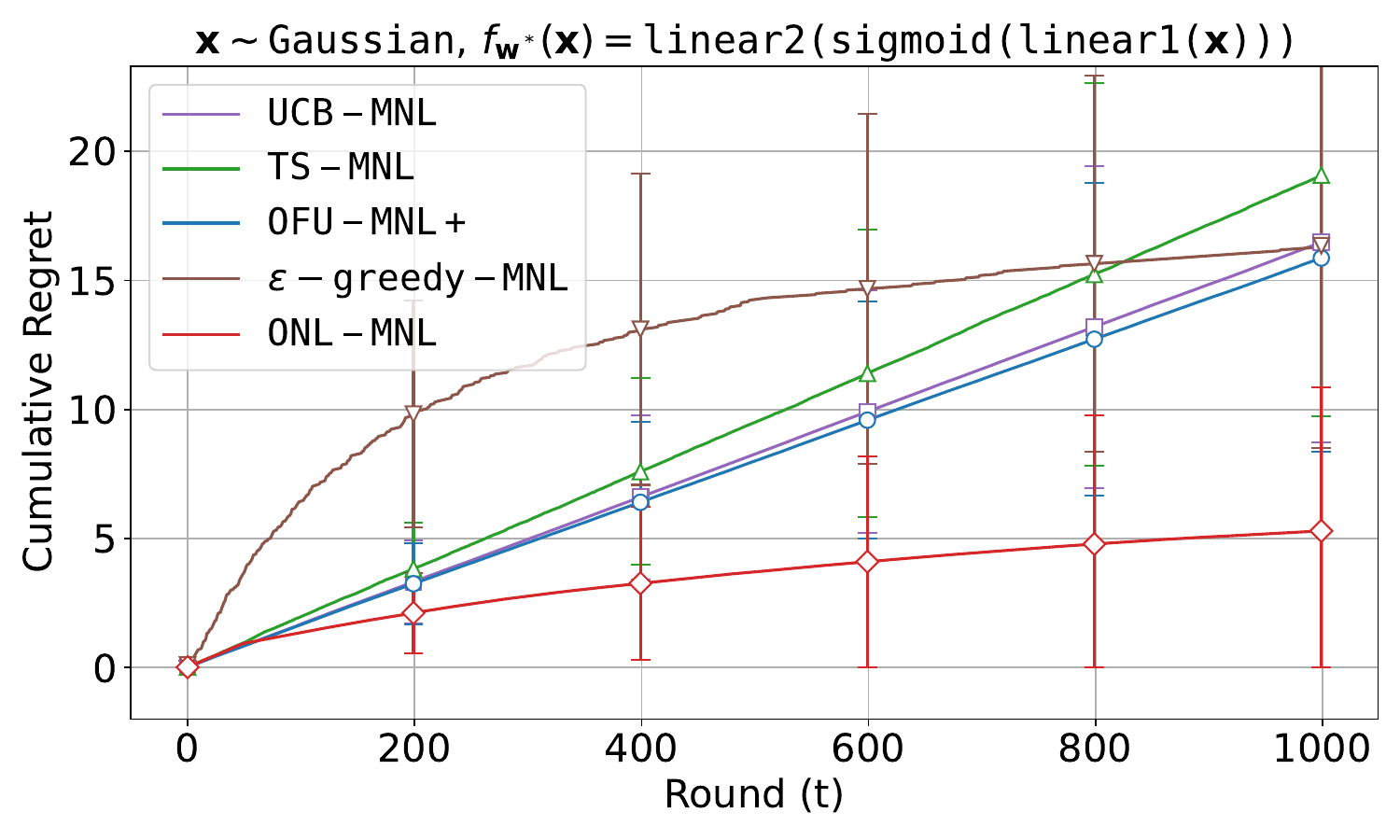}
    \caption{$N=100$, $d=3$, $K=15$}
  \end{subfigure}\hfill
  \begin{subfigure}[b]{0.49\textwidth}
    \centering
    \includegraphics[width=\linewidth]{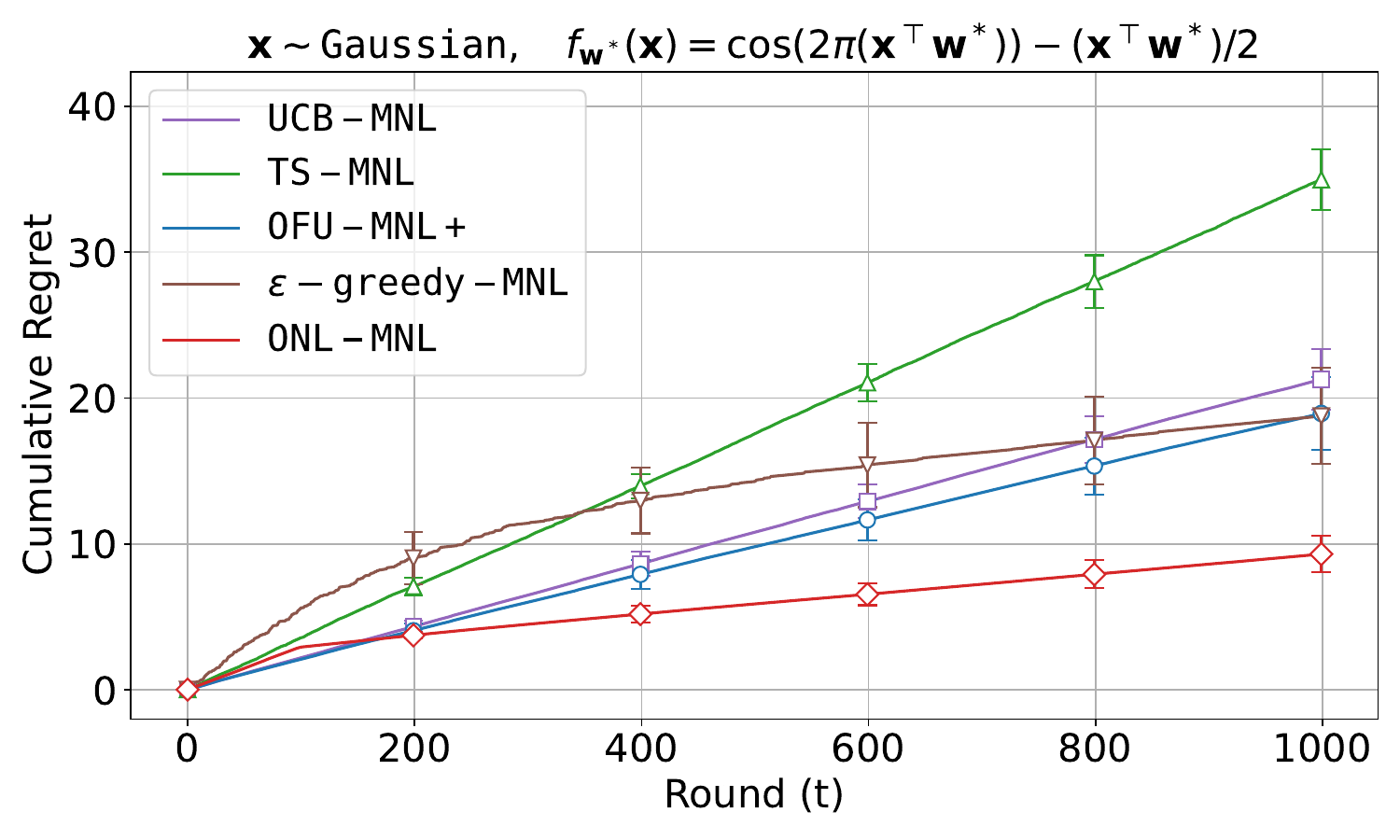}
    \caption{$N=100$, $d=3$, $K=15$}
  \end{subfigure}

  \begin{subfigure}[b]{0.49\textwidth}
    \centering
    \includegraphics[width=\linewidth]{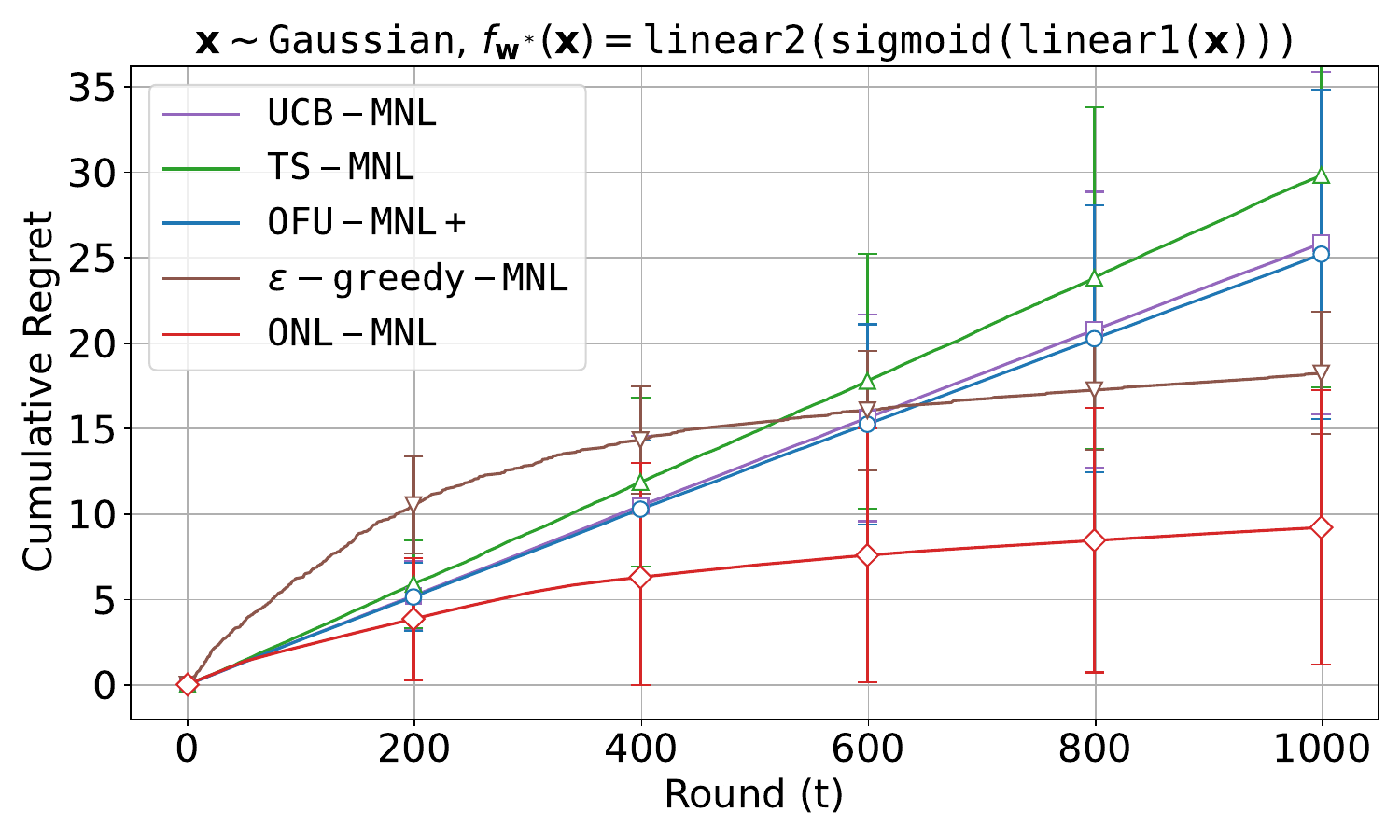}
    \caption{$N=100$, $d=5$, $K=10$}
  \end{subfigure}\hfill
  \begin{subfigure}[b]{0.49\textwidth}
    \centering
    \includegraphics[width=\linewidth]{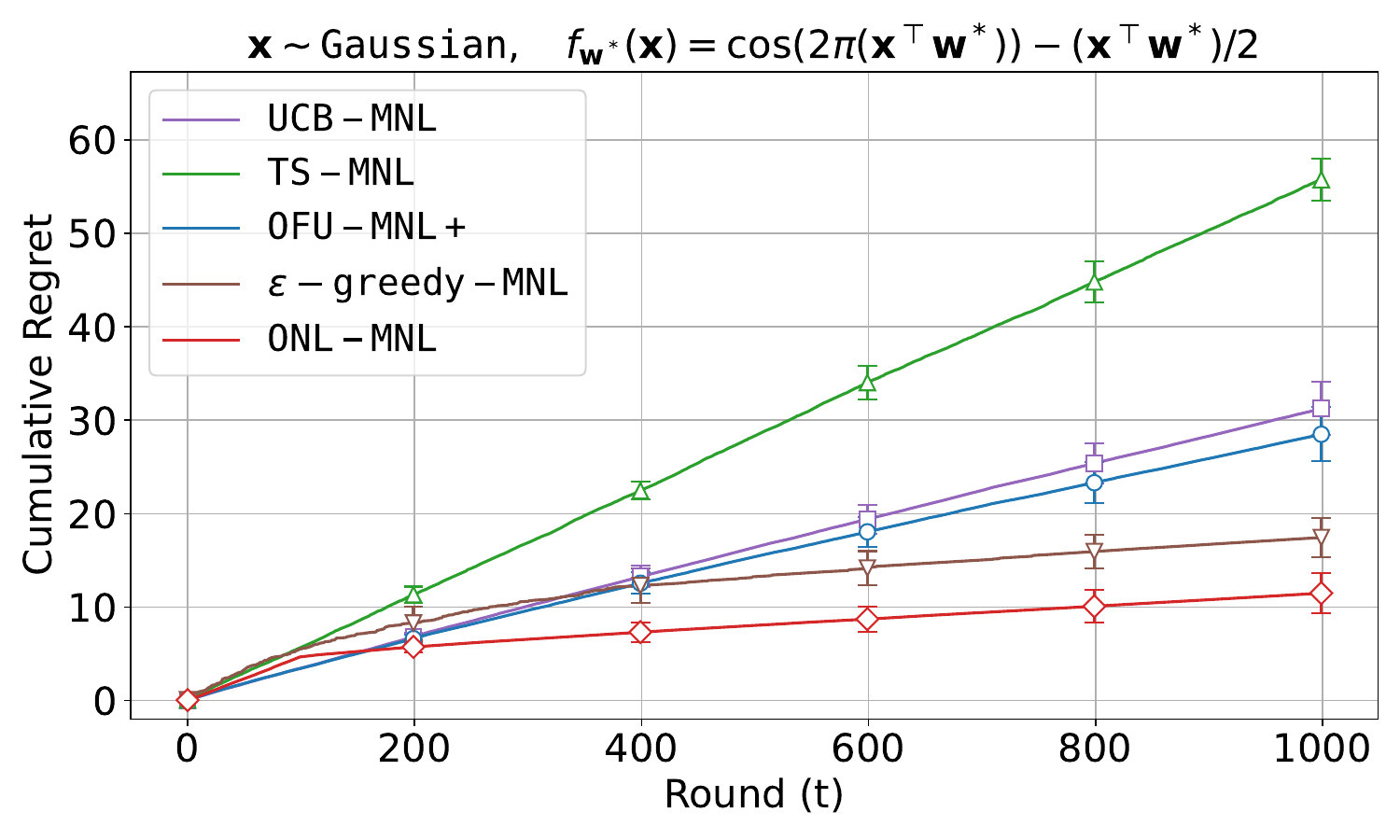}
    \caption{$N=100$, $d=5$, $K=10$}
  \end{subfigure}

  \begin{subfigure}[b]{0.49\textwidth}
    \centering
    \includegraphics[width=\linewidth]{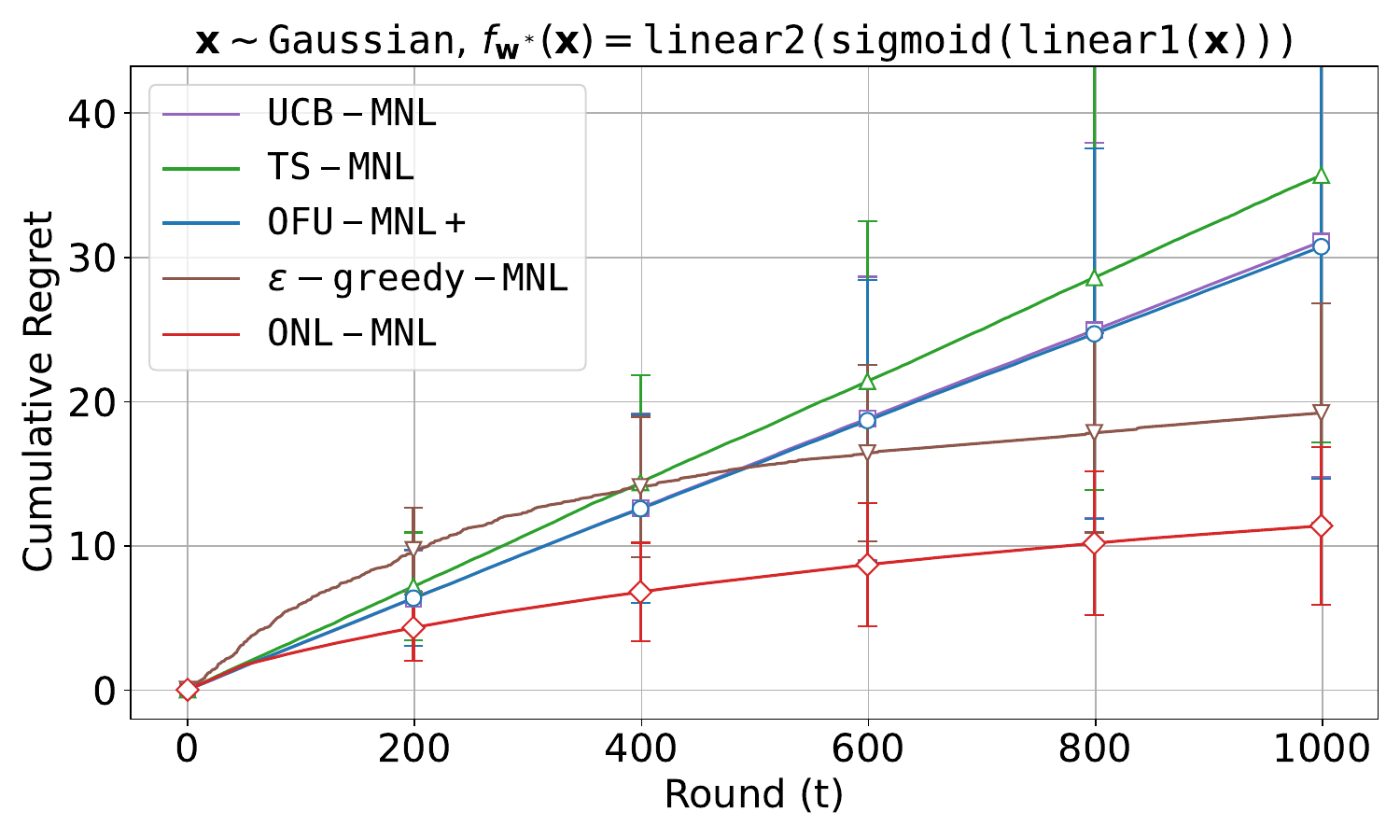}
    \caption{$N=100$, $d=10$, $K=10$}
  \end{subfigure}\hfill
  \begin{subfigure}[b]{0.49\textwidth}
    \centering
    \includegraphics[width=\linewidth]{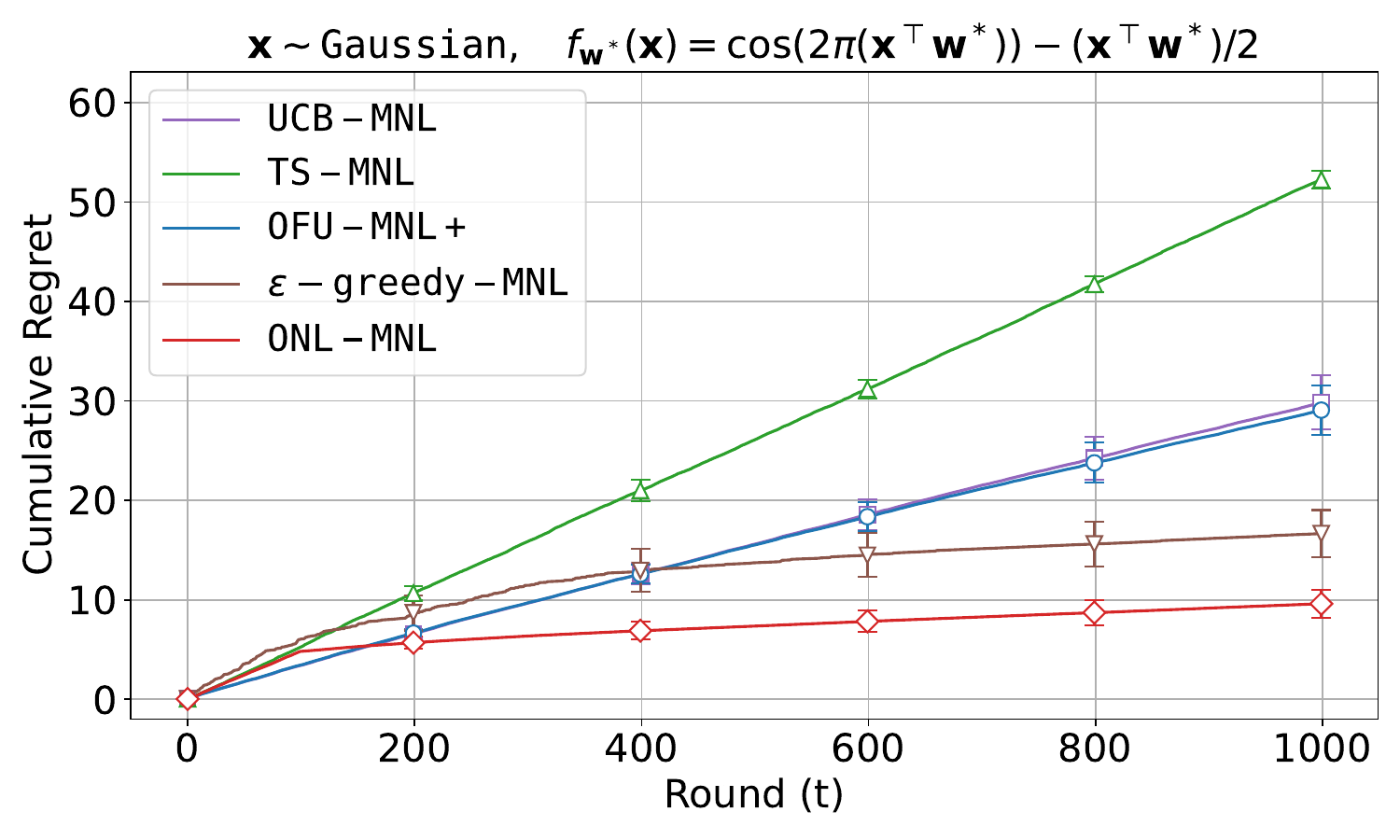}
    \caption{$N=100$, $d=10$, $K=10$}
  \end{subfigure}

  \caption{
  Cumulative regret comparison between $\algname$ (ours) and baseline methods under Gaussian contexts with varying context feature dimensions $d \in \{3, 5, 10\}$ and assortment sizes $K \in \{10, 15\}$.
  The left column shows results under the realizable setting, while the right column corresponds to the misspecified setting.
  Each curve represents the average cumulative regret over 10 independent runs.
  }
  \label{fig:add_exp_gaussian}
\end{figure*}

\begin{figure*}[t!]
  \centering

  \begin{subfigure}[b]{0.49\textwidth}
    \centering
    \includegraphics[width=\linewidth]{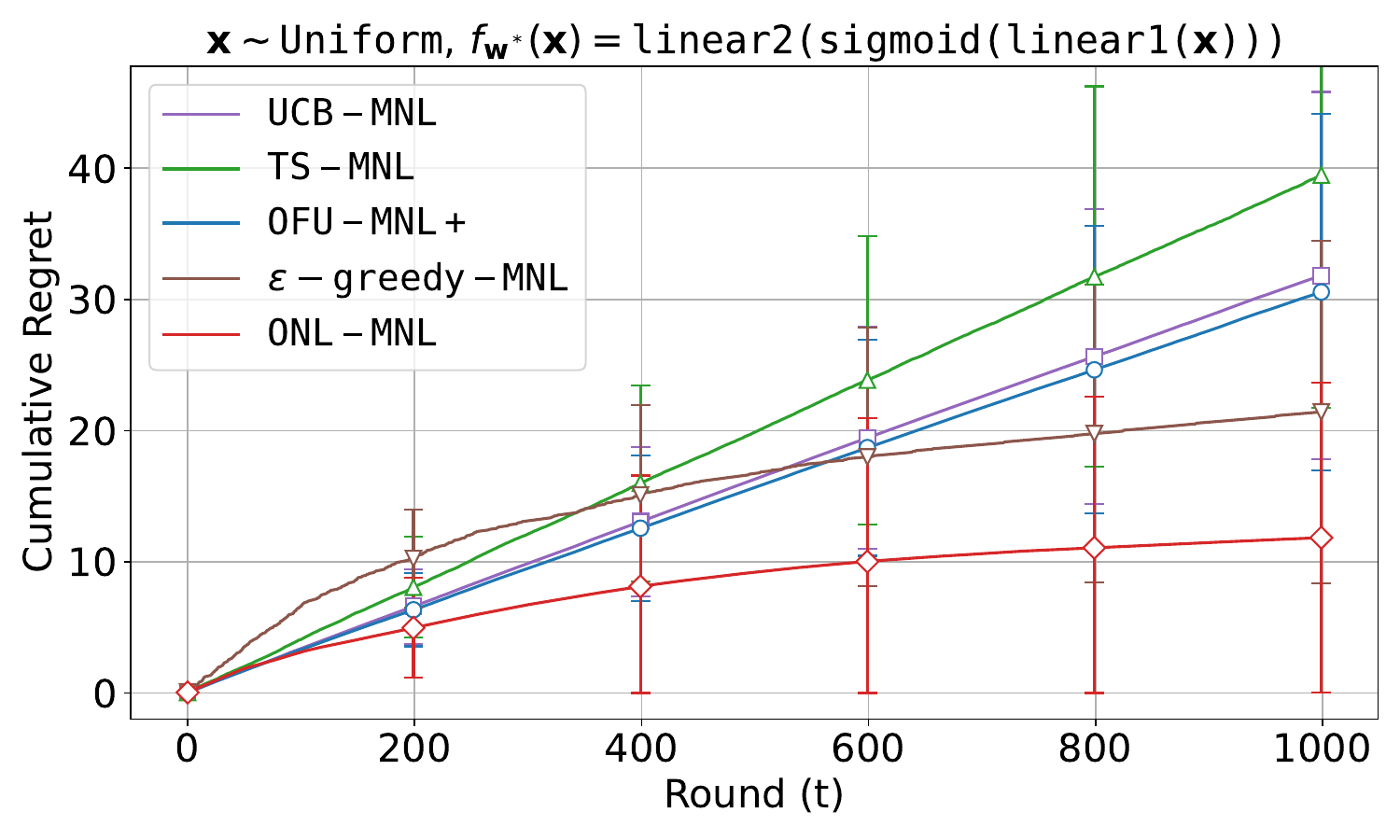}
    \caption{$N=100$, $d=3$, $K=10$}
  \end{subfigure}\hfill
  \begin{subfigure}[b]{0.49\textwidth}
    \centering
    \includegraphics[width=\linewidth]{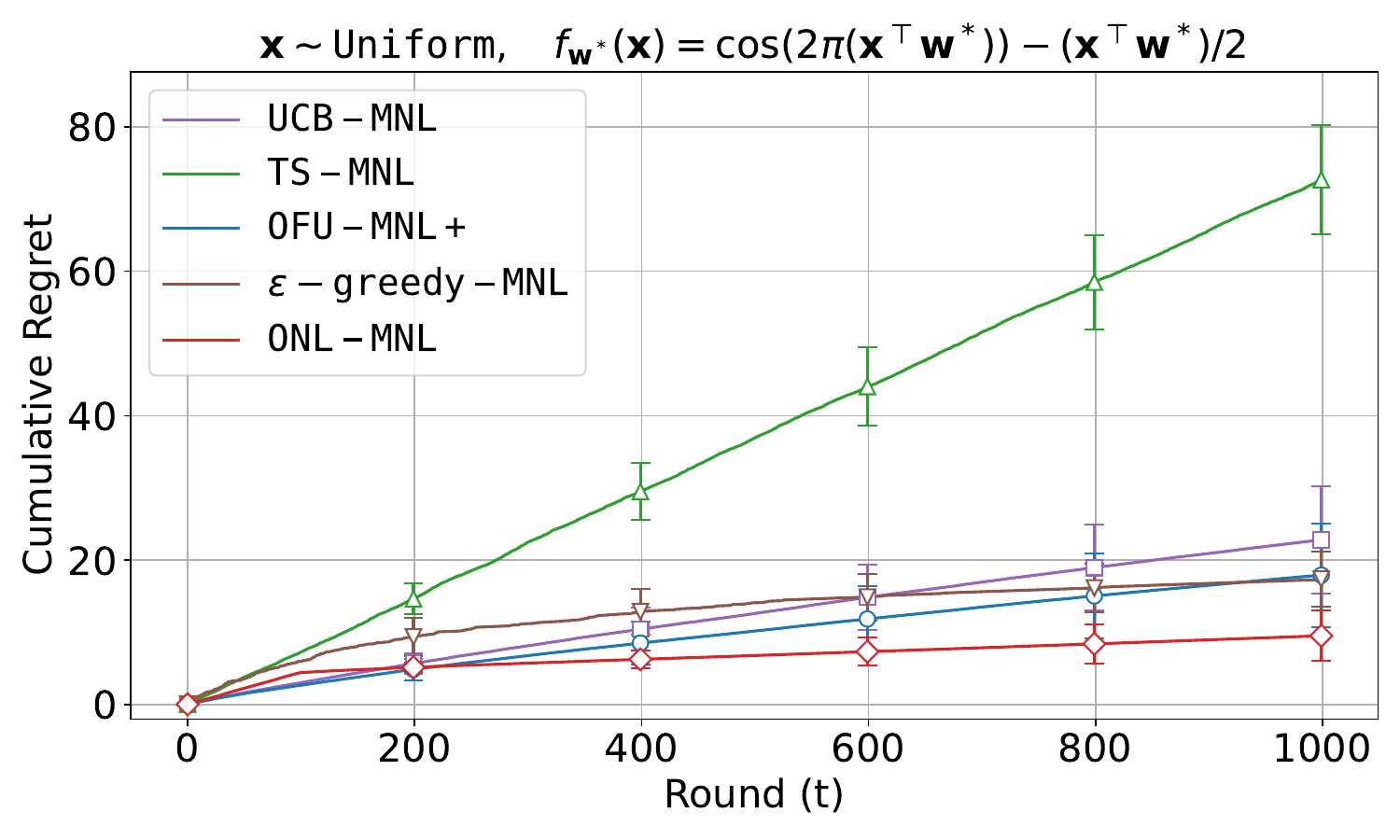}
    \caption{$N=100$, $d=3$, $K=10$}
  \end{subfigure}

  \begin{subfigure}[b]{0.49\textwidth}
    \centering
    \includegraphics[width=\linewidth]{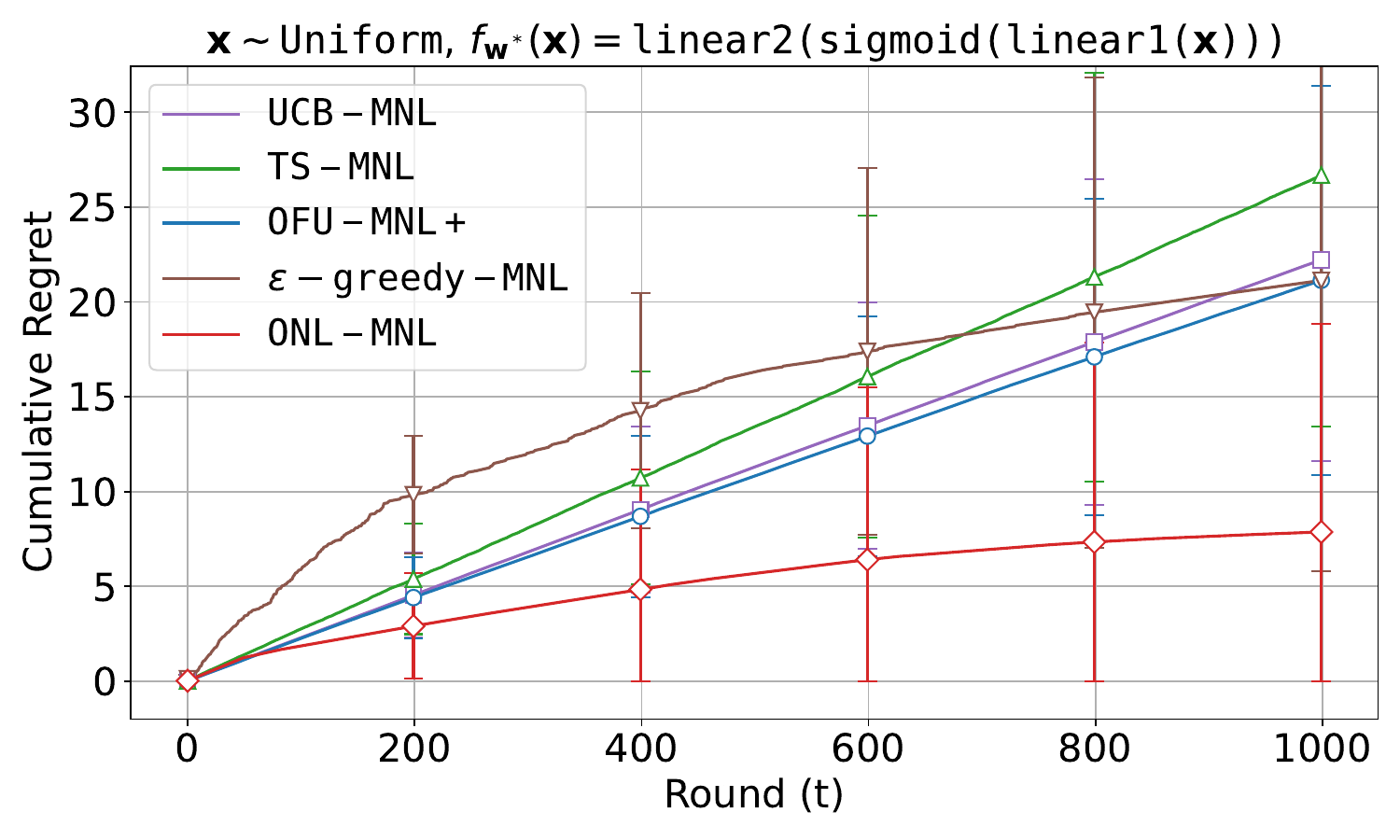}
    \caption{$N=100$, $d=3$, $K=15$}
  \end{subfigure}\hfill
  \begin{subfigure}[b]{0.49\textwidth}
    \centering
    \includegraphics[width=\linewidth]{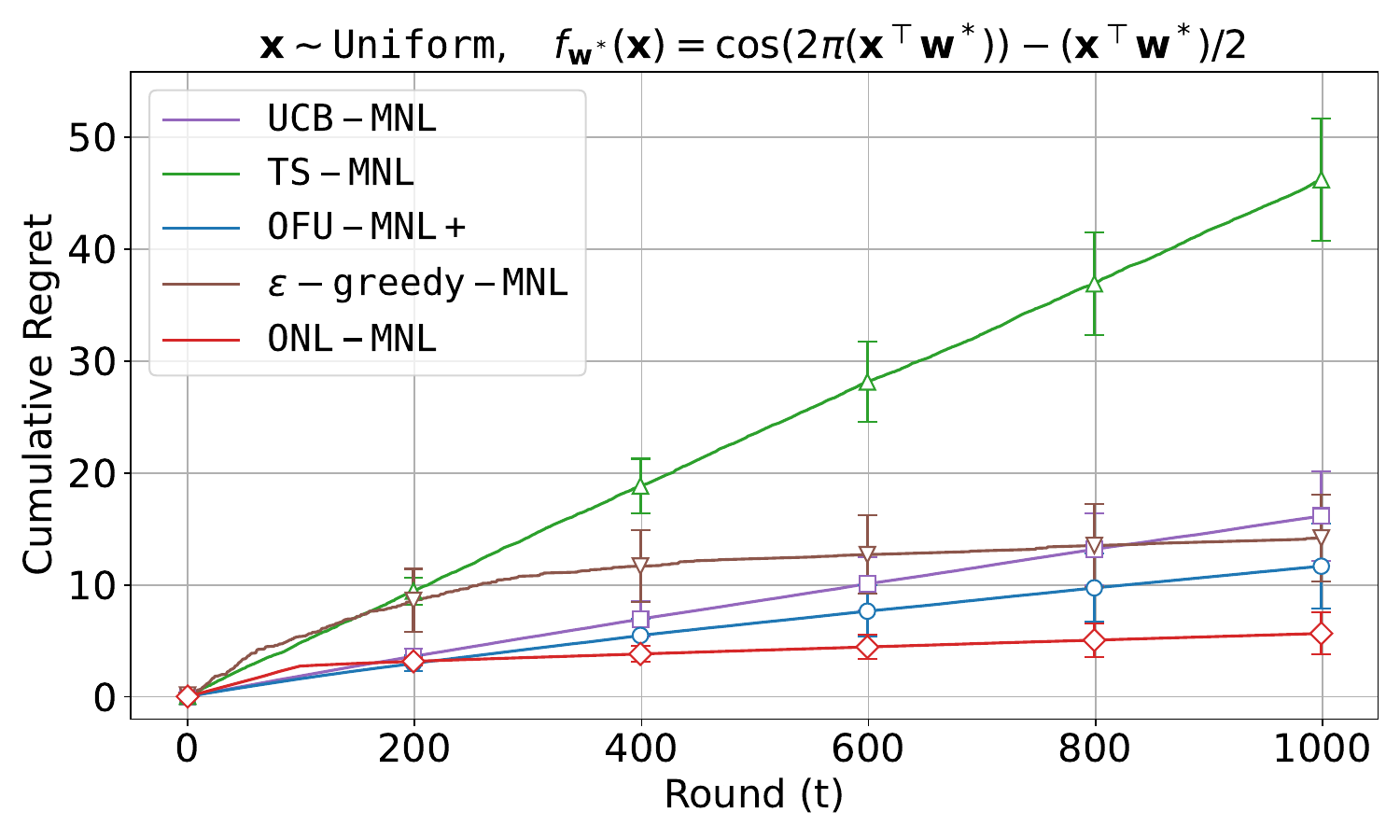}
    \caption{$N=100$, $d=3$, $K=15$}
  \end{subfigure}

  \begin{subfigure}[b]{0.49\textwidth}
    \centering
    \includegraphics[width=\linewidth]{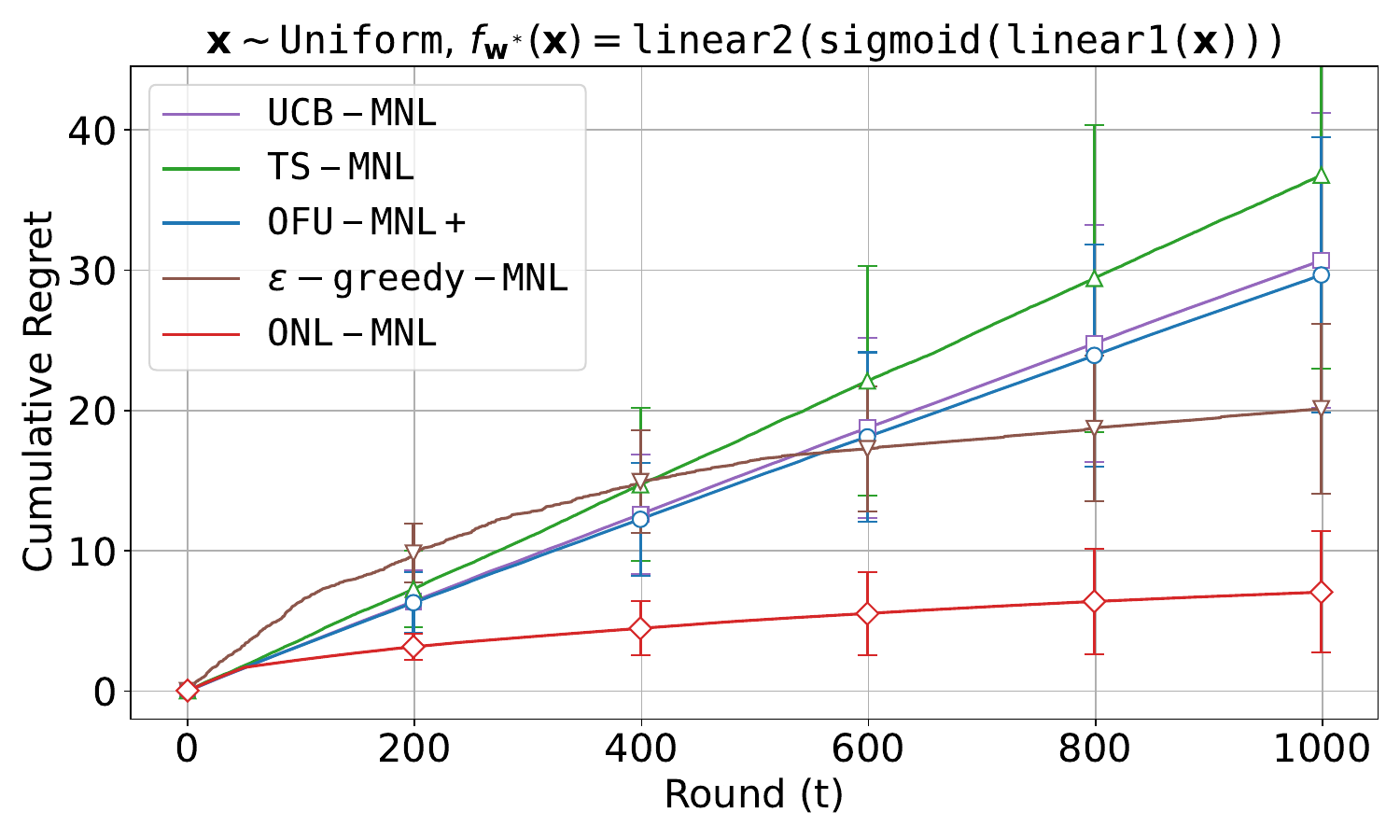}
    \caption{$N=100$, $d=5$, $K=10$}
  \end{subfigure}\hfill
  \begin{subfigure}[b]{0.49\textwidth}
    \centering
    \includegraphics[width=\linewidth]{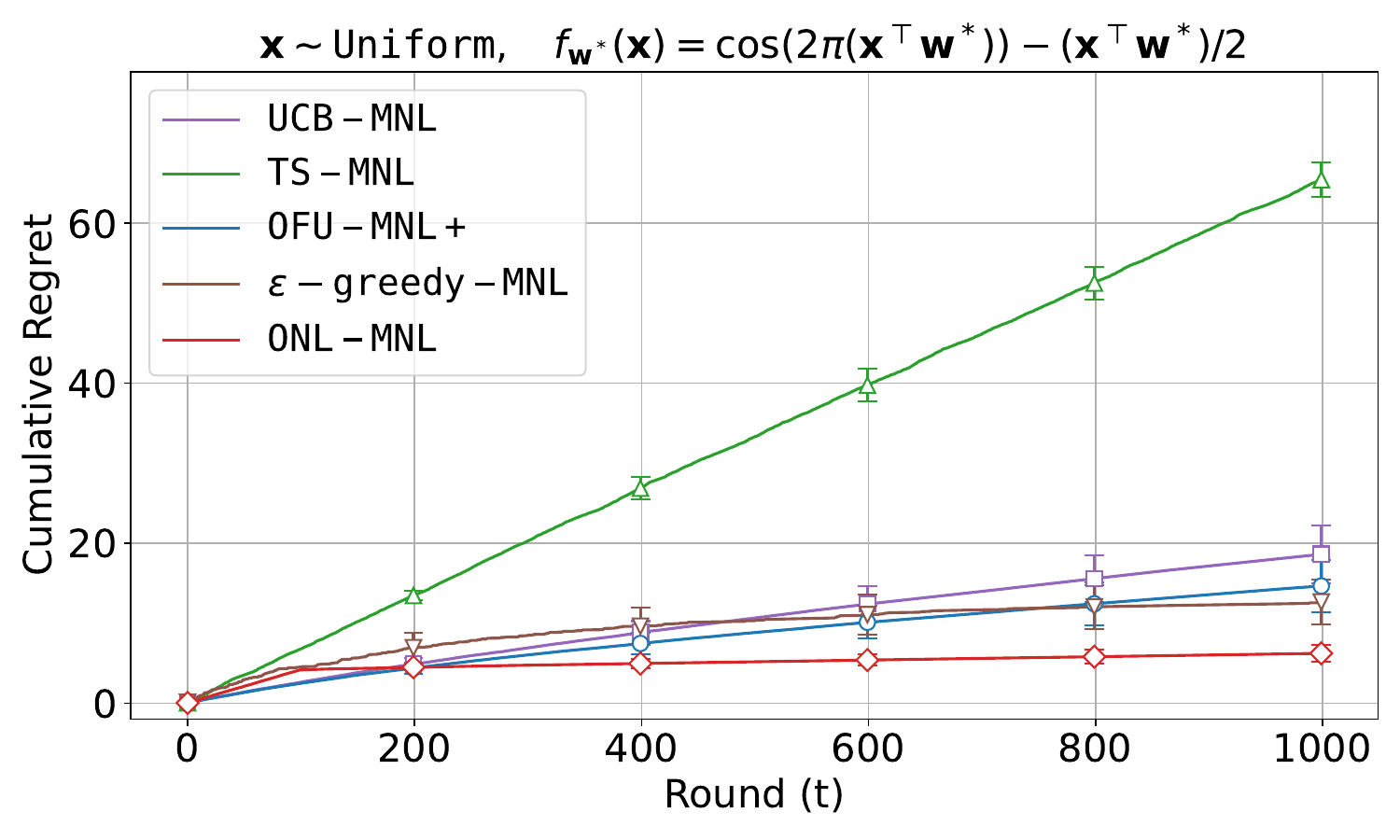}
    \caption{$N=100$, $d=5$, $K=10$}
  \end{subfigure}

  \begin{subfigure}[b]{0.49\textwidth}
    \centering
    \includegraphics[width=\linewidth]{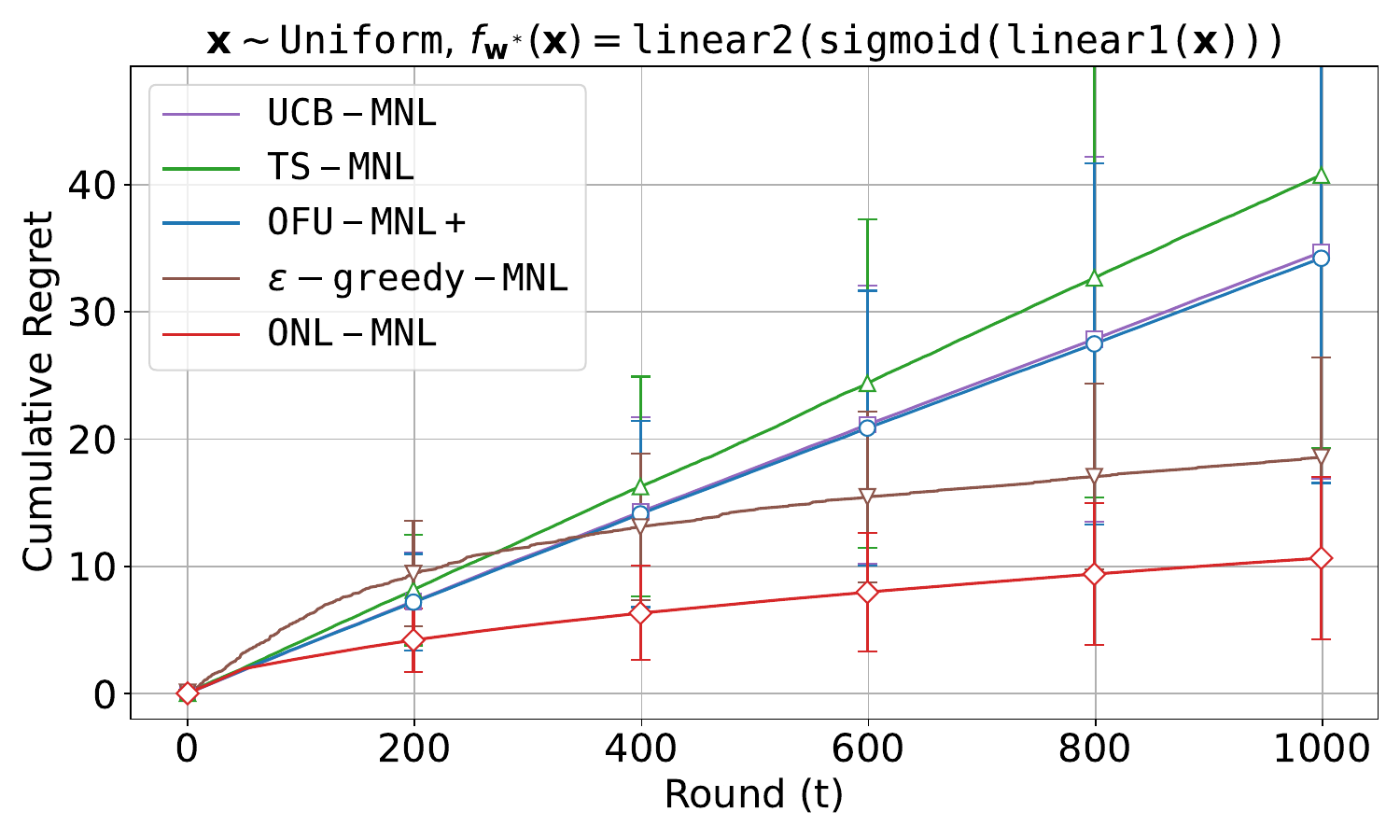}
    \caption{$N=100$, $d=10$, $K=10$}
  \end{subfigure}\hfill
  \begin{subfigure}[b]{0.49\textwidth}
    \centering
    \includegraphics[width=\linewidth]{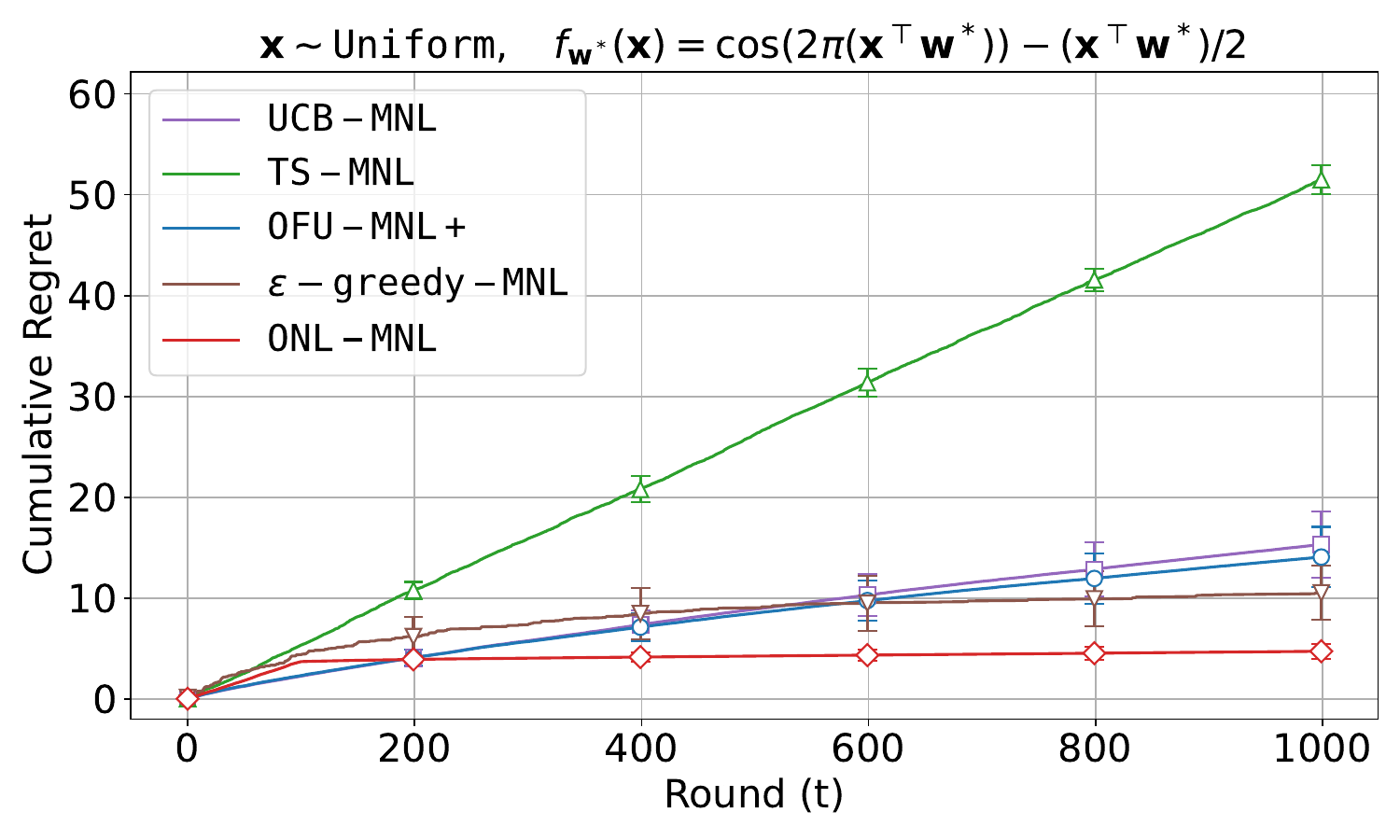}
    \caption{$N=100$, $d=10$, $K=10$}
  \end{subfigure}

  \caption{
  Cumulative regret comparison between $\algname$ (ours) and baseline methods under uniform contexts with varying context feature dimensions $d \in \{3, 5, 10\}$ and assortment sizes $K \in \{10, 15\}$.
  The left column shows results under the realizable setting, while the right column corresponds to the misspecified setting.
  Each curve represents the average cumulative regret over 10 independent runs.
  }
  \label{fig:add_exp_uniform}
\end{figure*}

At each round of the online experiment, we randomly sampled $N = 100$ movies from the remaining 10,000 held-out samples, and asked the algorithm to choose an assortment of size $K = 5$.
As in our main experiments, we used a uniform revenue of 1 for each item in the assortment (since revenue information is not given in the dataset).
The experimental results (cumulative regret over time step $T=1000$) are presented in Figure~\ref{fig:main exp_realworld}.

As shown in Figure~\ref{fig:main exp_realworld}, our proposed algorithm significantly outperforms the baseline methods.
These results demonstrate that our proposed algorithm not only achieves superior performance on synthetic benchmarks but also generalizes effectively to semi-synthetic settings constructed from real-world data. 
The consistent performance advantage over existing baselines highlights the superior ability of our approach to learn complex, non-linear utility structures, demonstrating strong robustness and adaptability to realistic, high-dimensional feature representations.
This experiment thus provides empirical evidence that our method can handle complex contextual structures beyond linear utility assumptions, supporting its potential applicability in real-world recommendation and decision-making environments.

\paragraph{Implementation of $\varepsilon \texttt{-greedy-MNL}$.}
We implement $\varepsilon \texttt{-greedy-MNL}$~\cite{zhang2024contextual} based on the $\varepsilon$-greedy-style method described in Eq.(4) of~\citet{zhang2024contextual}.
Since the official code for $\varepsilon \texttt{-greedy-MNL}$ is not publicly available, we tailor the implementation to our problem setting.
To isolate and assess the effectiveness of the exploration strategy, we set the utility function class of $\varepsilon \texttt{-greedy-MNL}$ to match the true utility function—a two-layer neural network with sigmoid activation.
We adopt an epoch-based update schedule, where the first epoch starts with length $1$, and each subsequent epoch doubles in length.
That is, after time step $t = 2^k - 1$ $(k = 1, 2, 3, \ldots)$, we update the utility parameter for the next epoch using observations collected in the current epoch.
We approximate the offline regression oracle using the Adam optimizer with a learning rate of $10^{-4}$ over $2000$ iterations.
The $\varepsilon$ parameter starts at $0.1$ and decays multiplicatively by a factor of $0.995$ after each step, with a minimum threshold of $0.001$.

\paragraph{Implementation of~$\algname$.}
After Phase I, the pilot estimator $\hat{\wb}_0$ is approximated by minimizing the negative log-likelihood using the Adam optimizer with a learning rate of $10^{-4}$ for $2000$ iterations.
The regularization parameter $\lambda$ is set as $\lambda = c_\lambda \cdot \kappa^{-5/2} d_w \sqrt{T}$, and the confidence radius $\beta_t$ is set as $\beta_t = c_\beta \cdot \kappa^{-4} d_w \frac{t}{T}$. Both scaling constants, $c_\lambda$ and $c_\beta$, are selected via grid search.

\end{document}